\relax
\documentclass[11pt]{article}
\usepackage{fullpage}
\usepackage{times}  
\usepackage{helvet}  
\usepackage{courier}  
\usepackage{url}  
\usepackage{graphicx}  

\usepackage[utf8]{inputenc} 
\usepackage[T1]{fontenc}    
\usepackage{url}            
\usepackage{booktabs}       
\usepackage{amsfonts}       
\usepackage{nicefrac}       
\usepackage{microtype}      
\usepackage{color}

\usepackage{graphicx} 
\usepackage{subfigure}

\usepackage{algorithm}
\usepackage{algorithmic}
\usepackage{amssymb}
\usepackage{amsthm}
\usepackage{amsmath,epsfig}
\usepackage{epstopdf}

\usepackage{enumitem}
\newif\ifsubmit

\submitfalse

\ifsubmit
\newcommand{\bo}[1]{}
\else
\newcommand{\bo}[1]{\textcolor{blue}{Bo: #1}}
\fi

\newtheorem{theorem}{Theorem}

\begin{document}

\title{Orthogonal Weight Normalization: Solution to Optimization over Multiple Dependent Stiefel Manifolds in Deep Neural Networks}
\author{
Lei Huang, Xianglong Liu, Bo Lang\\
Beihang University\\
\texttt{\small\{huanglei,xlliu,langbo\}@nlsde.buaa.edu.cn} \\
\and
Adams Wei Yu \\
CMU\\
\texttt{\small weiyu@cs.cmu.edu} \\
\and
Yongliang Wang \\
JD.COM\\
\texttt{\small wangyongliang1@jd.com} \\
\and
Bo Li \\
UC Berkeley\\
\texttt{\small crystalboli@berkeley.edu}
}
\date{}
\maketitle
\begin{abstract}
   Orthogonal matrix has shown advantages in training Recurrent Neural Networks (RNNs), but such matrix is limited to be square for the hidden-to-hidden transformation in RNNs. In this paper, we generalize such square orthogonal matrix to  orthogonal rectangular matrix and formulating this problem in feed-forward Neural Networks (FNNs) as Optimization over Multiple Dependent Stiefel Manifolds (OMDSM).
    We show that the rectangular orthogonal matrix can stabilize the distribution of network activations and regularize FNNs.
    We also propose a novel orthogonal weight normalization method to solve OMDSM. Particularly, it constructs orthogonal transformation over proxy parameters to ensure the weight matrix is orthogonal
    and back-propagates gradient information through the transformation during training.
   To guarantee stability,
we minimize the distortions between proxy parameters and canonical weights over all tractable orthogonal transformations.
In addition, we design an orthogonal linear module (OLM) to learn orthogonal filter banks in practice, which can be used as an alternative to standard linear module.
Extensive experiments demonstrate that by simply substituting OLM for standard linear module without revising any experimental protocols, our method largely improves the performance of the state-of-the-art networks, including Inception and residual networks on CIFAR and ImageNet datasets. In particular,
we have reduced the test error of wide residual network on CIFAR-100 from $20.04 \%$ to $18.61 \%$ with such simple substitution. Our code is available online for result reproduction.
\end{abstract}

\section{Introduction}
\label{sec_intro}
 Standard deep neural networks (DNNs) can be viewed as a composition of multiple simple nonlinear functions, each of which usually consists of one linear transformation with learnable weights or parameters followed by an element-wise nonlinearity. Such hierarchy and deep architectures equip DNNs with large capacity to represent complicated relationships between inputs and outputs. However, they also introduce potential risk of overfitting. Many methods have been proposed to address this issue, \emph{e.g.} weight decay \cite{1992_WD_Krogh} and Dropout \cite{2014_JMLR_Nitish} are commonly applied by perturbing objectives or adding random noise directly. These techniques can improve generalization of networks, but hurt optimization efficiency, which means one needs to train more epochs to achieve better performance.
This naturally rises one question: is there any technique that can regularize DNNs to guarantee generalization while still guarantee efficient convergence?

To achieve this goal, we focus on the orthogonality constraint, which is imposed in linear transformation between layers of DNNs. This technique performs optimization over low embedded submanifolds, where weights are orthogonal, and thus regularizes networks. Besides, the orthogonality implies energy preservation, which is extensively explored for filter banks in signal processing and guarantees that energy of activations will not be amplified~\cite{2006_TIP_Zhou} . Therefore, it can stabilize the distribution of activations over layers within DNNs ~\cite{2015_NIPS_Guillaume,2017_ICLR_guez} and make optimization more efficient.

Orthogonal matrix has been actively explored in Recurrent Neural Networks (RNNs)~\cite{2016_CoRR_Dorobantu,2016_ICML_Arjovsky,2016_NIPS_Wisdom,2017_ICML_Eugene,2017_ICML_Mhammedi}. It helps to avoiding gradient vanishing and explosion problem in RNNs due to its energy preservation property~\cite{2016_CoRR_Dorobantu}.
However, the orthogonal matrix here is limited to be square for the hidden-to-hidden transformation in RNNs. More general setting
of learning \emph{orthogonal rectangular matrix} is barely studied in DNNs~\cite{2017_Corr_Harandi}, especially in deep Convolutional Neural Networks (CNNs)~\cite{2016_Corr_Ozay}.
We formulate such a problem as Optimization over Multiple Dependent Stiefel Manifolds (OMDSM), due to the fact that  the weight matrix with orthogonality constraint in each layer is an embedded Stiefel Manifold \cite{2008_Book_Absil} and the weight matrix in certain layer is affected by those in preceding layers in DNNs.

To solve OMDSM problem, one straightforward idea is to use Riemannian optimization method that is extensively used for single manifold  or multiple independent manifolds problem, either in optimization communities ~\cite{2008_Book_Absil,2011_Yale_Tagarre,2012_SIAM_Absil,2013_MP_Wen,2015_JMLR_Zoubin} or in applications to the hidden-to-hidden transformation of RNNs ~\cite{2016_NIPS_Wisdom,2017_ICML_Eugene}. However, Riemannian optimization methods suffer instability in convergence or inferior performance in deep feed-forward neural networks based on our comprehensive experiments. Therefore, a stable
 method is  highly required for OMDSM problem.

Inspired by the orthogonality-for-vectors problem~\cite{2012_Bk_Garthwaite}  and the fact that eigenvalue decomposition is differentiable ~\cite{2015_ICCV_Ionescu},
 we propose a novel proxy parameters based solution referred to as \emph{orthogonal weight normalization}. Specifically, we devise explicitly a transformation that maps the proxy parameters to canonical weights such that the canonical weights are orthogonal. Updating is performed on the proxy parameters when gradient signal is ensured to back-propagate through the transformation. To guarantee stability, we minimize the distortions between proxy parameters and canonical weights over all tractable orthogonal transformations.

Based on \emph{orthogonal weight normalization}, we design orthogonal linear module for practical purpose. This module is a linear transformation with orthogonality, and can be used as an alternative of standard linear modules for DNNs. At the same time, this module is capable of stabilizing the distribution of activation in each layer, and therefore facilitates optimization process. Our method can also  cooperate well with other practical techniques in deep learning community, e.g., batch normalization ~\cite{2015_ICML_Ioffe}, Adam optimization \cite{2014_CoRR_Kingma} and Dropout \cite{2014_JMLR_Nitish}, and moreover improve their original performance.

Comprehensive experiments are conducted over Multilayer Perceptrons (MLPs) and CNNs.
By simply substituting the orthogonal linear modules for standard ones  without revising any experimental protocols, our method improves the performance of various state-of-the-art CNN architectures, including BN-Inception ~\cite{2015_ICML_Ioffe} and  residual networks ~\cite{2015_CVPR_He} over CIFAR \cite{2009_TR_Alex}  and ImageNet~\cite{2015_ImageNet} datasets. For example, on wide residual network, we improve the performance on CIFAR-10 and CIFAR-100 ~with test error as $3.73 \%$ and $18.61 \%$, respectively, compared to the best reported results  $4.17 \%$ and $20.04 \% $ in ~\cite{2016_CoRR_Zagoruyko}.

In summarization, our main contributions are as follows.
\begin{itemize}[itemsep=0pt]
\item
To the best of our knowledge, this is the first work to formulate the problem of learning orthogonal filters in DNNs as optimization over multiple dependent Stiefel manifolds problem (OMDSM). We further analyze two remarkable properties of orthogonal filters for DNNs: stabilizing the distributions of activation and regularizing the networks.
\item
 We conduct comprehensive experiments to show that several extensively used Riemannian optimization methods for single Stiefel manifold suffer severe instability in solving OMDSM,
  We thus propose a novel \emph{orthogonal weight normalization} method to solve OMDSM and show that the solution is stable and efficient in convergence.
 \item We devise an orthogonal linear module  to perform as an alternative to standard linear module for practical purpose.
\item
We apply the proposed method to various architectures including BN-Inception and residual networks, and achieve significant performance improvement over large scale datasets, including ImageNet.
\end{itemize}

\section{Optimization over Multiple Dependent Stiefel Manifolds}
\label{sec:solve_optimization}
Let $X\subseteq \mathcal{R}^d$ be the feature space, with $d$ the number of features.
Suppose the training set $\{(\mathbf{x}_i, \mathbf{y}_i)\}_{i=1}^{M}$ is comprised of feature vector $x_i \in X$ generated according to some unknown distribution $x_i \sim \mathcal{D}$, with $y_i$ the corresponding labels.
A standard feed-forward neural network with \emph{L}-layers can be viewed as a function $f(\mathbf{x}; \theta)$ parameterized by $\theta$, which is expected to fit the given training data and generalize well for unseen data points. Here $f(\mathbf{x}; \theta)$ is a composition of multiple simple nonlinear functions. Each of them usually consists of a linear transformation $\mathbf{s}^l= \mathbf{W}^{l} \mathbf{h}^{l-1}+ \mathbf{b}^l$ with learnable weights  $\mathbf{W}^l \in \mathbb{R}^{n_l \times d_l}$ and biases $\mathbf{b}^l \in \mathbb{R}^{{n}_l}$, followed by an element-wise nonlinearity: $\mathbf{h}^l=\varphi(\mathbf{s}^l)$. Here $l \in \{1,2,...,L\}$ indexes the layers.  Under this notation, the learnable parameters are $\theta=\{ {\mathbf{W}^l}, \mathbf{b}^l| l=1,2,\ldots,L \}$.
Training  neural networks is to minimize the discrepancy between the desired output $\mathbf{y}$ and the predicted output $f(\mathbf{x}; \theta)$. This discrepancy is usually described by a loss function $\mathcal{L}(\mathbf{y}, f(\mathbf{x}; \theta))$, and thus the objective is to optimize $\theta$ by minimizing the loss function: $\theta^* =\arg \min_{\theta} \mathbb{E}_{(\mathbf{x},\mathbf{y})\in D} [\mathcal{L}(\mathbf{y}, f(\mathbf{x}; \theta))]$.

\subsection{Formulation}
\label{sec:motivation}
 This paper targets to train deep neural networks (DNNs) with orthogonal rectangular weight matrix $\mathbf{W}^l \in \mathbb{R}^{n_l \times d_l} $ in each layer. Particularly, we expect to learn orthogonal filters of each layer (the rows of $\mathbf{W}$ ). We thus formulate it as a constrained optimization problem:
  \begin{eqnarray}
\label{eqn:objective}
	\theta^* & = \arg \min_{\theta} \mathbb{E}_{(\mathbf{x},\mathbf{y})\in D} \left[\mathcal{L} \left(\mathbf{y}, f \left(\mathbf{x}; \theta \right) \right) \right]  \nonumber \\
      & s.t.~~~~~  \mathbf{W}^l \in \mathcal{O}_l^{n_l \times d_l}, l=1,2,...,L
\end{eqnarray}
where the matrix family $\mathcal{O}_l^{n_l \times d_l}= \{ \mathbf{W}^l \in \mathbb{R}^{n_l \times d_l}: \mathbf{W}^l (\mathbf{W}^l)^T= \mathbf{I} \}$\footnote{ We first assume $n_l \leq d_l$ and will discuss how to handle the case $n_l > d_l$ in subsequent sections.} is real Stiefel manifold \cite{2008_Book_Absil,2015_JMLR_Zoubin}, which is an embedded sub-manifold of $\mathbb{R}^{n_l \times d_l}$. 
  The formulated problem has following characteristics: (1) the optimization space is over multiple  embedded submanifolds; (2) the embedded submanifolds $\{ \mathcal{O}_1^{n_1 \times d_1},...,\mathcal{O}_L^{n_L \times d_L} \}$ is dependent due to the fact that the optimization of weight matrix $\mathbf{W}^l$  is affected by those in preceding layers $\{ \mathbf{W}^i, i<l \}$; (3) moreover, the dependencies amplify as the network becomes deeper.
We thus call such a problem as Optimization over Multiple Dependent Stiefel Manifolds (OMDSM).
To our best knowledge,  we are the first to learn orthogonal filters for deep feed-forward neural networks and formulate such a problem as OMDSM. Indeed, the previous works ~\cite{2016_NIPS_Wisdom,2017_ICML_Eugene} that learning orthogonal hidden-to-hidden transformation in RNNs is over single manifold due to weight sharing of hidden-to-hidden transformation.

\subsection{Properties of Orthogonal Weight Matrix}
\label{sec:property}
Before solving OMDSM, we first introduce two remarkable properties of orthogonal weight matrix for DNNs.

\subsubsection{Stabilize the Distribution of Activations}

 Orthogonal weight matrix can  stabilize the distributions of activations in DNNs as illustrated in the following theorem.

 \begin{theorem}
 \label{prop1}
 Let $\mathbf{s}= \mathbf{W} \mathbf{x}$, where $\mathbf{W} \mathbf{W}^T = \mathbf{I}$ and $\mathbf{W} \in \mathbb{R}^{n \times d}$.  (1) Assume the mean of $\mathbf{x}$ is $\mathbb{E}_{\mathbf{x}}[\mathbf{x}]=\mathbf{0}$, and covariance matrix of $\mathbf{x}$ is $cov(\mathbf{x})=\sigma^2 \mathbf{I}$. Then  $\mathbb{E}_{\mathbf{s}}[\mathbf{s}]=\mathbf{0}$, $cov(\mathbf{s})=\sigma^2 \mathbf{I}$. (2) If $n=d$, we have $\|\mathbf{s}\|= \|\mathbf{x}\|$. (3) Given the back-propagated gradient $\frac{\partial \mathcal{L}}{\partial \mathbf{s}}$, we have  $\| \frac{\partial \mathcal{L}}{\partial \mathbf{x}} \|= \|\frac{\partial \mathcal{L}}{\partial \mathbf{s}}\|.$
 \end{theorem}
The proof of Theorem 1 is shown in Appendix ~\ref{proof_p1}.
The first point of Theorem 1 shows that in each layer of DNNs the weight matrix with orthonormality can maintain the activation $\mathbf{s}$ to be normalized and even de-correlated if the input is whitened. The normalized and de-correlated activation is well known for improving the conditioning of the \emph{Fisher information matrix} and accelerating the training of deep neural networks \cite{1998_NN_Yann,2015_NIPS_Guillaume,YuHLSC17}.
 Besides, orthogonal filters can well keep the norm of the activation and back-propagated gradient information in DNNs as shown by the second and third point of Theorem 1.
 \subsubsection{Regularize Neural Networks}
  Orthogonal weight matrix  can also ensure each filter to be \emph{orthonomal}: i.e. $\mathbf{w}_i^T \mathbf{w}_j=0, i \neq j$ and $ \| \mathbf{w}_i \|_2=1$, where $\mathbf{w}_i \in \mathbb{R}^d$ indicates the weight vector of the $i$-th neuron and $\|\mathbf{w}_i\|_2$ denotes the Euclidean norm of $\mathbf{w}_i$. This provides $n(n+1)/2$ constraints. Therefore, orthogonal weight matrix  regularizes the neural networks as the embedded Stiefel manifold $\mathcal{O}^{n \times d}$ with degree of freedom $nd- n(n+1)/2$~\cite{2008_Book_Absil}. Note that this regularization may harm the representation capacity if neural networks is not enough deep. We can  relax the constraint of orthonormal to orthogonal, which means we don't need $\| \mathbf{w}_i \|_2=1$. A practical method is to introduce a learnable scalar parameter $g$ to fine tune the norm of $\mathbf{w}$ \cite{2016_CoRR_Salimans}. This trick can recover the representation capacity of orthogonal weight layer to some extent, that is practical in shallow neural networks but for deep CNNs, it is unnecessary based on our observation. We also discuss how to trade off the regularization and optimization efficiency of orthogonal weight matrix in subsequent sections.

\section{Orthogonal Weight Normalization}

To solve OMDSM problem, one straightforward idea is to use  Riemannian optimization methods that are used for the hidden-to-hidden transform in RNNs~\cite{2016_NIPS_Wisdom,2017_ICML_Eugene}. However, we find that the Riemannian optimization methods to solve OMDSM suffered instability in convergence or inferior performance as shown in the experiment section.

Here we propose a novel algorithm to solve OMDSM problem via re-parameterization ~\cite{2016_CoRR_Salimans}.   For each layer $l$,  we represent the weight matrix $\mathbf{W}^{l}$ in terms of the proxy parameter matrix  $\mathbf{V}^{l} \in \mathbb{R}^{n_l \times d_l}$  as $\mathbf{W}^{l}= \phi(\mathbf{V}^{l})$, and parameter update is performed with respect to $\mathbf{V}^{l}$.
 By devising a transformation $\phi: \mathbb{R}^{n_l \times d_l} \rightarrow \mathbb{R}^{n_l \times d_l}$ such that $\phi(\mathbf{V}^{l}) * \phi(\mathbf{V}^{l})^T = \mathbf{I}$,   we can ensure the weight matrix $\mathbf{W}^{l}$ is  orthogonal. Besides, we require the gradient information back-propagates through the transformation $\phi$. An illustrative example is shown in Figure 1.  Without loss of generality, we drop the layer indexes of $\mathbf{W}^{l}$ and $\mathbf{V}^{l}$ for clarity.

\begin{figure}[t]
\centering
  \includegraphics[width=0.6\linewidth]{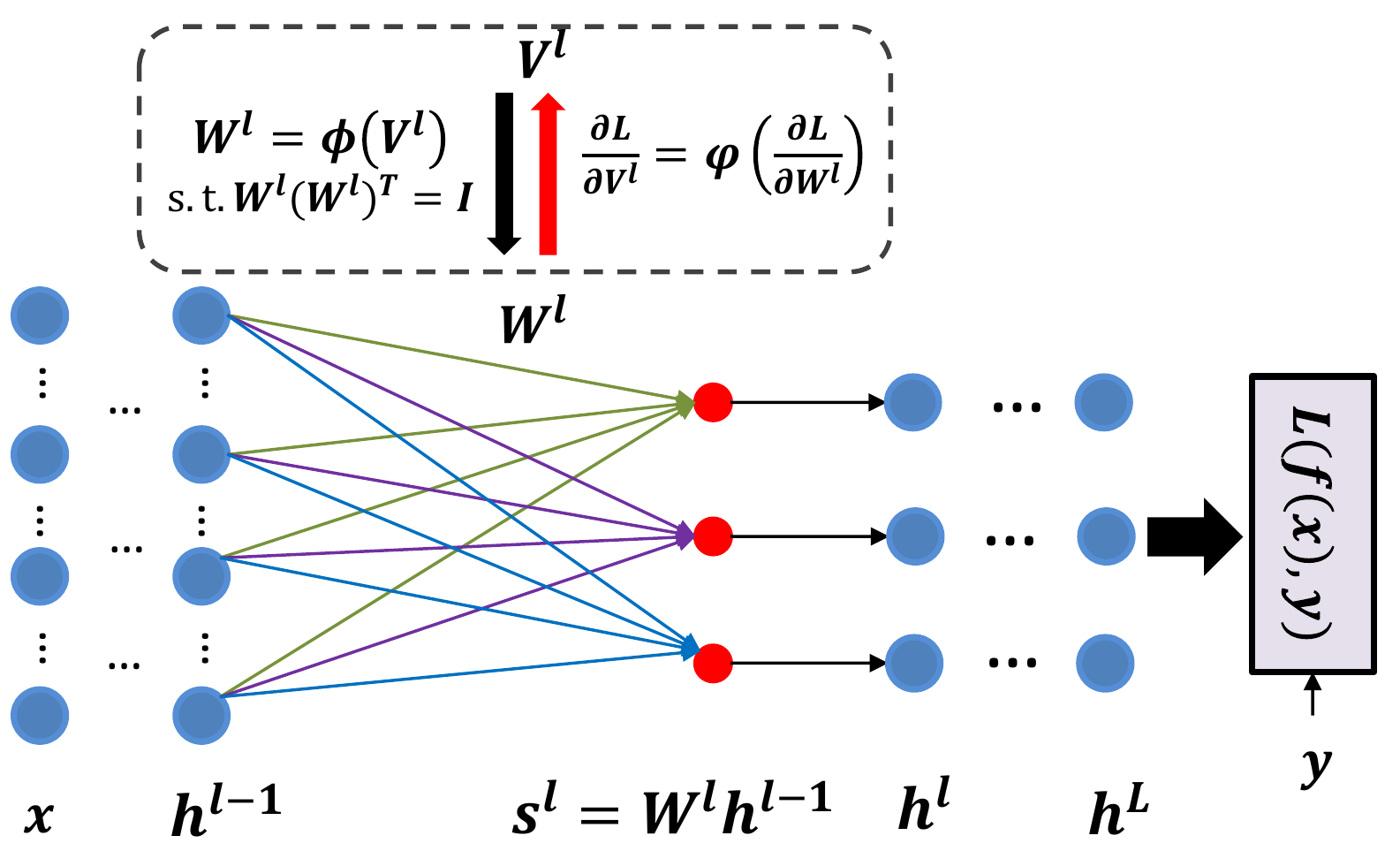}
  \caption{An illustrative example of orthogonal weight normalization in certain layer of neural networks (for brevity, we leave out the bias nodes).}
  \label{fig:motivation}
\end{figure}

\subsubsection{Devising Transformation}
 Inspired by the classic problem of orthogonality-for-vectors~\cite{2012_Bk_Garthwaite}, we represent $\phi(\mathbf{V})$  as linear transformation $\phi(\mathbf{V})=\mathbf{P} \mathbf{V}$.
 In general, vectors in this problem are usually assumed to be zero-centered. We therefore first center $\mathbf{V}$ by: $\mathbf{V}_C= \mathbf{V}- \mathbf{c} \mathbf{1}_d^T$ where $\mathbf{c}=\frac{1}{d}\mathbf{V}  \mathbf{1}_{d}$ and  $\mathbf{1}_d$ is $d$-dimension vector with all ones. The transformation is performed over $\mathbf{V}_C$.

There can be infinite $\mathbf{P}$ satisfying $\mathbf{W}=\mathbf{P} \mathbf{V}_C$ and $\mathbf{W} \mathbf{W}^T= \mathbf{I}$. For example, if $\mathbf{\hat{P}}$ is the solution,  $\mathbf{Q} \mathbf{\hat{P}}$ is also the solution where $\mathbf{Q}$ is an arbitrary orthogonal matrix $\mathbf{Q} \in \mathbb{R}^{n \times n}$, since we have $\mathbf{W} \mathbf{W}^T= \mathbf{Q} \mathbf{\hat{P}} \mathbf{V}_C \mathbf{V}_C^T \mathbf{\hat{P}}^T \mathbf{Q}^T= \mathbf{Q}  \mathbf{Q}^T =\mathbf{I} $. The question is which $\mathbf{P}$ should be chosen?

In order to achieve a stable solution, we expect the singular values of  Jacobians $\partial \mathbf{W}/\partial \mathbf{V}$ close to 1~\cite{2013_CoRR_Saxe}. However, this constraint is difficult to be formulated. We thus look for a relaxation and tractable constraint as minimizing the distortion between $\mathbf{W}$ and $\mathbf{V}_C$ in a least square way:
\begin{small}
  \begin{eqnarray}
\label{eqn:orthogonalTransform}
	& \min_{\mathbf{P}} tr \left( \left(\mathbf{W}-\mathbf{V}_C \right) \left(\mathbf{W}-\mathbf{V}_C \right)^T \right)  \nonumber \\
   & ~~~~~~ s.t. ~~~~ \mathbf{W}=\mathbf{P} \mathbf{V}_C ~~and ~~\mathbf{W} \mathbf{W}^T = \mathbf{I},
\end{eqnarray}
\end{small}
 \hspace{-0.01\linewidth}
 where $tr(\cdot)$ indicates the trace of matrix. We omit the derivation of solving this optimization to Appendix \ref{proof_p2}. The solution is $\mathbf{P}^*=\mathbf{D} \Lambda^{-1/2} \mathbf{D}^T $,
 where $\Lambda=\mbox{diag}(\sigma_1, \ldots,\sigma_n)$ and $\mathbf{D}$ represent the eigenvalues and eigenvectors of the covariance matrix $\Sigma=(\mathbf{V}-\mathbf{c}\mathbf{1}_d^T) (\mathbf{V}-\mathbf{c}\mathbf{1}_d^T)^T$.
 Based on this solution, we use the transformation as follows:
 \begin{small}
 \begin{eqnarray}
\label{eqn:whitening}
	\mathbf{W}=\phi(\mathbf{V})=\mathbf{D}\Lambda^{-1/2} \mathbf{D}^T (\mathbf{V}-\mathbf{c}  \mathbf{1}_d^T).
\end{eqnarray}
\end{small}
 \hspace{-0.01\linewidth}
We also consider another transformation $\mathbf{P}_{var}=\Lambda^{-1/2} \mathbf{D}^T$ without minimizing such distortions, and observe that  $\mathbf{P}_{var}$ suffers the instability problem and fails convergence  in subsequent experiments. Therefore, we hypothesize that minimizing distortions formulated by Eqn.~\ref{eqn:orthogonalTransform}   is essential  to ensure the stability of solving OMDSM.

\subsubsection{Back-Propagation}
We target to update proxy parameters $\mathbf{V}$, and therefore it is necessary to back-propagate the gradient information through the transformation $\phi(\mathbf{V})$.
 To achieve this, we use the result from matrix differential calculus \cite{2015_ICCV_Ionescu}, which combines the derivatives of eigenvalues and eigenvectors based on chain rule: given
 $\frac{\partial \mathcal{L}}{\partial \mathbf{D}} \in \mathbb{R}^{n\times n}$ and
$\frac{\partial \mathcal{L}}{\partial \Lambda} \in \mathbb{R}^{n\times n}$, where $\mathcal{L}$ is the loss function, the
back-propagate derivatives are $\frac{\partial \mathcal{L}}{\partial \Sigma}=\mathbf{D} ( ( \mathbf{K}^T \odot (\mathbf{D}^T \frac{\partial \mathcal{L}}{\partial \mathbf{D}} ) ) + (\frac{\partial \mathcal{L}}{\partial \Lambda})_{diag} ) \mathbf{D}^T$,
where $\mathbf{K} \in \mathbb{R}^{n\times n}$ is 0-diagonal and structured as
$\mathbf{K}_{ij}= \frac{1}{\sigma_i  -\sigma_j} [i\neq j]$,
 and $(\frac{\partial
  \mathcal{L}}{\partial \Lambda})_{diag}$ sets all off-diagonal elements of $\frac{\partial
  \mathcal{L}}{\partial \Lambda}$ to zero. The $\odot$ operator represents element-wise matrix multiplication.
Based on the chain rule, the back-propagated formulations for calculating $ \frac{\partial \mathcal{L}}{\partial \mathbf{V}}$ are shown as below.
\begin{small}
\begin{eqnarray}
\label{eq_dLdLambda}
&  \frac{\partial \mathcal{L}}{\partial \Lambda}&=-\frac{1}{2}   \mathbf{D}^T \frac{\partial \mathcal{L}}{\partial \mathbf{W}}  \mathbf{W}^T \mathbf{D}   \Lambda^{-1}   \nonumber \\
\label{eq_dLdD}
&  \frac{\partial \mathcal{L}}{\partial \mathbf{D}}&= \mathbf{D} \Lambda^{\frac{1}{2}} \mathbf{D}^T \mathbf{W}  \frac{\partial \mathcal{L}}{\partial \mathbf{W}}^T  \mathbf{D} \Lambda^{-\frac{1}{2}}
+  \frac{\partial \mathcal{L}}{\partial \mathbf{W}} \mathbf{W}^T \mathbf{D}  \nonumber \\
\label{eq_dLdSigma}
&  \frac{\partial \mathcal{L}}{\partial \Sigma}&=\mathbf{D} ( ( \mathbf{K}^T \odot (\mathbf{D}^T \frac{\partial \mathcal{L}}{\partial \mathbf{D}} ) ) +
(\frac{\partial \mathcal{L}}{\partial \Lambda} )_{diag} ) \mathbf{D}^T  \nonumber \\
\label{eq_dLdc}
&  \frac{\partial \mathcal{L}}{\partial \mathbf{c}}&= - \mathbf{1}_d^T  \frac{\partial \mathcal{L}}{\partial \mathbf{W}} ^T \mathbf{D} \Lambda^{-\frac{1}{2}} \mathbf{D}^T-	2 \cdot \mathbf{1}_d^T  (\mathbf{V}-\mathbf{c}  \mathbf{1}_d^T)^T  (\frac{\partial \mathcal{L}}{\partial \Sigma})_{s}  \nonumber \\
\label{eq_dLdx}
&  \frac{\partial \mathcal{L}}{\partial \mathbf{V}}&=  \mathbf{D} \Lambda^{-\frac{1}{2}} \mathbf{D}^T \frac{\partial \mathcal{L}}{\partial \mathbf{W}}
 + 2 (\frac{\partial \mathcal{L}}{\partial \Sigma} )_{s} (\mathbf{V}-\mathbf{c}  \mathbf{1}_d^T )  +\frac{1}{d} \frac{\partial \mathcal{L}}{\partial \mathbf{c}}^T  \mathbf{1}_d^T  \nonumber
\end{eqnarray}
\end{small}
where $(\frac{\partial \mathcal{L}}{\partial \Sigma})_{s}$ means symmetrizing $\frac{\partial \mathcal{L}}{\partial \Sigma}$ by $(\frac{\partial \mathcal{L}}{\partial \Sigma})_{s}=\frac{1}{2} (\frac{\partial \mathcal{L}}{\partial \Sigma}^T+\frac{\partial \mathcal{L}}{\partial \Sigma})$.
  Given $\frac{\partial \mathcal{L}}{\partial \mathbf{V}}$, we can apply regular gradient decent or other tractable optimization methods to update $\mathbf{V}$.  Note that symmetrizing $\frac{\partial \mathcal{L}}{\partial \Sigma}$ is necessary based on the perturbation theory, since wiggling $\mathbf{c}$ or $\mathbf{V}$  will make $\Sigma$ wiggle symmetrically.

 \begin{algorithm}[H]
  \caption{Forward pass of OLM.}
   \label{alg_forward}
    \begin{algorithmic}[1]
      \STATE \textbf{Input}: mini-batch input $\mathbf{H} \in \mathbb{R}^{d \times m} $ and parameters:  $\mathbf{b}  \in \mathbb{R}^{n \times 1}$, $\mathbf{V} \in \mathbb{R}^{n \times d} $.
     \STATE \textbf{Output}:  $\mathbf{S} \in \mathbb{R}^{n \times m} $ and $\mathbf{W} \in \mathbb{R}^{n \times d} $.
    \STATE	Calculate: $\Sigma = (\mathbf{V}-\frac{1}{d}\mathbf{V}  \mathbf{1}_{d}  \mathbf{1}_d^T)(\mathbf{V}-\frac{1}{d}\mathbf{V}  \mathbf{1}_{d}  \mathbf{1}_d^T)^T $.
    \STATE	Eigenvalue decomposition:  $\Sigma=\mathbf{D} \Lambda \mathbf{D}^T$.
    \STATE  Calculate $\mathbf{W}$ based on Eqn. \ref{eqn:whitening}.
    \STATE Calculate $\mathbf{S}$ as standard linear module does.
\end{algorithmic}
  \end{algorithm}

 \begin{algorithm}[H]
 \caption{Backward pass of OLM.}
   \label{alg_backprop}
\begin{algorithmic}[1]
     \STATE \textbf{Input}: activation derivative
      $\frac{\partial \mathcal{L}}{\partial \mathbf{S}} \in \mathbb{R}^{n \times m} $ and variables from respective forward pass.
     \STATE \textbf{Output}: $ \{ \frac{\partial \mathcal{L}}{\partial \mathbf{H} } \in \mathbb{R}^{d \times m} \}$,  $\mathbf{V} \in \mathbb{R}^{n \times d}$ and $\mathbf{b}  \in \mathbb{R}^{n \times 1}$.
    \STATE Calculate: $\frac{\partial \mathcal{L}}{\partial \mathbf{W}}$, 	$\frac{\partial \mathcal{L}}{\partial \mathbf{b}}$ and $ \frac{\partial \mathcal{L}}{\partial \mathbf{H}}$ as standard linear module does.
    \STATE Calculate $\frac{\partial \mathcal{L}}{\partial \mathbf{V}} $ base on Eqn.  \ref{eq_dLdx}
    \STATE Update $\mathbf{V}$ and $\mathbf{b}$.
\end{algorithmic}
  \end{algorithm}

\subsection{Orthogonal Linear Module}
Based on our orthogonal weight normalization method for solving  OMDSM, we build up the Orthogonal Linear Module (OLM) from practical perspective.  Algorithm \ref{alg_forward} and \ref{alg_backprop} summarize the forward and backward pass of  OLM, respectively. This module can be an alternative of standard linear module. Based on this, we can train DNNs with orthogonality constraints by simply substituting it for standard linear module without any extra efforts. After training,  we calculate the weight matrix $\mathbf{W}$ based on Eqn. \ref{eqn:whitening}.  Then $\mathbf{W}$  will be saved and used for inference as the standard module does.

\subsubsection{Convolutional Layer}
With regards to the convolutional layer parameterized by weights $\mathbf{W}^C \in \mathbb{R}^{n \times d \times F_h \times F_w}$ where  $F_h$ and  $F_w$ are the height and width of the filter,
 it takes feature maps $X \in \mathbb{R}^{d \times h \times r}$ as input, where $h$ and $r$ are the height and width of the feature maps, respectively. We denote $\Delta$ the set of spatial locations and $\Omega$ the set of spatial offsets. For each output feature map $k$ and its spatial location $\delta \in \Delta$, the convolutional layer computes the activation $\{ s_{k,\delta}\}$ as: $s_{k,\delta}=\sum_{i=1}^{d} \sum_{\tau \in \Omega }  w_{k,i,\tau} h_{i, \delta+\tau}  =<\mathbf{w}_k, \mathbf{h}_{\delta}>$.
 Here $\mathbf{w}_k$ eventually can be viewed as unrolled filter produced by $\mathbf{W}^C$. We thus reshape $\mathbf{W}^C$ as $\mathbf{W} \in \mathbb{R}^{n \times p }$  where $p=d \cdot F_h \cdot F_w$, and the  orthogonalization is executed over the unrolled weight matrix $\mathbf{W} \in \mathbb{R}^{n \times (d \cdot F_h \cdot F_w)}$.

 \subsubsection{Group Based Orthogonalization}
 In previous sections, we assume $n<=d$, and obtain the solution of OMDSM  such that the rows of $\mathbf{W}$ is orthogonal.
  To handle the case with $n>d$, we propose the \emph{group based orthogonalization} method. That is, we divide the weights $\{ w_i\}_{i=1}^{n}$ into groups with size $N_G<=d$ and the orthogonalization is performed over each group, such that the weights in each group is orthogonal.

 One appealing property of \emph{group based orthogonalization} is that we can use group size $N_G$ to control to what extent we regularize the networks. Assume $N_G$ can be divided by $n$, the free dimension of embedded manifold is $nd-n(N_G+1)/2$ by using \emph{orthogonal group} method. If we use $N_G=1$, this method reduces to Weight Normalization \cite{2016_CoRR_Salimans} without learnable scalar parameters.

 Besides, \emph{group based orthogonalization} is a practical strategy in real  application, especially reducing the computational burden. Actually, the cost of eigen decomposition with high dimension in GPU is expensive. When using group with small size (e.g., 64), the eigen decomposition is not the bottleneck of computation, compared to convolution operation. This make our orthogonal linear module possible to be applied in very deep and high dimensional CNNs.

\begin{figure*}[t]
\centering
  \subfigure[EI+QR]{
  \includegraphics[width=0.23\linewidth]{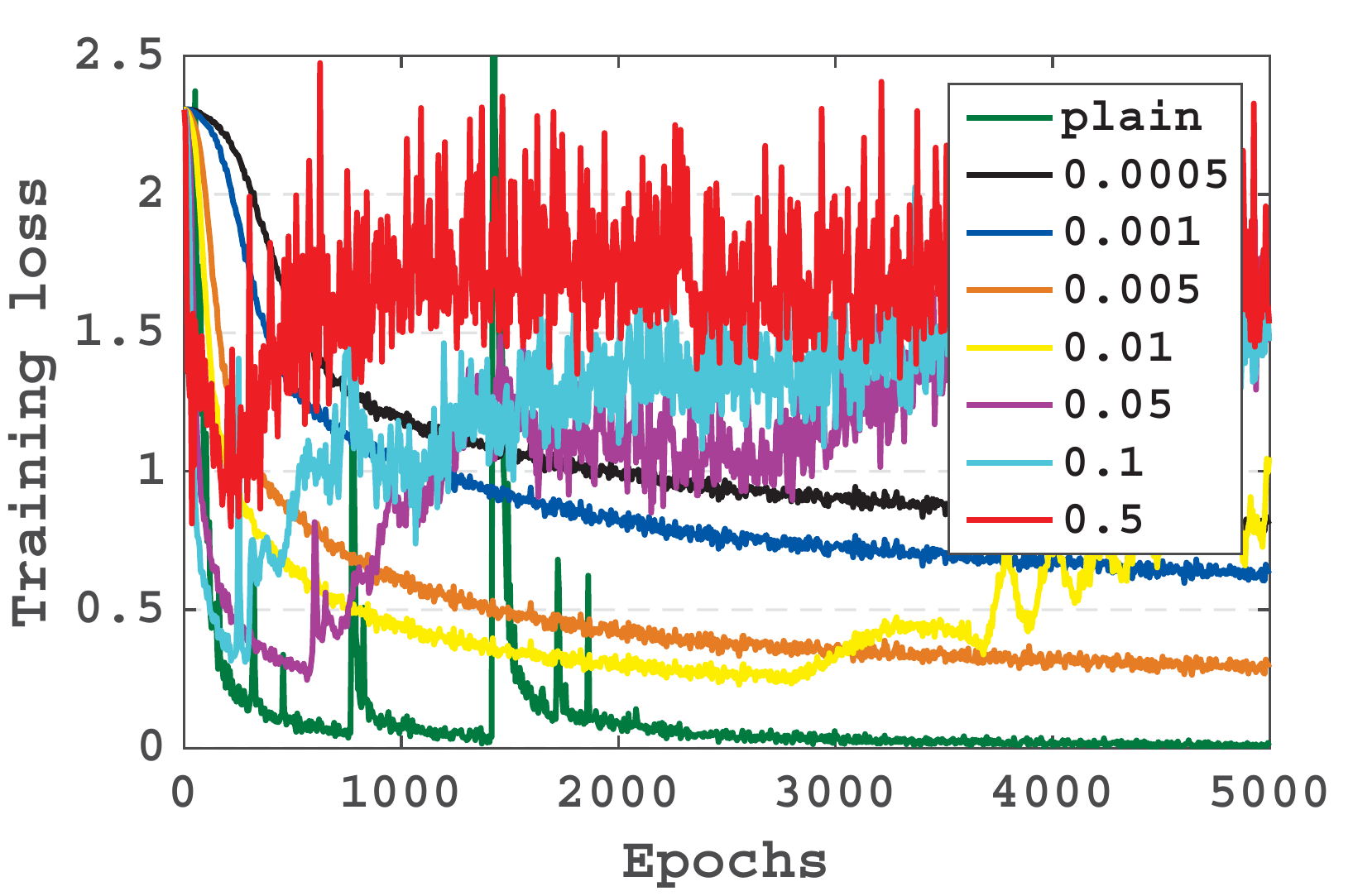}
  }
  \subfigure[CI+QR]{
  \includegraphics[width=0.23\linewidth]{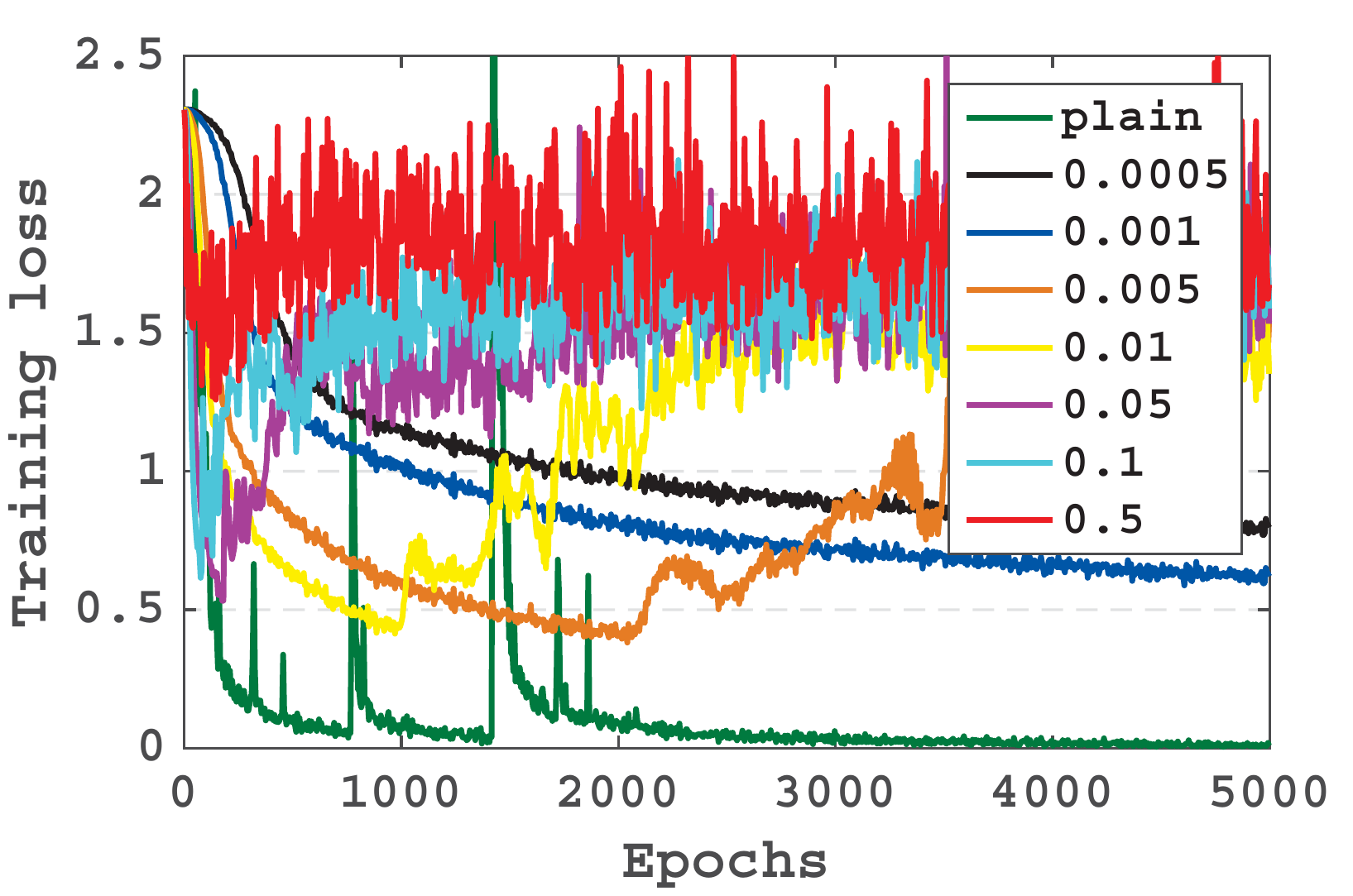}
  }
  \subfigure[CayT]{
  \includegraphics[width=0.23\linewidth]{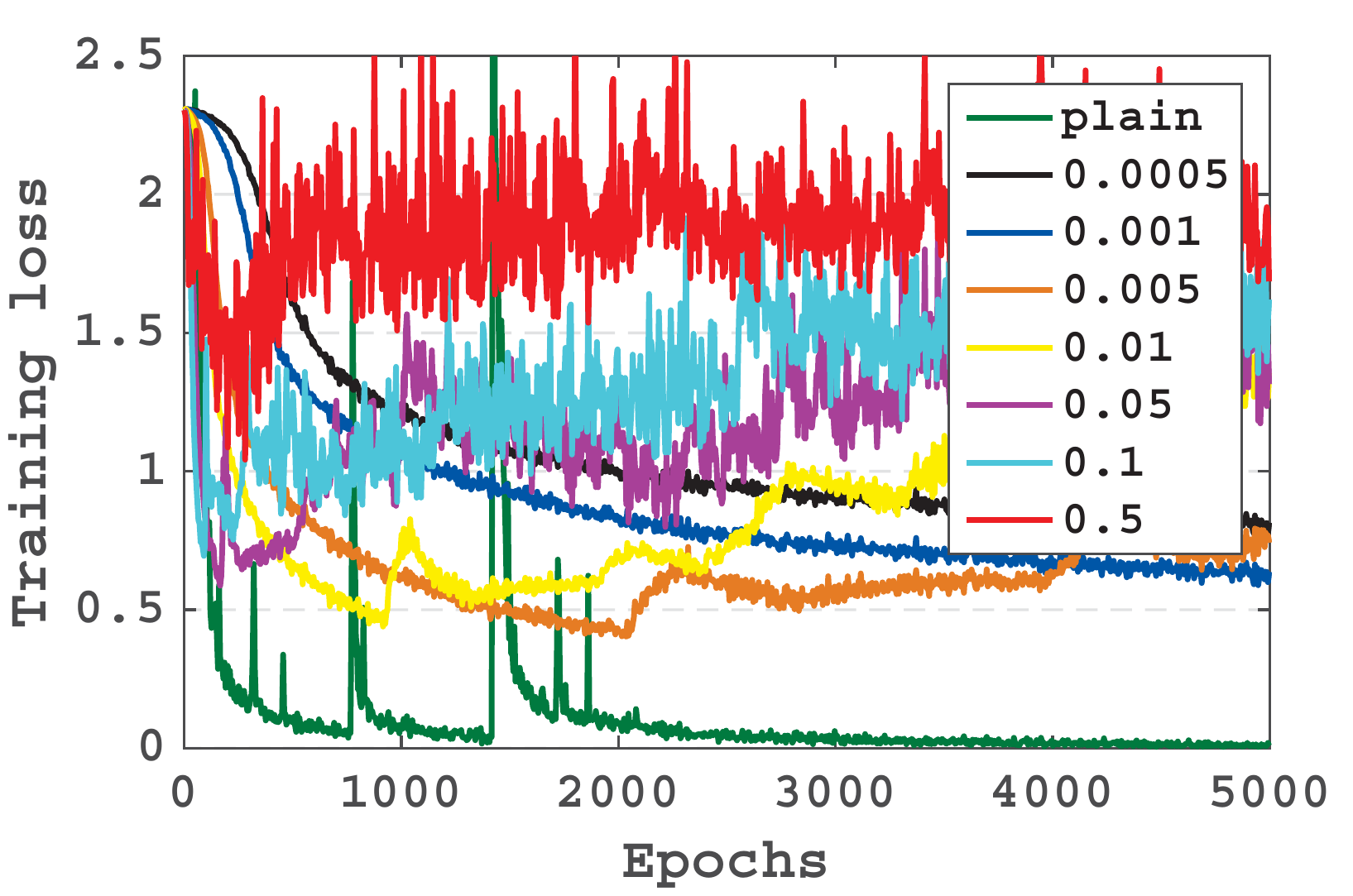}
  }
  \subfigure[Our OLM]{
  \includegraphics[width=0.23\linewidth]{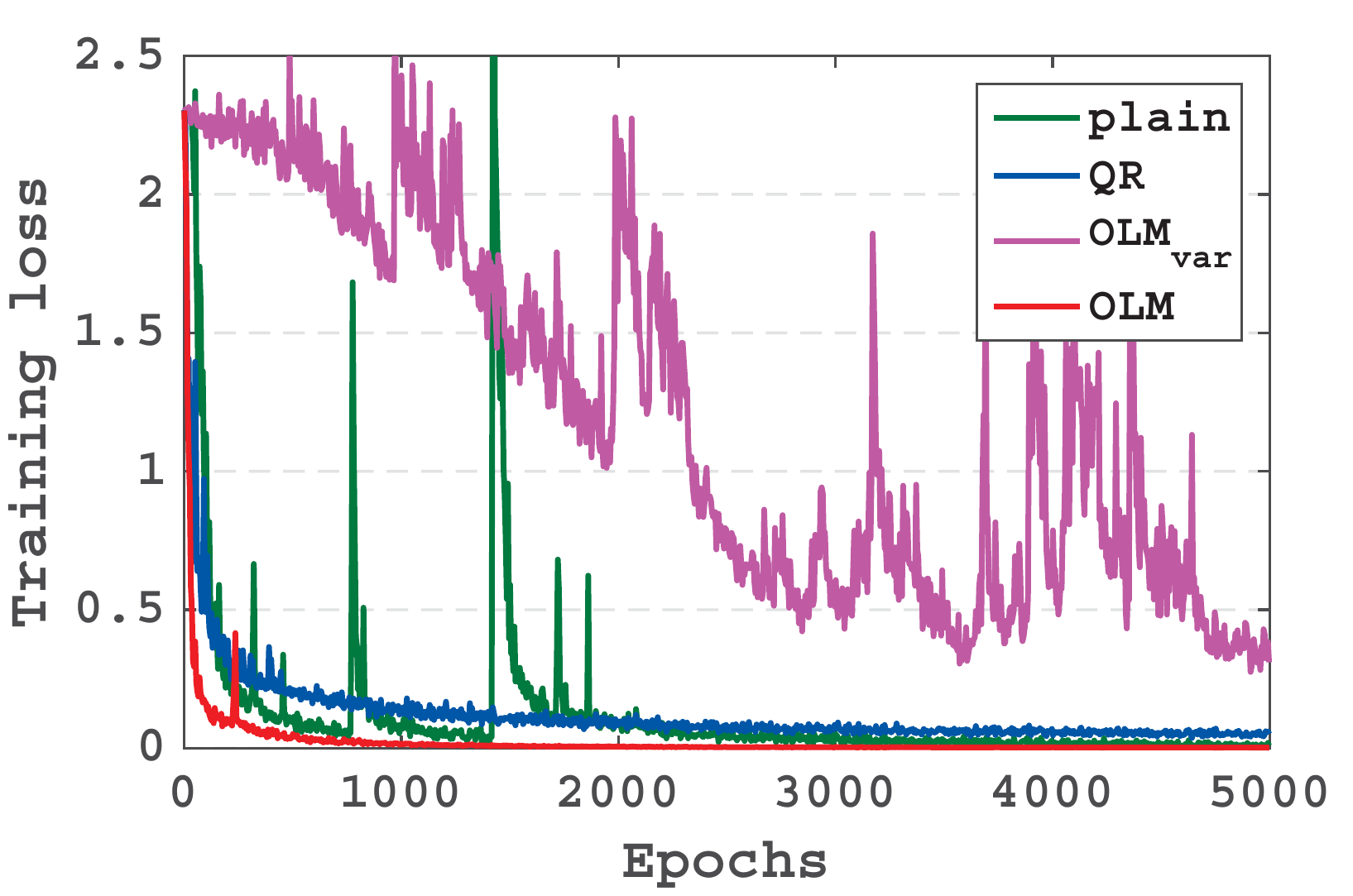}
  }
  \caption{\small Results of Riemannian optimization  methods to solve OMDSM on MNIST dataset under the 4-layer MLP. We  show the training loss curves for different learning rate of `EI+QR', `CI+QR' and  `CayT' compared to the baseline `plain'  in (a), (b) and (c) respectively. We compare our methods to baselines  and report the best performance among all learning rates based on the training loss for each method in (d).}
  \label{fig:exp_MLP_1}
\end{figure*}

\subsubsection{Computational Complexity}
 We show our method is scalable from complexity analysis here and provide empirical results later for large CNNs. Given a convolutional layer with filters $\mathbf{W} \in \mathbb{R}^{n \times d \times F_h \times F_w}$, and $m$ mini-batch data $\{\mathbf{x}_i \in \mathbb{R}^{d \times h \times w}\}_{i=1}^m$. The computational complexity of our method with group size $N_G$ is $O(n N_G d F_h F_w + n N_G^2 +nmdhwF_h F_w)$ per iteration, and if we control a small group size $N_G \ll mhw$, it will be close to the standard convolutional layer as $O(n m d h w F_h F_w)$.

\section{Experiments}
\label{s:exp}
In this section, we first conduct comprehensive experiments to explore different methods to solve the OMDSM problem, and show the advantages of our proposed orthogonal weight normalization solution in terms of the stability and efficiency in optimization.
We then evaluate the effectiveness of the proposed method that learns orthogonal weight matrix in DNNs, by simply replacing our OLM with standard ones on MLPs and CNNs.
Codes to reproduce our results are available from: \textcolor[rgb]{0.00,0.50,1.00}{https://github.com/huangleiBuaa/OthogonalWN}.

\subsection{Comparing Methods for Solving OMDSM}
\label{exp:OMDSM_solve}
In this section,  we use 3 widely used Riemannian optimization methods for solving OMDSM and compared two other  baselines. For completeness, we provide a brief review for Riemannian optimization shown in Appendix \ref{Riem_p3} and for more details please refer to \cite{2008_Book_Absil} and references therein.

 We design comprehensive experiments on MNIST dataset to compare methods for solving OMDSM. The compared methods including:
(1) `EI+QR': using Riemannian gradient with Euclidean inner product  and  QR-retraction \cite{2017_Corr_Harandi};
(2) `CI+QR': using Riemannian gradient  with canonical inner product and QR-retraction;
(3) `CayT': using the Cayley transformation~\cite{2016_NIPS_Wisdom,2017_ICML_Eugene};
(4) `QR': a conventional method that runs  the ordinary gradient descent based on gradient $\frac{\partial F}{\partial \mathbf{W}}$ and projects the solution back to the manifold $\mathbb{M}$ by QR decomposition;
(5) `$OLM_{var}$': using orthogonal transformation: $\mathbf{P}_{var}=\Lambda^{-1/2} \mathbf{D}^T$;
(6) `OLM': our proposed orthogonal transformation by minimizing distortions: $\mathbf{P}^*=\mathbf{D} \Lambda^{-1/2} \mathbf{D}^T$.
 The baseline is the standard network without any orthogonal constraints referred to as `plain'.

We use MLP architecture with 4 hidden layers. The number of neurons in each hidden layer is 100. We train the model with stochastic gradient descent and mini-batch size of 1024. We tried a broadly learning rate  in ranges of $\{0.0005, 0.001, 0.005, 0.01, 0.05,0.1,0.5,1,5\}$.

 We firstly explored the performance  of Riemannian optimization methods for solving OMDSM problem. Figure \ref{fig:exp_MLP_1} (a), (b) and (c) show the training loss curves for different learning rate of `EI+QR', `CI+QR' and  `CayT' respectively, compared to the baseline `plain'. From Figure \ref{fig:exp_MLP_1}, we can find that under larger learning rate (e.g., larger than 0.05) these Riemannian optimization methods suffer severe instability and divergence, even though they show good performance in the initial iterations. They  can also obtain stable optimization behaviours under small learning rate but are significantly slower in convergence than the baseline `plain' and suffer worse performance.

 We then compared our proposed method with the baseline `plain' and the conventional method `QR',  and report the best performance among all learning rates based on the training loss for each method in Figure \ref{fig:exp_MLP_1} (d).
We can find that the conventional method `QR' performs stably. However, it also suffers inferior performance of final training loss compared to `plain'.
 The proposed `\emph{OLM}' works stably and converges the fastest.  Besides, we find that `$OLM_{var}$'  suffered instability, which means that minimizing distortions formulated by Eqn.~\ref{eqn:orthogonalTransform}   is essential  to ensure the stability of solving OMDSM.

We also explore 6-layer and 8-layer MLPs and further with  mini-batch size of 512 and 256. We observe the similar phenomena shown in Appendix \ref{exp_p3}.  Especially with the number of layer increasing, `\emph{OLM}' shows more advantages compared to  other methods.
 These comprehensive experiments strongly support our empirical conclusions that: (1) Riemannian optimization methods probably do not work for the OMDSM problem, and if work, they must be under fine designed algorithms or tuned hyper-parameters; (2) deep feed-forward neural networks (e.g., MLP in this experiment) equipped with orthogonal weight matrix is easier for optimization by our `OLM' solution.

  \begin{figure}[t]
\centering
  \subfigure[training error]{
  \includegraphics[width=0.44\linewidth]{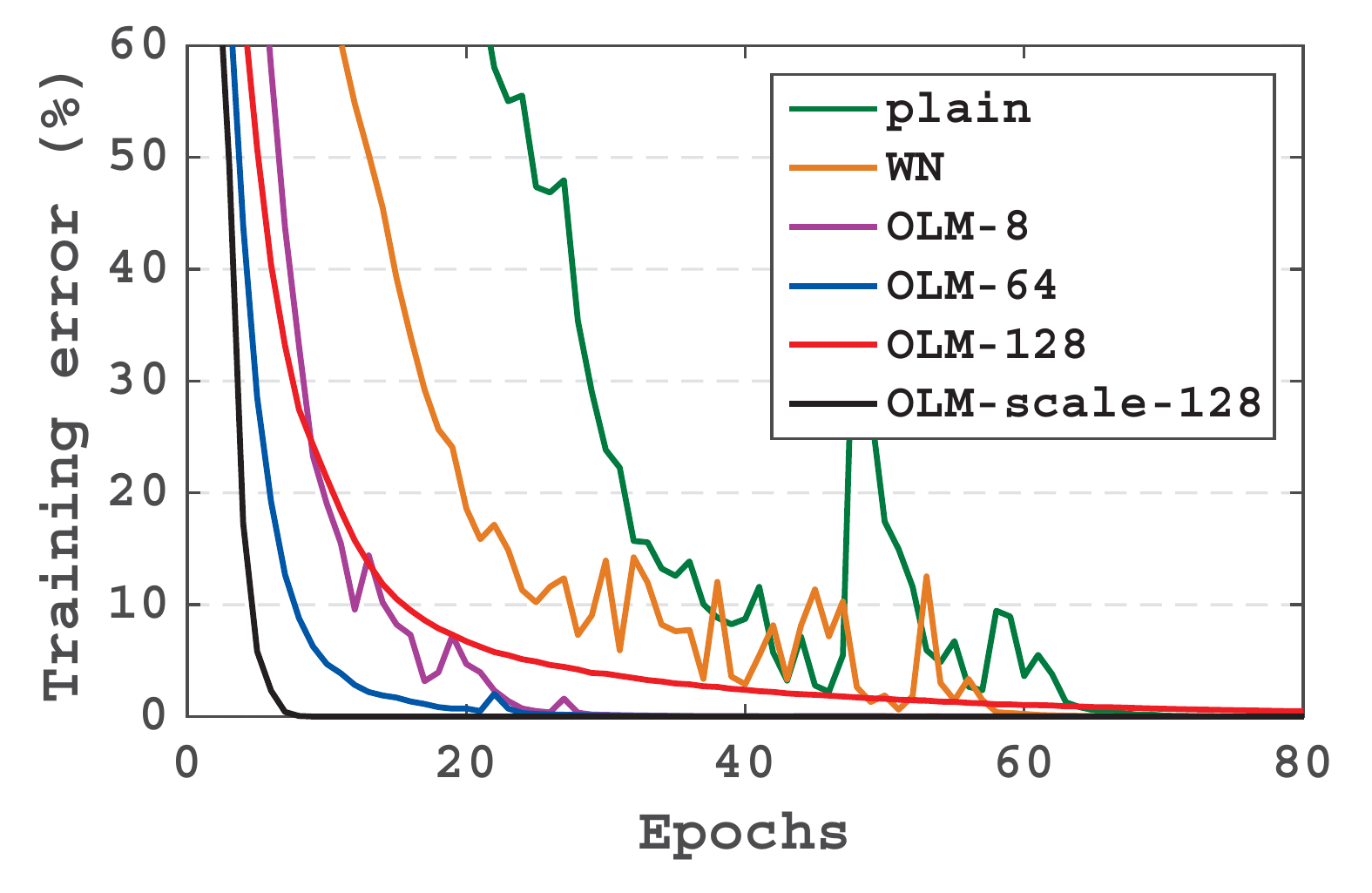}
  }
  \subfigure[test error]{
  \includegraphics[width=0.44\linewidth]{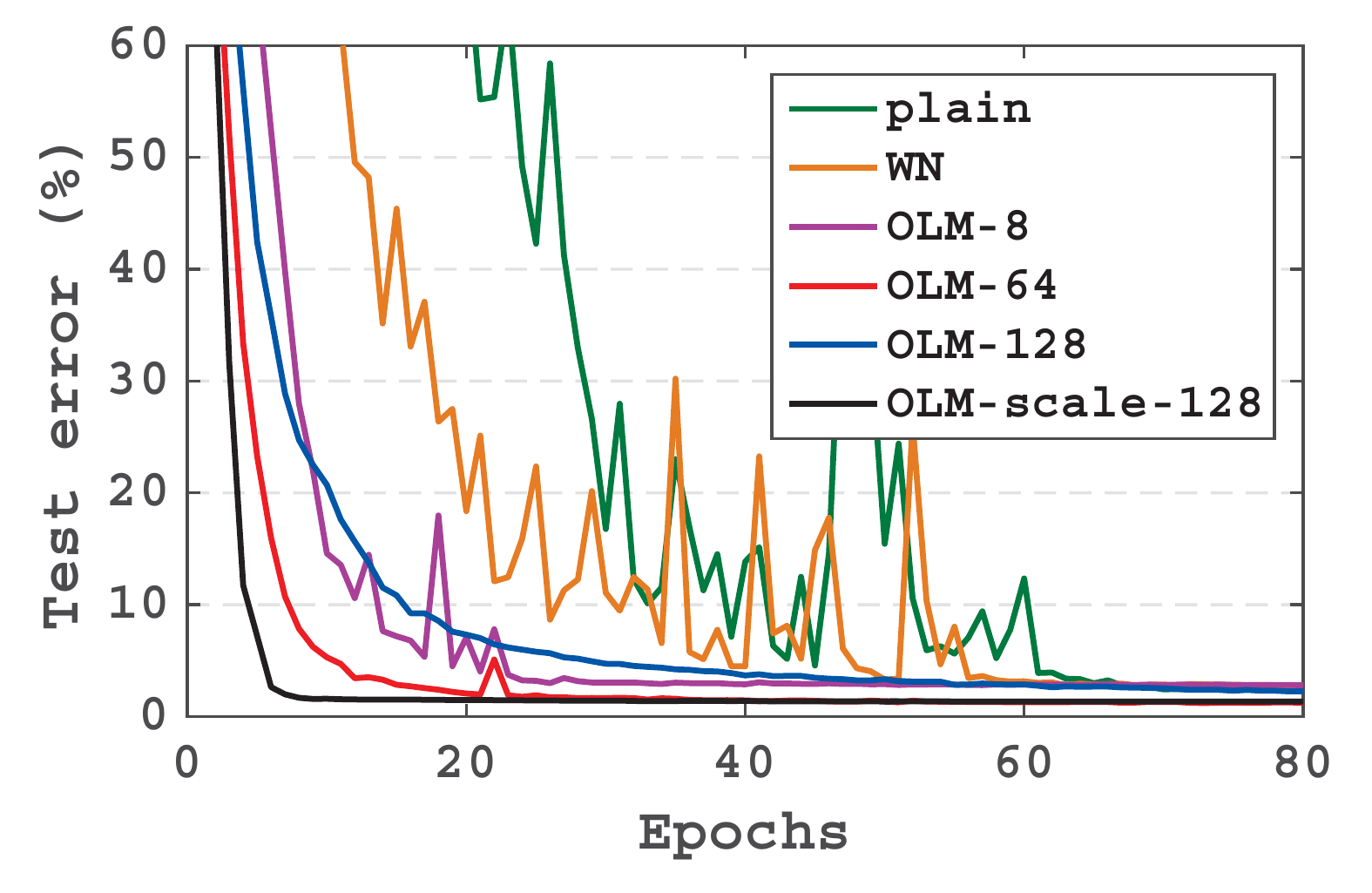}
  }
  \caption{\small Performance comparisons in MLP architecture on PIE dataset. We compare the effect of different group size of OLM. }
  \label{fig:exp_MLP_2}
\end{figure}

  \begin{figure}[t]
\centering
  \subfigure[batch normalization]{
  \includegraphics[width=0.44\linewidth]{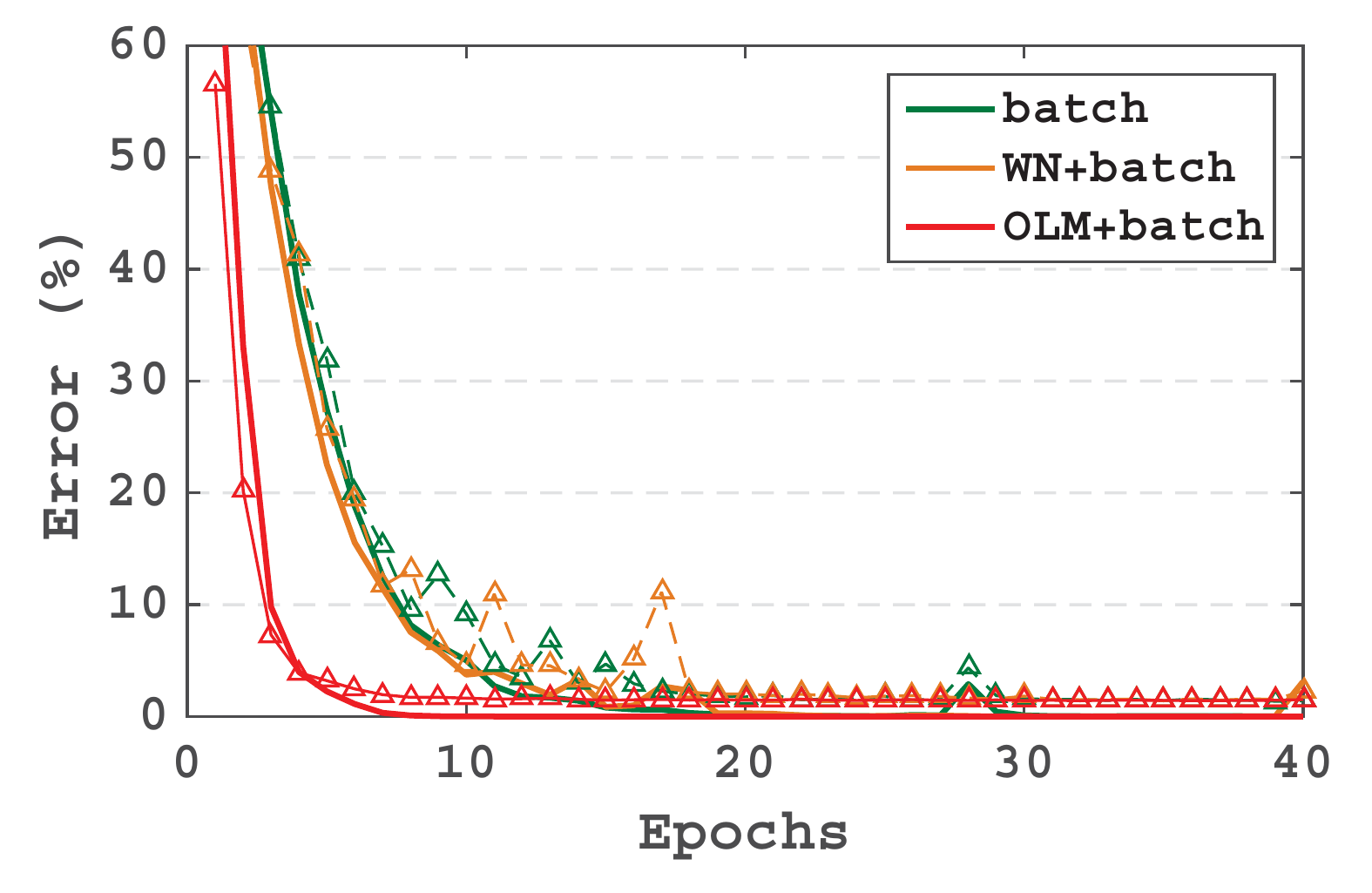}
  }
  \subfigure[Adam optimization]{
  \includegraphics[width=0.44\linewidth]{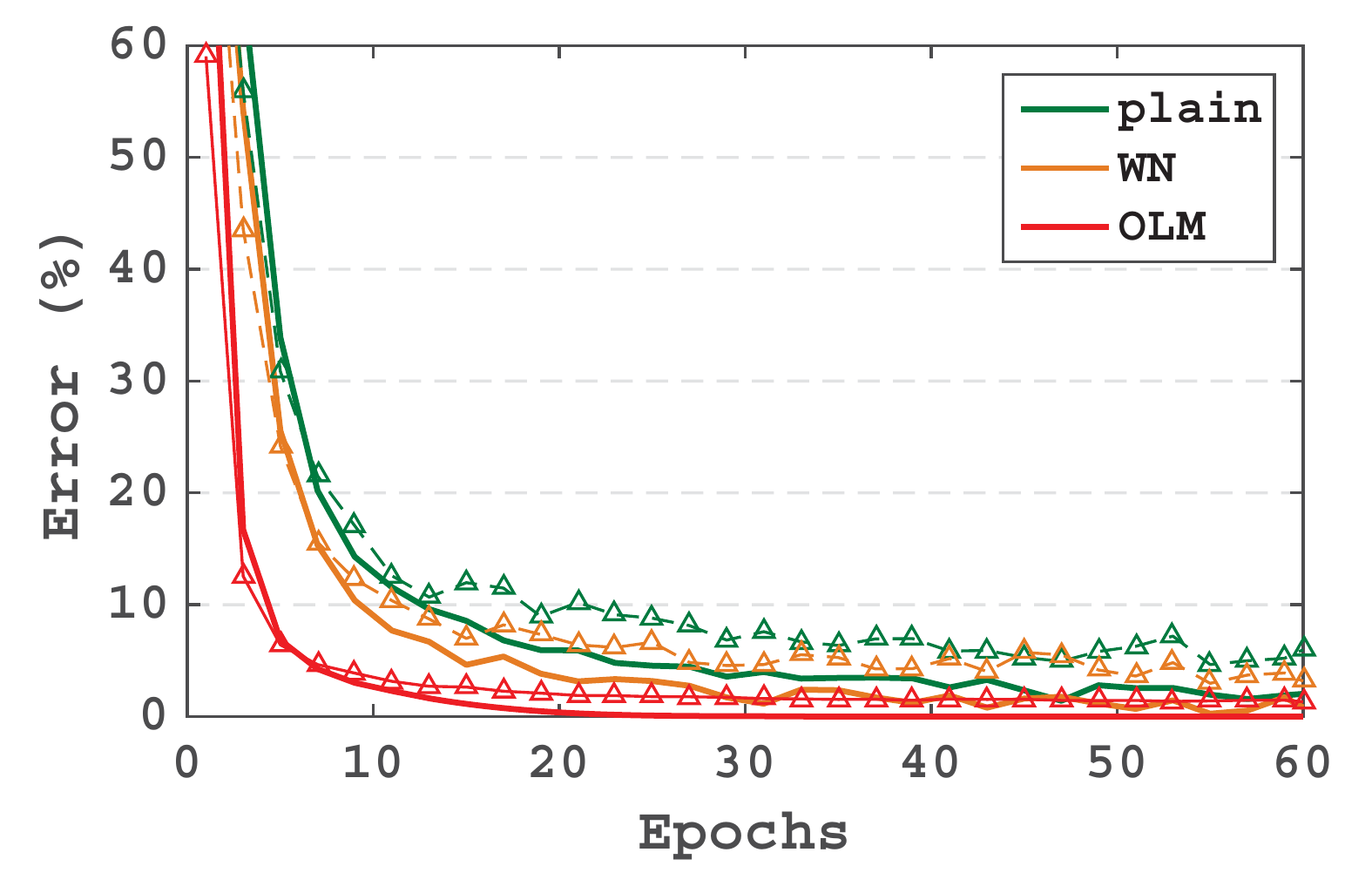}
  }
  \caption{\small Performance comparisons in MLP architecture on PIE dataset by combining (a) batch normalization; (b) Adam optimization. We evaluate the training error (solid lines) and test error (dash lines marked with triangle). }
  \label{fig:exp_MLP_3}
\end{figure}

\subsection{MLP Architecture}

 Now we investigate the performance of OLM in MLP architecture. On PIE face recognition dataset with 11,554 images from 68 classes, we sample 1,340 images as the test set and others as training set.
 Here, we employ standard networks (referred as \emph{plain}) and networks with Weight Normalization (\emph{WN}) ~\cite{2016_CoRR_Salimans} as baselines for comparisons.  \emph{WN} is one of the most related study that normalizes the weights as unit norm via re-parameterization as OLM does, but it does not introduce the orthogonality for the weights matrix.
 For all methods, we train a 6-layers MLP with the number of neurons in each hidden layer as 128,128,128,128,128, and Relu as nonlinearity. The mini-batch size is set to 256. We evaluate the training error and test error as a function with respect to epochs.

\subsubsection{Using Different Group Sizes} We explore the effects of group size $N_G$ on the performance when applying OLM. In this setup, we employ stochastic gradient descent (SGD) optimization and the learning rates are selected based on the validation set ($10\%$ samples of the training set) from $\{0.05, 0.1, 0.2, 0.5, 1\}$. Figure \ref{fig:exp_MLP_2}  shows the performance of OLM using different $N_G$ (`OLM-$N_G$'), compared with \emph{plain} and \emph{WN} methods. We can find that OLM achieves significantly better performance in all cases, which means introducing orthogonality to weight matrix can largely improve the network performance. Another observation is that though increasing group size would help improve orthogonalization, too large group size will reduce the performance. This is mainly because a large $N_G=128$ provides overmuch regularization. Fortunately, when we add extra learnable scale (indicated by `OLM-scale-128') to recover the model capacity as described in previous section, it can help to achieve the best performance.

\subsubsection{Combining with Batch Normalization} Batch normalization~\cite{2015_ICML_Ioffe} has been shown to be helpful for training the deep architectures~\cite{2015_ICML_Ioffe,2016_CoRR_He}. Here, we show that OLM enjoys good compatibility to incorporate well with batch normalization, which still outperforms others in this case. Figure \ref{fig:exp_MLP_3} (a) shows the results of training/test error with respect to epochs. We can see that \emph{WN} with batch normalization (`WN+batch') has no advantages compared with the standard network with batch normalization (`batch'), while `OLM+batch' consistently achieves the best performance.

\subsubsection{Applying Adam Optimization} We also try different optimization technique such as Adam~\cite{2014_CoRR_Kingma} optimization. The hyper-parameters are selected from learning rates in $\{ 0.001, 0.002, 0.005, 0.01 \}$. We show error rates based on Adam optimization in Figure \ref{fig:exp_MLP_3} (b). From the figure, we can see OLM also obtains the best performance.

\subsection{CNN Architectures}
\label{sec:CNN_debug}

\begin{figure}[tbp]
\centering
  \subfigure[CIFAR-10]{
  \includegraphics[width=0.44\linewidth]{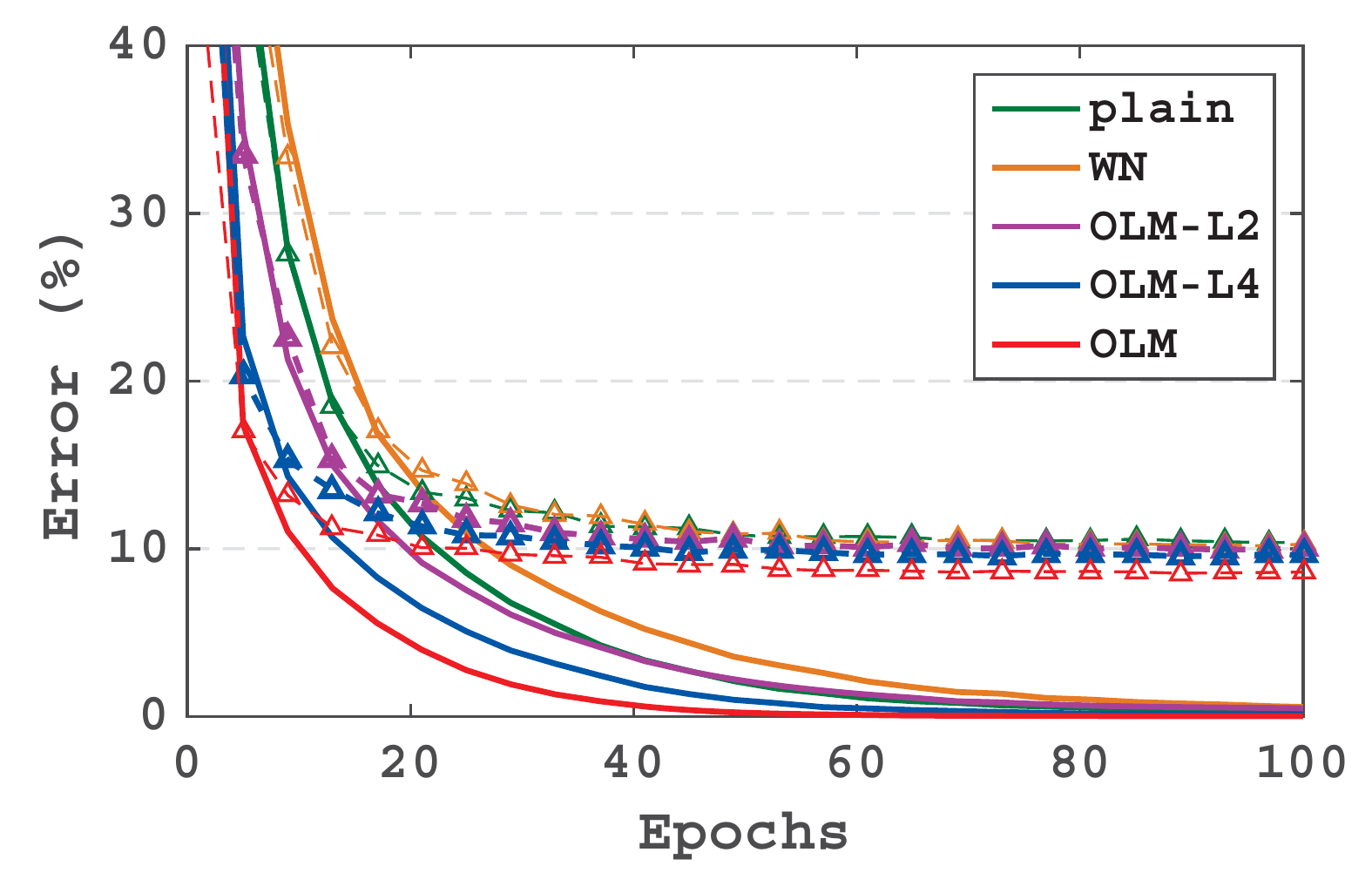}
  }
   \subfigure[CIFAR-100]{
  \includegraphics[width=0.44\linewidth]{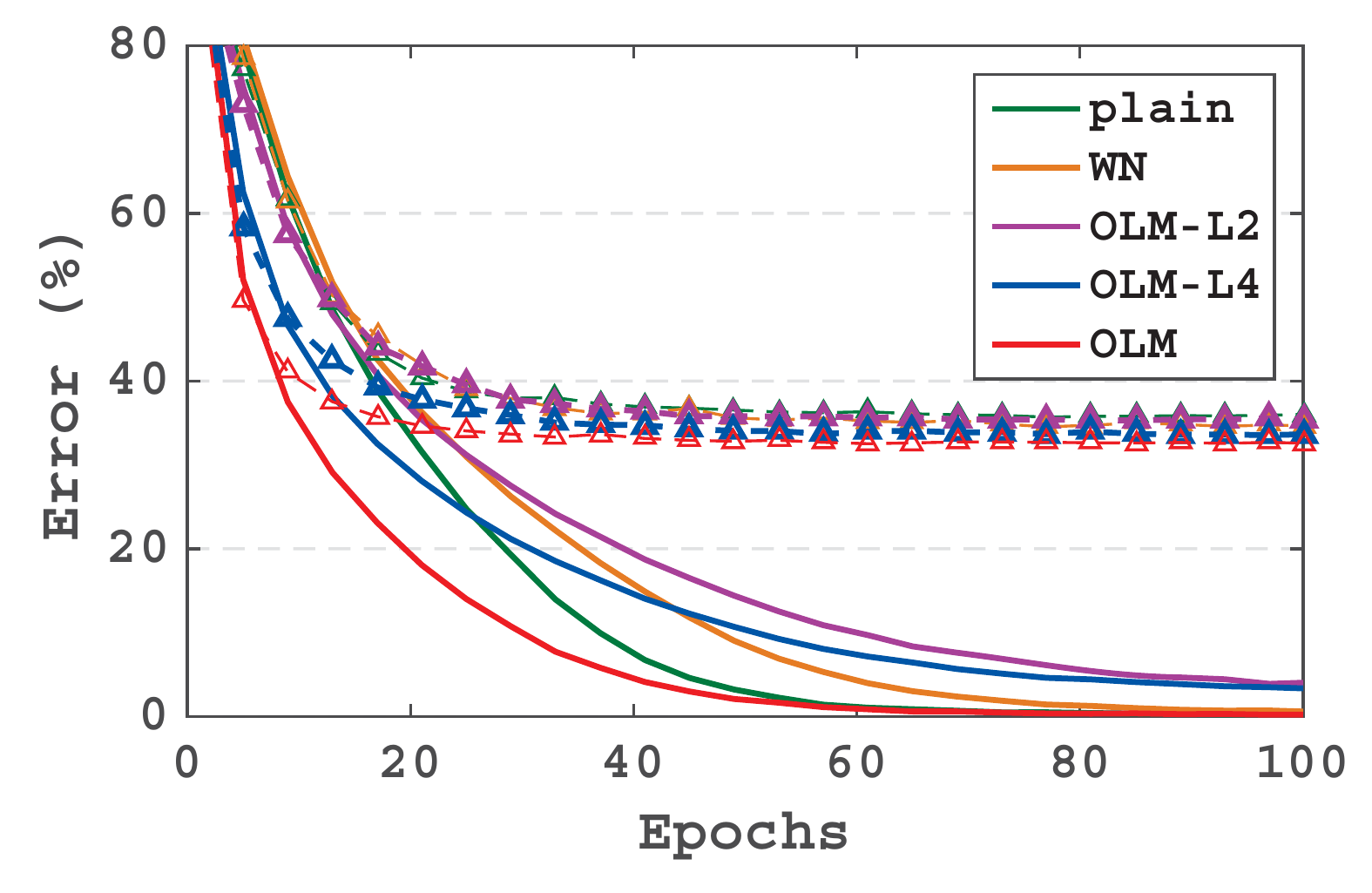}
  }
  \caption{\small Experimental results on VGG-style architectures over CIFAR datasets. We evaluate the training error (solid lines) and test error (dash lines marked with triangle) with respect to epochs, and all results are averaged over 5 runs.}
  \label{fig:exp_Conv}
\end{figure}

\begin{figure}[tbp]
\centering
  \subfigure[CIFAR-10]{
  \includegraphics[width=0.44\linewidth]{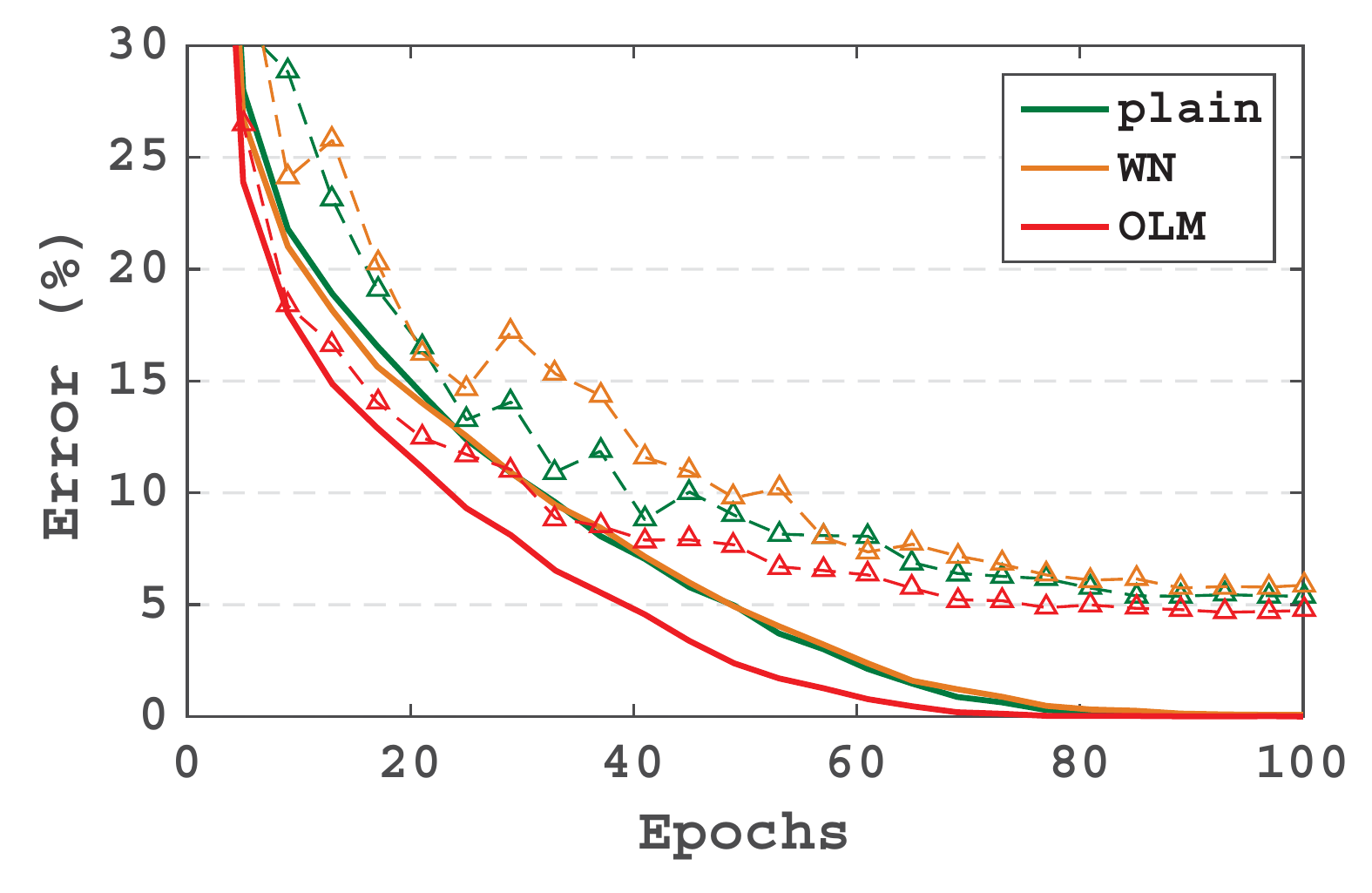}
  }
  \subfigure[CIFAR-100]{
  \includegraphics[width=0.44\linewidth]{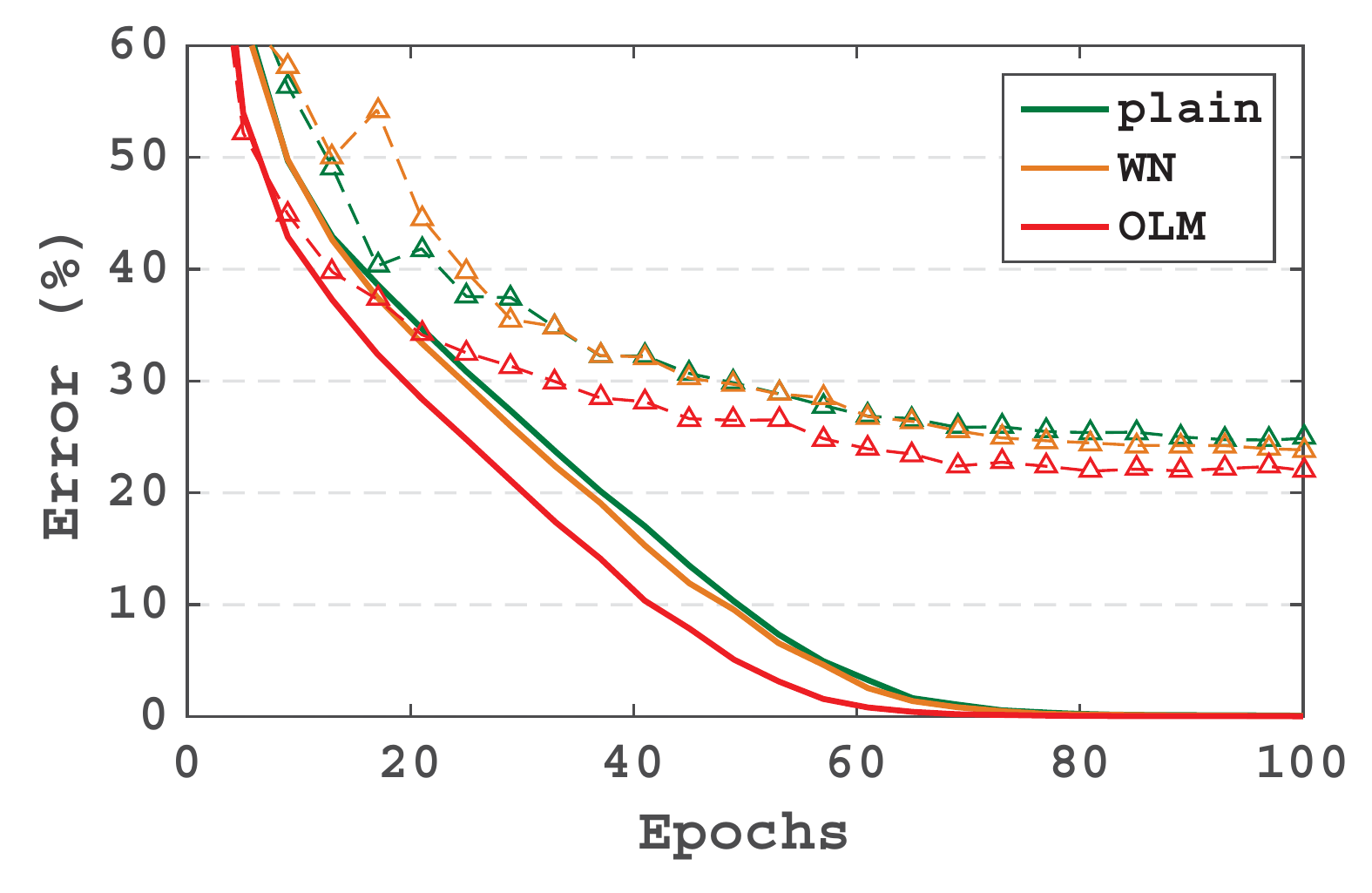}
  }
  \caption{\small Experimental results on BN-Inception over CIFAR datasets. We evaluate the training error (solid lines) and test error (dash lines marked with triangle) with respect to epochs, and all results are averaged over 5 runs.}
  \label{fig:exp_Conv_BNInception}
\end{figure}

In this section, We evaluate our method on a VGG-style CNN~\cite{2014_CoRR_Simonyan}, BN-Inception ~\cite{2014_CoRR_Szegedy,2015_ICML_Ioffe}, and Wide Residual Networks ~\cite{2016_CoRR_Zagoruyko}  for image classification, respectively on CIFAR-10 and CIFAR-100 ~\cite{2009_TR_Alex}. For each dataset, We use the official training set of 50k images  and the standard test set of 10k images. The data preprocessing and data argumentation follow the commonly used mean$\&$std normalization and flip translation as described in~\cite{2015_CVPR_He}. For OLM method, we replace all convolution layers with our OLM modules by default on CNNs, if we do not specify it.  Among all experiments, the group size $N_G$ of OLM is set as 64.
\subsubsection{VGG-style network}
 We adopt the $3 \times 3$ convolutional layer as the following specification: $\rightarrow conv(64) \rightarrow conv(128) \rightarrow maxPool(2,2,2,2) \rightarrow conv(256) \rightarrow conv(256) \rightarrow maxPool(2,2,2,2) \rightarrow conv(512)\rightarrow  conv(512) \\ \rightarrow AvePool(8,8,1,1)\rightarrow fc(512 \times ClassNum)$. SGD is used as our optimization method with mini-batch size of 256. The best initial learning rate is chosen from $\{0.01, 0.05, 0.1\}$ over the validation set of 5k examples from the training set, and exponentially decayed to 1\% in the last (100th) epoch. We set the momentum to 0.9 and weight decay to $5\times 10^{-4}$.
 Table \ref{table:VGG-error} reports the test error, from which we can find OLM achieves the best performance consistently on both datasets.
 Figure \ref{fig:exp_Conv} (a) and (b) show the training and test errors with respect to epochs on CIFAR-10 and CIFAR-100, respectively. On CIFAR-100, to achieve the final test error  of \emph{plain} as 36.02 $\%$, OLM takes only 17 epochs. Similarly, on CIFAR-10, OLM only takes 21 epochs to achieve the final test error of \emph{plain} as 10.39 $\%$. While on both datasets, 'plain' takes about 100 epochs.
 Results demonstrate that OLM converges significantly faster in terms of training epochs and achieves better error rate compared to baselines.

We  also study the effect of OLM on different layers. We optionally replace the first 2 and 4 convolution layers with OLM modules (referred as OLM-L2 and OLM-L4 respectively). From Figure \ref{fig:exp_Conv} and Table \ref{table:VGG-error}, we can find that with the numbers of used OLM increasing, the VGG-style network achieves better performance both in optimization efficiency and generalization.

\subsubsection{BN-Inception}
For BN-inception network, batch normalization ~\cite{2015_ICML_Ioffe} is inserted after each linear layer based on original Inception architecture ~\cite{2014_CoRR_Szegedy}. Again, we train the network using SGD, with the momentum 0.9, weight decay $5\times 10^{-4}$ and the batch size 64. The initial learning rate is set to 0.1 and decays exponentially every two epochs until the end of 100 epoches with 0.001.
Table \ref{table:BN-error} reports the test error after training and  Figure \ref{fig:exp_Conv} (c) and (d) show the training/test error with respect to epochs on CIFAR-10 and CIFAR-100, respectively.
 We can find that OLM converges faster in terms of training epochs and achieve better optimum, compared to baselines, which indicate consistent conclusions for VGG-style network above.

\begin{table}[t]
\caption{Test error ($\%$) on VGG-style over CIFAR datasets. We report the `mean $\pm std$' computed over 5 independent runs.}
\label{table:VGG-error}
\begin{center}
\begin{small}
\begin{tabular}{lcr}
\hline
   & CIFAR-10  & CIFAR-100  \\
\hline
plain    & 10.39 $\pm$ 0.14 & 36.02 $\pm$ 0.40 \\
WN   & 10.29$\pm$ 0.39 & 34.66 $\pm$ 0.75\\
OLM-L2   & 10.06 $\pm$ 0.23&35.42 $\pm$ 0.32\\
 OLM-L4   & 9.61 $\pm$ 0.23& 33.66 $\pm$ 0.11\\
OLM  & \textbf{8.61} $\pm$ 0.18&\textbf{32.58} $\pm$ 0.10\\
\hline
\end{tabular}
\end{small}
\end{center}
\end{table}

\begin{table}[t]
\caption{Test error ($\%$) on BN-Inception over CIFAR datasets. We report the `mean $\pm std$' computed over 5 independent runs.}
\label{table:BN-error}
\begin{center}
\begin{small}
\begin{tabular}{lcr}
\hline
   & CIFAR-10  & CIFAR-100  \\
\hline
plain    & 5.38$\pm$ 0.18  & 24.87 $\pm$ 0.15    \\
WN & 5.87$\pm$ 0.35 & 23.85 $\pm$ 0.28  \\
OLM   & \textbf{4.74$\pm$ 0.16} &\textbf{22.02} $\pm$ 0.13 \\
\hline
\end{tabular}
\end{small}
\end{center}
\end{table}

\begin{table}[t]
\caption{Test errors ($\%$) of different methods on CIFAR-10 and CIFAR-100. For OLM, we report the `mean $\pm std$' computed over 5 independent runs.
`WRN-28-10*' indicates the new results given by authors on their Github.
}
\label{table:error}
\centering
\begin{small}
\begin{tabular}{l|c|c}
    \toprule
  & CIFAR-10 &CIFAR-100  \\
\hline
pre-Resnet-1001 & 4.62& 22.71\\
WRN-28-10   & 4.17& 20.04\\
WRN-28-10*   & 3.89 & 18.85\\
WRN-28-10-OLM (ours)  & \textbf{3.73}  $\pm$ 0.12 & 18.76 $\pm$ 0.40 \\
WRN-28-10-OLM-L1 (ours) & 3.82 $\pm$ 0.19 & \textbf{18.61} $\pm$ 0.14 \\
\bottomrule
\end{tabular}
\end{small}
\end{table}

\begin{table}[t]
\caption{Top-5 test error ($\%$, single model and single-crop) on ImageNet dataset.}
\label{tab:imageNet}
\begin{center}
\begin{small}
\begin{tabular}{lccccr}
\hline
   & AlexNet & BN-Inception & ResNet & Pre-ResNet   \\
\hline
plain   & 20.91    &    12.5    &    9.84 &   9.79 \\
OLM     & \textbf{20.43}  &  \textbf{9.83} & \textbf{9.68} &  \textbf{9.45} \\
\hline
\end{tabular}
\end{small}
\end{center}
\end{table}

\subsubsection{Wide Residual Netwok}
\label{sec:exp_wideResNet}
 Wide Residual Network (WRN) has been reported to achieve  state-of-the-art results on CIFARs \cite{2016_CoRR_Zagoruyko}. We adopt WRN architecture with depth 28 and width 10 and the same experimental setting as in~\cite{2016_CoRR_Zagoruyko}.
Instead of ZCA whitening, we preprocess the data using per-pixel mean subtract and standard variance divided as described in ~\cite{2015_CVPR_He}.
We implement two setups  of OLM: (1) replace all the convolutional layers by WRN (\emph{WRN-OLM}); (2) only replace the first convolutional layer in WRN (\emph{WRN-OLM-L1}). Table \ref{table:error} reports the test errors.
 We can see that OLM can further improve the state-of-the-art results achieved by WRN. For example, on CIFAR-100, our method \emph{WRN-OLM} achieves the best 18.61 test error, compared to 20.04 of WRN reported in~\cite{2016_CoRR_Zagoruyko}. Another interesting observation is that \emph{WRN-OLM-L1} obtains the best performance on CIFAR-10 with test error as $3.73 \%$, compare to $4.17\%$ of WRN, which means that we can improve residual networks by only constraining the first convolution layer orthogonal and the extra computation cost is negligible.

\subsubsection{Computation Cost}
We also evaluate computational cost per iteration in our current Torch-based implementation, where the convolution relies on the fastest \emph{cudnn} package. In the small VGG-style architecture with batch size of 256, OLM costs 0.46s, while \emph{plain} and \emph{WN} cost 0.26s and 0.38s, respectively.
On large WRN network, OLM costs 3.12s compared to 1.1s of \emph{plain}.
Note that, our current implementation of OLM can be further optimized.

\subsection{Large Scale Classification on ImageNet Challenge}
To further validate the effectiveness of OLM on large-scale dataset, we employ ImageNet 2012 consisting of more than 1.2M images from 1,000 classes~\cite{2015_ImageNet}. We use the given 1.28M labeled images for training and the validation set with 50k images for testing. We evaluate the classification performance based on  top-5 error.
We apply the well-known AlexNet ~\cite{2012_NIPS_Krizhevsky} with batch normalization inserted after the convolution layers, BN-Inception\footnote{We use the public Torch implementation of AlexNet and BN-Inception available on: https://github.com/soumith/imagenet-multiGPU.torch}, ResNet ~\cite{2015_CVPR_He} with 34 layers  and its advanced version Pre-ResNet ~\cite{2016_CoRR_He}\footnote{We use the public Torch implementation available on: https://github.com/facebook/fb.resnet.torch}. In AlexNet and BN-Inception, we replace all the convolution layers with OLM modules for our method,  and in ResNet and Pre-ResNet, we only replace the first convolution layer with OLM module, which is shown effective with negligible computation cost based on previous experiment.

We run our experiments on one GPU.
To guarantee a fair comparison between our method with the baseline, we keep all the experiments settings the same as the publicly available Torch implementation (e.g., we apply stochastic gradient descent with  momentum of 0.9, weight decay of 0.0001, and set the initial learning rate to 0.1). The exception is that we use mini-batch size of 64 and 50 training epochs considering the GPU memory limitations and training  time costs. Regarding learning rate annealing, we use exponential decay to 0.001, which has slightly better performance than the method of lowering by a factor of 10 after epoch 20 and epoch 40 for each method.
 The final test errors are shown in Table \ref{tab:imageNet}. We can find that our proposed OLM method obtains better performance compared to the baselines over  AlexNet, BN-Inception, ResNet and Pre-ResNet architectures.

\section{Related Work and Discussion}

In optimization community, there exist methods to solve the optimization problem over matrix manifolds with orthogonal constraints ~\cite{2011_Yale_Tagarre,2012_SIAM_Absil,2013_MP_Wen,2008_Book_Absil,2015_JMLR_Zoubin}.
They are usually limited for one manifold (one linear mapping) ~\cite{2015_JMLR_Zoubin}, and are mainly based on full batch gradient. When applying to DNNs, it suffered instability in convergence or inferior performance, as observed by either ~\cite{2017_Corr_Harandi,2017_ICML_Eugene} or our experiments.

In deep learning community, there exist researches using \emph{orthogonal matrix} ~\cite{2016_ICML_Arjovsky,2016_NIPS_Wisdom,2016_CoRR_Dorobantu,2017_ICML_Eugene} or normalization techniques~\cite{2015_ICML_Ioffe,2016_CoRR_Salimans,YuHLSC17} to avoid the gradient vanish and explosion problem.
Arjovsky \emph{et al.}~\cite{2016_ICML_Arjovsky} introduced the \emph{orthogonal matrix }for the hidden to hidden transformation in RNN. They constructed an expressive unitary weight matrix by composing several structured matrices that act as building blocks with parameters to be learned.
Wisdom \emph{et al.}~\cite{2016_NIPS_Wisdom} pointed out that Arjovsky's method ~\cite{2016_ICML_Arjovsky} has restricted representational capacity and proposed a method of optimizing a full-capacity unitary matrix by using Reimannian gradient in canonical inner product with Cayley transformation ~\cite{2016_NIPS_Wisdom,2017_ICML_Eugene}.
Dorobantu \emph{et al.}~\cite{2016_CoRR_Dorobantu} also proposed a simple method of updating orthogonal linear transformations in RNN in a way that maintains orthogonality. Vorontsov~\cite{2017_ICML_Eugene} advocate soft constraints on orthogonality when they find that hard constraints on orthogonality can negatively affect the speed of convergence and model performance.
However, these methods are limited for the hidden to hidden transformation in RNN because both methods require the weight matrix to be square matrix. Our methods are more general and can adapt to situations where the weight matrix are not square, especially for deep Convolutional Neural Networks (CNNs).

Recently, Harandi and Fernando ~\cite{2017_Corr_Harandi} have proposed Stiefel layer to guarantee fully connected layer to be orthogonal by using Reimannian gradient in Euclidean inner product with QR-retraction. However, this method only shows the experimental results of replacing the last one or two layers  of neural networks with Stiefel layers. Besides, this method is not stable as shown in their experiments: when they constrain the last two fully connected layer to be orthogonal in VGG networks, the accuracy performance is low. We also observe similar phenomena as shown in Section \ref{exp:OMDSM_solve} that `EI+QR' shows instability in convergence. Ozay and Okatani ~\cite{2016_Corr_Ozay} also used Riemannian optimization to guarantee convolutional kernels within a channel are orthogonal, while our ensure filters (among channels) orthogonal.
Some methods ~\cite{2017_CVPR_Xie,2017_ICLR_guez} used orthogonal regularizer as a penalty posed  on the objective function.
These methods can not learn orthogonal filters, because they relax the constraints and only penalize their violations. The solution of these methods is on infeasible set of the optimization problem with orthogonality constraint  while our on feasible set.

\section{Conclusions and Further Work}
We formulate learning orthogonal linear transformation in DNNs as Optimization over Multiple Dependent Stiefel Manifolds (OMDSM) and propose the Orthogonal Weight Normalization method to solve it, which is stable and can be applied to large and deep networks. Base on this solution, we design Orthogonal Linear Module (OLM) which can be applied as an alternative to standard linear module. We show that neural networks equipped with OLM  can improve optimization efficiency and generalization ability.  In addition, new deep architectures that address domain-specific representation can also benefit from the proposed method by simply replacing standard linear module with OLM.

 Various shallow dimensional reduction methods have been  unified under the optimization framework with orthogonality constraints~\cite{2015_JMLR_Zoubin}. Our method has potentials to improve the performance of corresponding  unsupervised ~\cite{2017_IJCAI_Wang} and semi-supervised methods ~\cite{2015_NIPS_Rasmus} in DNNs. Besides, our method has great potential to be used in  improving the robust of the networks to  adversarial examples \cite{2017_ICML_Cisse}.



\small
\bibliographystyle{plain}
\bibliography{whitening}

\appendix
\section{Appendix}

\subsection{Proof of Theorem 1}
\label{proof_p1}

 \textbf{Theorem 1.} ~\emph{Let $\mathbf{s}= \mathbf{W} \mathbf{x}$, where $\mathbf{W} \mathbf{W}^T = \mathbf{I}$ and $\mathbf{W} \in \mathbb{R}^{n \times d}$.
  (1) Assume the mean of $\mathbf{x}$ is $\mathbb{E}_{\mathbf{x}}[\mathbf{x}]=\mathbf{0}$, and covariance matrix of $\mathbf{x}$ is $cov(\mathbf{x})=\sigma^2 \mathbf{I}$. Then  $\mathbb{E}_{\mathbf{s}}[\mathbf{s}]=\mathbf{0}$, $cov(\mathbf{s})=\sigma^2 \mathbf{I}$.
   (2) If $n=d$, we have $\|\mathbf{s}\|= \|\mathbf{x}\|$.
   (3) Given the back-propagated gradient $\frac{\partial \mathcal{L}}{\partial \mathbf{s}}$, we have  $\| \frac{\partial \mathcal{L}}{\partial \mathbf{x}} \|= \|\frac{\partial \mathcal{L}}{\partial \mathbf{s}}\|.$}

 \proof (1) It's easy to calculate:
\begin{eqnarray}
	&\mathbb{E}_{\mathbf{s}}[\mathbf{s}]&=\mathbf{w}^T \mathbb{E}_{\mathbf{x}} [ \mathbf{x}]=\mathbf{w}^T \mu \mathbf{1}=0
\end{eqnarray}

The covariance of $\mathbf{s}$ is given by
\begin{eqnarray}
cov(\mathbf{s})&=&\mathbb{E}_{\mathbf{s}}[\mathbf{s}-\mathbb{E}_{\mathbf{s}}[\mathbf{s}]]^2   \nonumber \\
          &=&\mathbb{E}_{\mathbf{x}}[\mathbf{W}(\mathbf{x}-\mathbb{E}_{\mathbf{x}}[\mathbf{x}])]^2   \nonumber \\
          &=&\mathbb{E}_{\mathbf{x}}[\mathbf{W} (\mathbf{x}-\mathbb{E}_{\mathbf{x}}[\mathbf{x}])] \cdot \mathbb{E}_{\mathbf{x}}[\mathbf{W}(\mathbf{x}-\mathbb{E}_{\mathbf{x}}[\mathbf{x}])]^T   \nonumber \\
          &=&\mathbf{W} \mathbb{E}_{\mathbf{x}}[(\mathbf{x}-\mathbb{E}_{\mathbf{x}}[\mathbf{x}])] \cdot \mathbb{E}_{\mathbf{x}}[(\mathbf{x}-\mathbb{E}_{\mathbf{x}}[\mathbf{x}])]^T \mathbf{W}^T  \nonumber \\
          &=&\mathbf{W} cov(\mathbf{x}) \mathbf{W}^T  \nonumber \\
          &=&\mathbf{W} \sigma^2 \mathbf{I} \mathbf{W}^T  \nonumber \\
           &=&\sigma^2 \mathbf{W} \mathbf{W}^T = \sigma^2
\end{eqnarray}

(2) If $n=d$, $\mathbf{W}$ is orthogonal matrix, therefore $\mathbf{W}^T \mathbf{W}
= \mathbf{W} \mathbf{W}^T=\mathbf{I}$.
 $\|\mathbf{s}\|=\mathbf{s}^T \mathbf{s}
 = \mathbf{x}^T \mathbf{W}^T \mathbf{W} \mathbf{x}
 =\mathbf{x}^T \mathbf{x}
 =\| \mathbf{x} \| $.

 (3) As similar as the proof of (2), $\| \frac{\partial \mathcal{L}}{\partial \mathbf{x}} \|
 =  \|\frac{\partial \mathcal{L}}{\partial \mathbf{s}} \mathbf{W}\|
 = \frac{\partial \mathcal{L}}{\partial \mathbf{s}} \mathbf{W} \mathbf{W}^T \frac{\partial \mathcal{L}}{\partial \mathbf{s}}^T
  = \frac{\partial \mathcal{L}}{\partial \mathbf{s}} \frac{\partial \mathcal{L}}{\partial \mathbf{s}}^T
 = \|\frac{\partial \mathcal{L}}{\partial \mathbf{s}}\|$

\qed

\subsection{Derivation of Minimizing Orthogonal Vectors Transformation Problem}
\label{proof_p2}
Given a matrix $\mathbf{V} \in \mathbb{R}^{n \times d}$ where $n \leq d$ and $Rank(\mathbf{V})=n$, we expect to transform it by $\mathbf{W} =\mathbf{P} \mathbf{V}$ such that $\mathbf{W} \mathbf{W}^T = I$ where $\mathbf{W} \in \mathbb{R}^{n \times d}$ and $\mathbf{P} \in \mathbb{R}^{n \times n}$. Besides,  we minimize the distortion between the transformed matrix $\mathbf{W}$ and   the original matrix $\mathbf{V}$ in a least square way. This can be formulated as:
  \begin{eqnarray}
\label{eqn:appendex_orth}
	& \min_{\mathbf{P}} tr((\mathbf{W}-\mathbf{V}) (\mathbf{W}-\mathbf{V})^T)  \nonumber \\
   & ~~~~~~ s.t. ~~~~ \mathbf{W}=\mathbf{P} \mathbf{V} ~~and ~~\mathbf{W} \mathbf{W}^T =\mathbf{I}
\end{eqnarray}

where $tr(\mathbf{A})$ denotes the trace of the matrix $\mathbf{A}$. The derivation of this problem is based on paper \cite{2006_TIT_Eldar} where a similar problem is considered in the
context of data whitening. To solve this problem, we first calculate the covariance matrix $\Sigma=\mathbf{V} \mathbf{V}^T$. Since $Rank(\mathbf{V})=n$, we have $\Sigma$ is real positive definite and thus invertible with positive eigenvalues. By performing eigenvalue decomposition on $\Sigma$, we have $\Sigma=\mathbf{D} \Lambda \mathbf{D}^T$ where $\Lambda=\mbox{diag}(\sigma_1, \ldots,\sigma_n)$ and $\mathbf{D}$ are the eigenvalues and eigenvectors respectively. Given the constraint $\mathbf{W} \mathbf{W}^T=\mathbf{I}$, we have $\mathbf{P}\mathbf{V} \mathbf{V}^T \mathbf{P}^T=\mathbf{I}$. Therefor, we have $\mathbf{P} \Sigma \mathbf{P}^T=\mathbf{I}$.

 By construction, we have
    \begin{eqnarray}
    \mathbf{P}=\mathbf{M} \Sigma^{-\frac{1}{2}}
    \end{eqnarray}
 where $\mathbf{M}$ is an $n \times n$ orthonomal matrix with $\mathbf{M} \mathbf{M}^T =\mathbf{M}^T \mathbf{M} =\mathbf{I}$. For convenience, we use $\Sigma^{-\frac{1}{2}}=\mathbf{D} \Lambda^{-1/2} \mathbf{D}^T$,  Therefore, we have:

   \begin{eqnarray}
\label{eqn:sovle}
 &&tr((\mathbf{W}-\mathbf{V}) (\mathbf{W}-\mathbf{V})^T) \nonumber \\
  &=& tr(\mathbf{W}\mathbf{W}^T -\mathbf{V} \mathbf{W}^T - \mathbf{W}\mathbf{V}^T +\mathbf{V} \mathbf{V}^T ) \nonumber  \\
 &=&tr(\mathbf{I}) +tr(\mathbf{\Sigma})-tr(\mathbf{V} \mathbf{W}^T)-tr((\mathbf{V} \mathbf{W}^T)^T)  \nonumber\\
 &=&tr(\mathbf{I}) +tr(\mathbf{\Sigma})-2tr(\mathbf{V} \mathbf{W}^T) )  \nonumber\\
  &=& d +tr(\mathbf{\Sigma})-2tr(\Sigma  \mathbf{P}^T) ) \nonumber \\
  &=& d +tr(\mathbf{\Sigma})-2tr(\mathbf{D} \Lambda \mathbf{D}^T \mathbf{D} \Lambda^{-1/2} \mathbf{D}^T  \mathbf{M}^T)   \nonumber\\
   &=& d +tr(\mathbf{\Sigma})-2tr(\mathbf{D} \Lambda^{1/2} \mathbf{D}^T  \mathbf{M}^T)
\end{eqnarray}

 Minimizing \ref{eqn:sovle} with respect to $\mathbf{P}$ is equivalent to maximizing $tr(\mathbf{D} \Lambda^{1/2} \mathbf{D}^T  \mathbf{M}^T) $ with respect to $\mathbf{M}$.  Note that $tr(\mathbf{D} \Lambda^{1/2} \mathbf{D}^T  \mathbf{M}^T)= tr( \Lambda^{1/2} \mathbf{D}^T  \mathbf{M}^T \mathbf{D})=tr(\Lambda^{1/2} \mathbf{Q})$ with  $\mathbf{Q}=\mathbf{D}^T  \mathbf{M}^T \mathbf{D}$ is orthogonal matrix. One important point fact is that $\mathbf{Q}_{ij} \leq 1$, otherwise the norm of the \emph{i}-th row or the \emph{j}-th column of $\mathbf{Q}$ will larger than 1, which is contradictory to that $\mathbf{Q}$ is real orthogonal matrix. Therefore, maximizing $tr(\Lambda^{1/2} \mathbf{Q}= \sum_{i=1}^n \Lambda_{ii}^{1/2} \mathbf{Q}_{ii})$ guarantees $\mathbf{Q}=\mathbf{I}$. We thus have $\mathbf{M}=\mathbf{D} \mathbf{I} \mathbf{D}^T=\mathbf{I}$. Therefore, we get the solution $\mathbf{P}=\Sigma^{-\frac{1}{2}}=\mathbf{D} \Lambda^{-1/2} \mathbf{D}^T$.

\subsection{Riemannian Optimization over Stiefel Manifold}
\label{Riem_p3}
For comparison purpose, here we  review the Riemannian optimization over Stiefel manifold briefly and for more details please refer to \cite{2008_Book_Absil} and references therein.
The objective is  $\arg \min_{\mathbf{W} \in \mathbb{M}} f(\mathbf{W})$, where $f$ is a real-value smooth function over $\mathbb{M}=\{ \mathbf{W} \in \mathbb{R}^{n \times d}: \mathbf{W}^T \mathbf{W}= \mathbf{I}, n \geqslant d \}$.
 Note that in this section, we follow the common description for  Riemannian optimization over Stiefel manifold with the columns of $\mathbf{W}$ being $d$ orthonormal vectors in $\mathbb{R}^n$, and therefore with the constraints $\mathbf{W}^T \mathbf{W}= \mathbf{I}$.
 It is different to the description of our formulation in the paper with constraints $\mathbf{W} \mathbf{W}^T= \mathbf{I}$ and $n \leqslant d$.

Conventional optimization techniques are based gradient descent method over manifold by iteratively seeking for updated points $\mathbf{W}_t \in \mathbb{M}$. In each iteration $t$, the keys are: (1) finding the Riemannian gradient $G^{\mathbb{M}}~f(\mathbf{W}_t) \in T_{\mathbf{W}_t}$ where $T_{\mathbf{W}_t}$ is the tangent space of $\mathbb{M}$ at current point $\mathbf{W}_t$; and (2) finding the descent direction and ensuring that the new points is on the manifold $\mathbb{M}$.

For obtaining the Riemannian gradient $G^{\mathbb{M}}~f(\mathbf{W})$, the inner dot  should be defined in $T_{\mathbf{W}}$. There are two extensively used inner products for tangent space of Stiefel manifold ~\cite{2013_MP_Wen}:
 (1) \emph{Euclidean inner product}: $<\mathbf{X}_1, \mathbf{X}_2>_e=tr(\mathbf{X}_1^T \mathbf{X}_2)$ and
  (2) \emph{canonical inner product}: $<\mathbf{X}_1, \mathbf{X}_2>_c=tr(\mathbf{X}_1^T (\mathbf{I}-\frac{1}{2} \mathbf{W} \mathbf{W}^T)\mathbf{X}_2)$ where $\mathbf{X}_1, \mathbf{X}_2 \in T_{\mathbf{W}}$ and $tr(\cdot)$ denote the trace of the matrix. Based on these two inner products, the respective Riemannian gradient can be obtained as~\cite{2013_MP_Wen}:
  \begin{eqnarray}
\label{eqn:RG-Euclean}
G_e^{\mathbb{M}}~f(\mathbf{W})=\frac{\partial f}{\partial \mathbf{W}}-\mathbf{W} \frac{\partial f}{\partial \mathbf{W}}^T \mathbf{W}
\end{eqnarray}
and
  \begin{eqnarray}
\label{eqn:RG-cannic}
G_c^{\mathbb{M}}~f(\mathbf{W})=\frac{\partial f}{\partial \mathbf{W}}-\frac{1}{2}(\mathbf{W}  \mathbf{W}^T \frac{\partial f}{\partial \mathbf{W}} + \mathbf{W} \frac{\partial f}{\partial \mathbf{W}}^T \mathbf{W})
\end{eqnarray}
 where $\frac{\partial f}{\partial \mathbf{W}}$ is the ordinary gradient.

 Given the Riemannian gradient, the next step is to find the descent direction and guarantee that the new point is on the manifold $\mathbb{M}$, which can be supported by the so called operation \emph{retraction}. One well recommended \emph{retraction}is the QR-Decomposition-Type retraction ~\cite{2013_TSP_Kaneko,2017_Corr_Harandi} that maps a tangent vector of $T_\mathbf{W}$ onto $\mathbb{M}$ by: $P_\mathbf{W}(\mathbf{Z})=qf(\mathbf{W}+\mathbf{Z})$,
 where $qf(\cdot)$ denotes the $Q$ factor of  the QR decomposition with $Q \in \mathbb{M}$, and the R-factor is an upper-trangular matrix with strictly positive elements on its main diagonal such that the decomposition is unique ~\cite{2013_TSP_Kaneko}. Given the Riemannian gradient $G^{\mathbb{M}}~f(\mathbf{W})$ and the learning rate $\eta$, the new point is:
   \begin{eqnarray}
\label{eqn:QR-retract}
\mathbf{W}_{t+1}= qf(\mathbf{W}_t - \eta ~G^{\mathbb{M}}~f(\mathbf{W}))
\end{eqnarray}

 Another well known technique to jointly move in the descent direction and make sure the new solution on the manifold $\mathbb{M}$ is Cayley transformation ~\cite{2013_MP_Wen,2016_NIPS_Wisdom,2017_ICML_Eugene}. It produces the feasible solution $\mathbf{W}_{k+1}$ with the current solution $\mathbf{W}_{k}$ by:
   \begin{eqnarray}
\label{eqn:CayT}
\mathbf{W}_{t+1}=(\mathbf{I}+ \frac{\eta}{2} A_t )^{-1} (\mathbf{I} - \frac{\eta}{2} A_t ) \mathbf{W}
\end{eqnarray}
 where $\eta$ is the learning rate and $A_t=\frac{\partial F}{\partial \mathbf{W}_t}^T  \mathbf{W}_t - \mathbf{W}_t^T \frac{\partial F}{\partial \mathbf{W}_t}$ that is induced by the defined canonical inner product in the tangent space.
%
%

\subsection{More Experimental Results}
\label{exp_p3}
\begin{figure*}[]
\centering
  \vspace{-0in}
  \subfigure[EI+QR]{
  \includegraphics[width=0.23\linewidth]{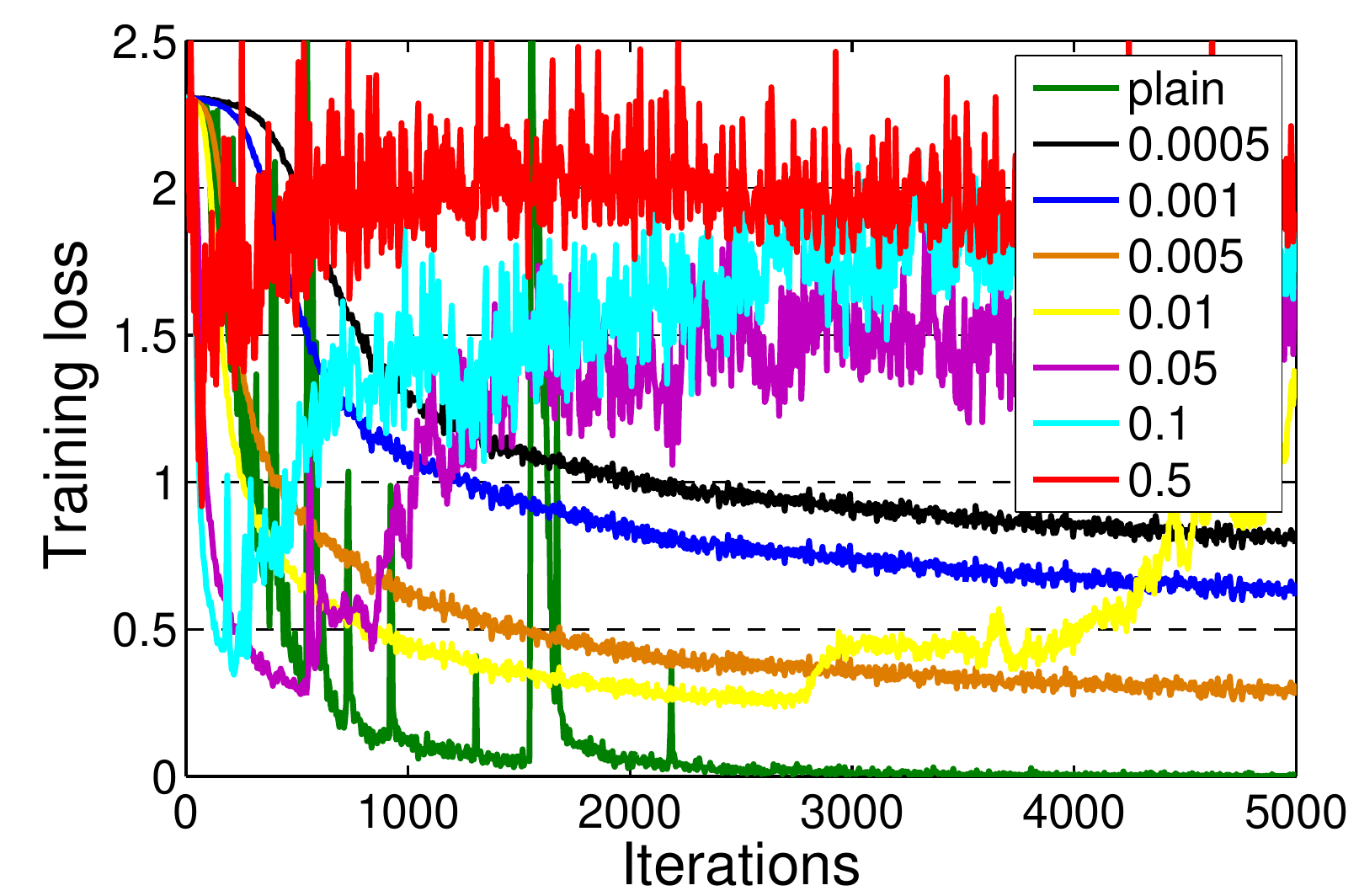}
  }
  \subfigure[CI+QR]{
  \includegraphics[width=0.23\linewidth]{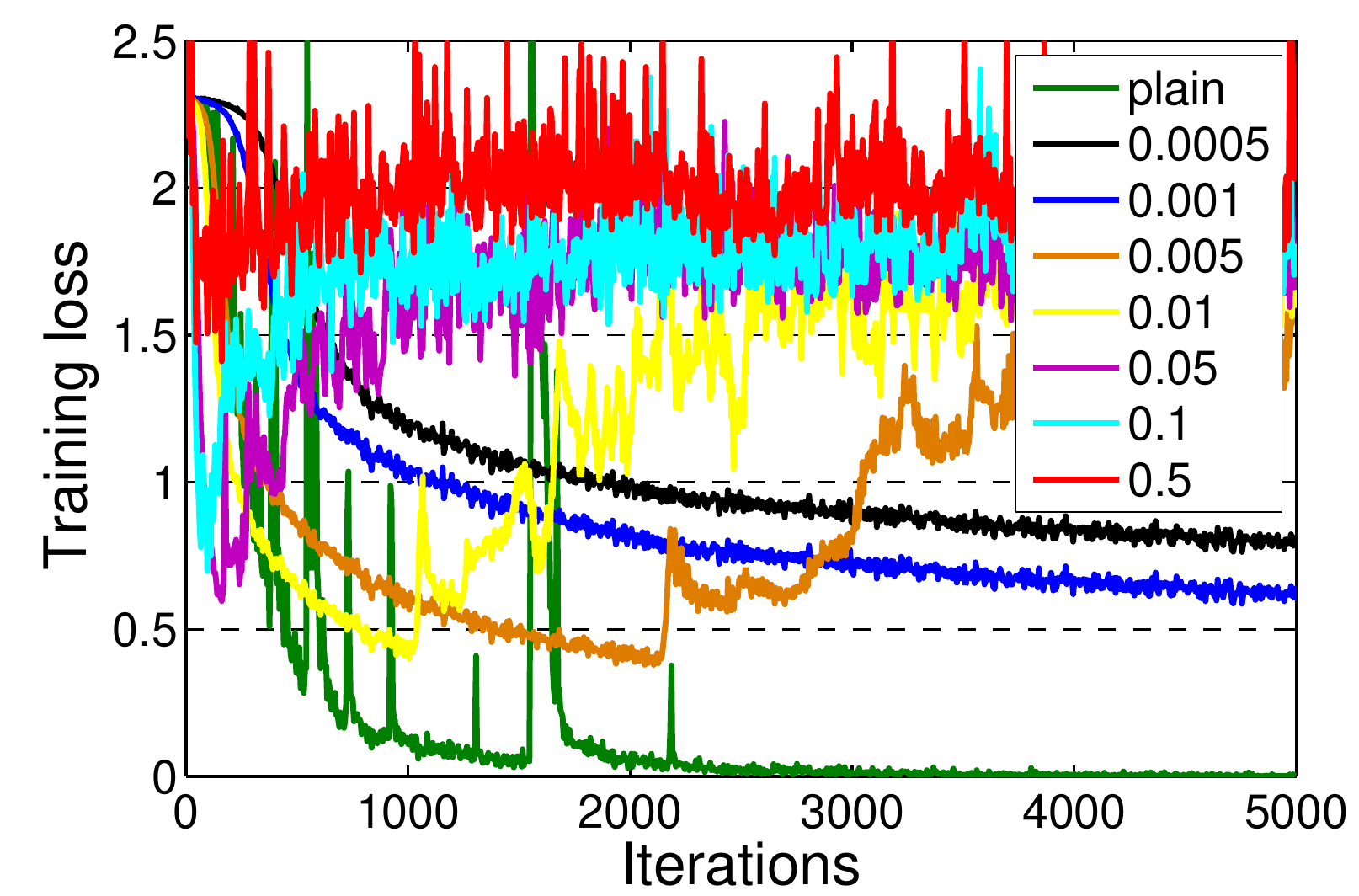}
  }
  \subfigure[CayT]{
  \includegraphics[width=0.23\linewidth]{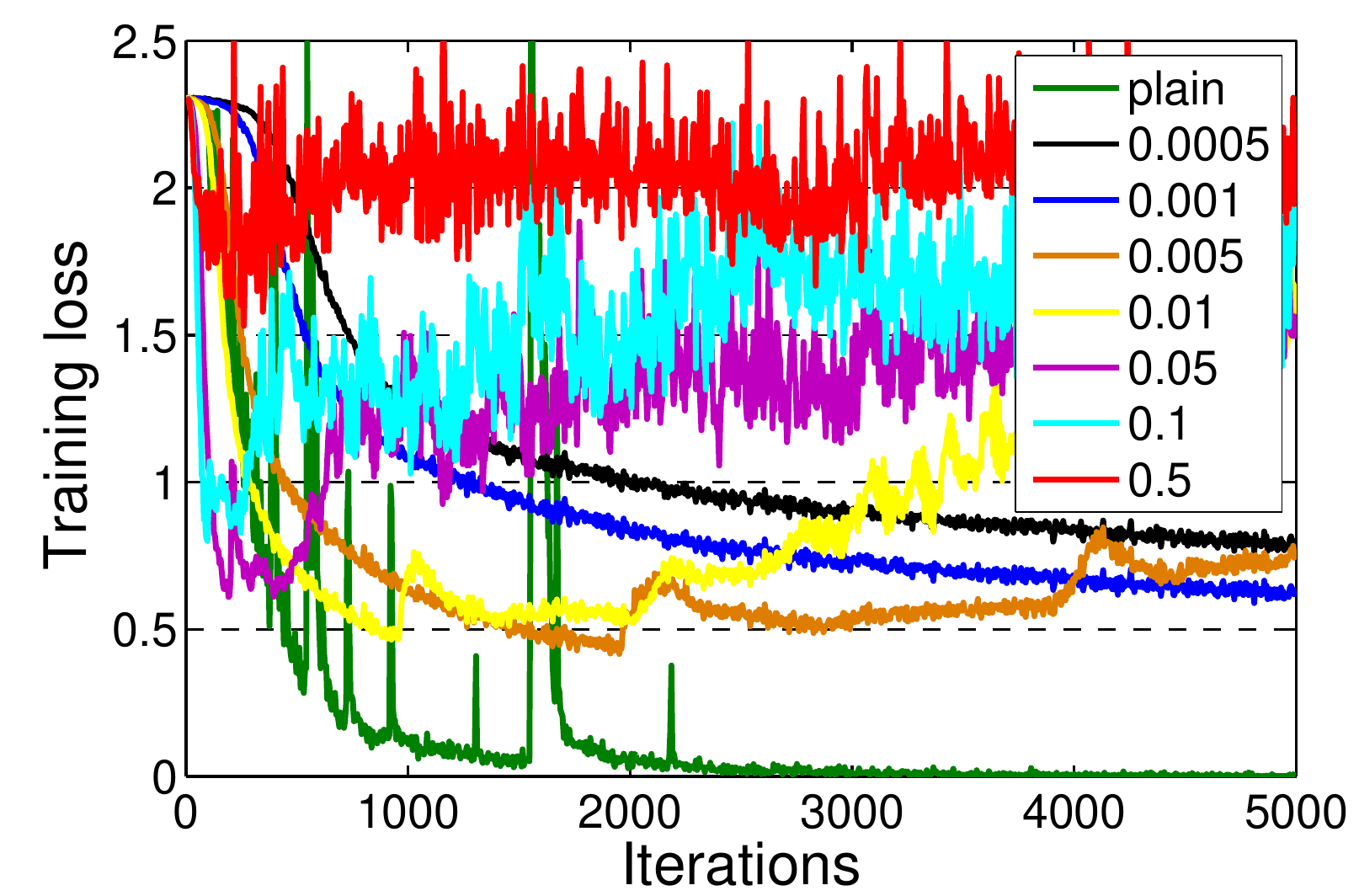}
  }
  \subfigure[Our OLM]{
  \includegraphics[width=0.23\linewidth]{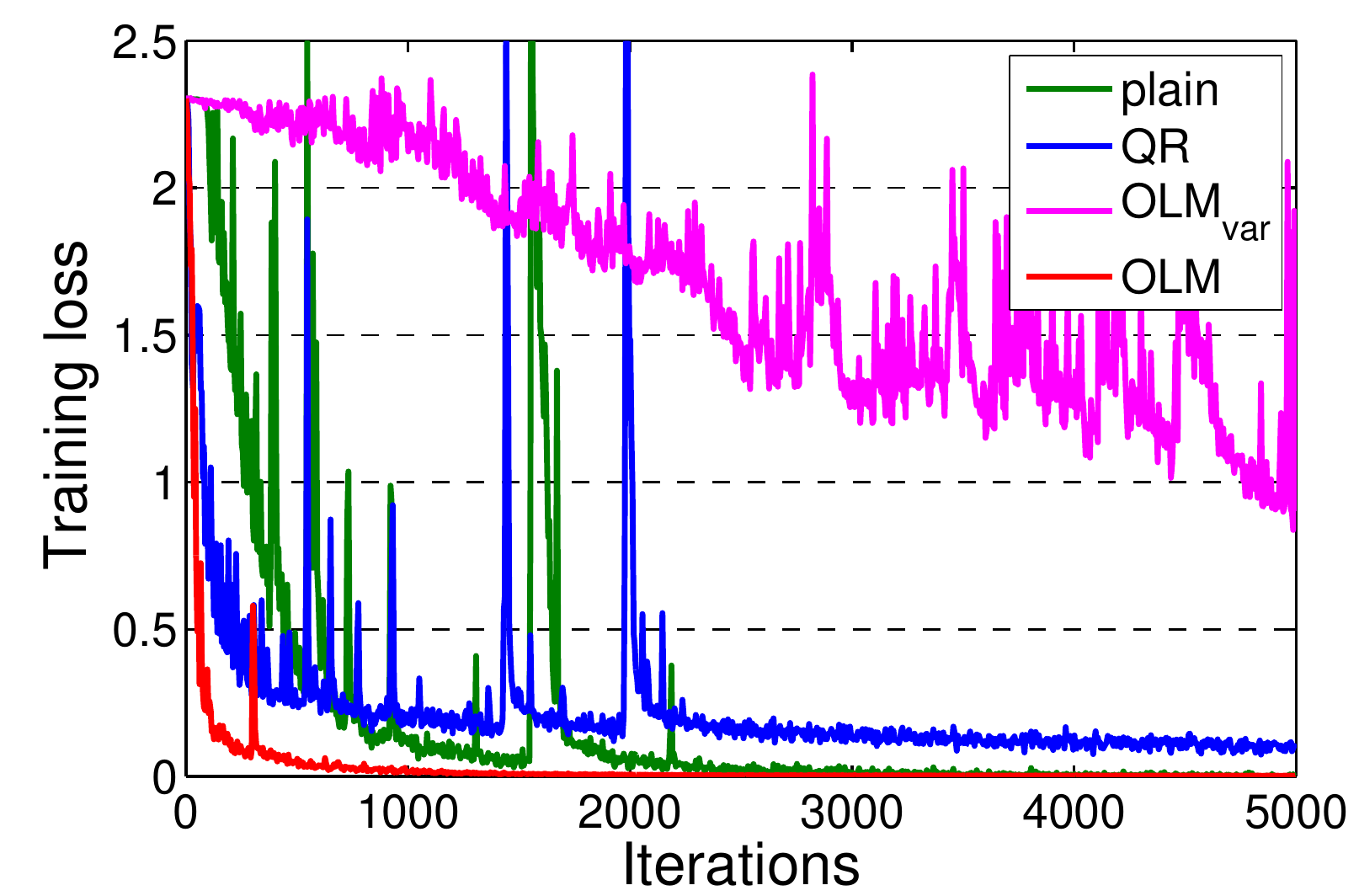}
  }
   \vspace{-0.15in}
  \caption{\small Results of Riemannian optimization  methods to solve OMDSM on MNIST dataset under the 6-layer MLP. We  train the model with batch size of 1024 and show the training loss curves for different learning rate of `EI+QR', `CI+QR' and  `CayT' compared to the baseline `plain'  in (a), (b) and (c) respectably. We compare our methods to baselines  and report the best performance among all learning rates based on the training loss for each method in (d).}
  \label{fig:exp_MLP_b1024_6layer}
  \vspace{-0.1in}
\end{figure*}

\begin{figure*}[]
\centering
  \vspace{-0in}
  \subfigure[EI+QR]{
  \includegraphics[width=0.23\linewidth]{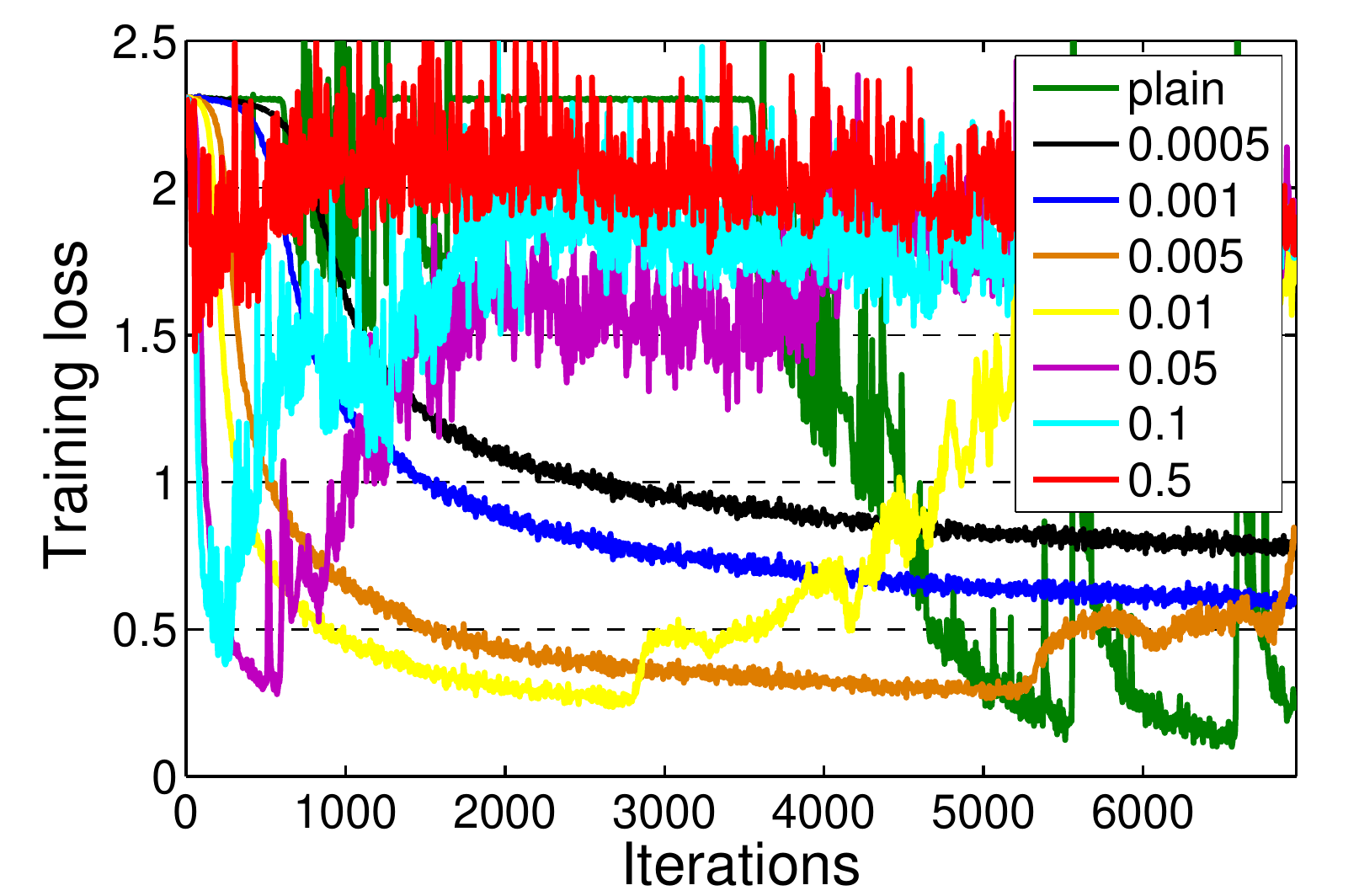}
  }
  \subfigure[CI+QR]{
  \includegraphics[width=0.23\linewidth]{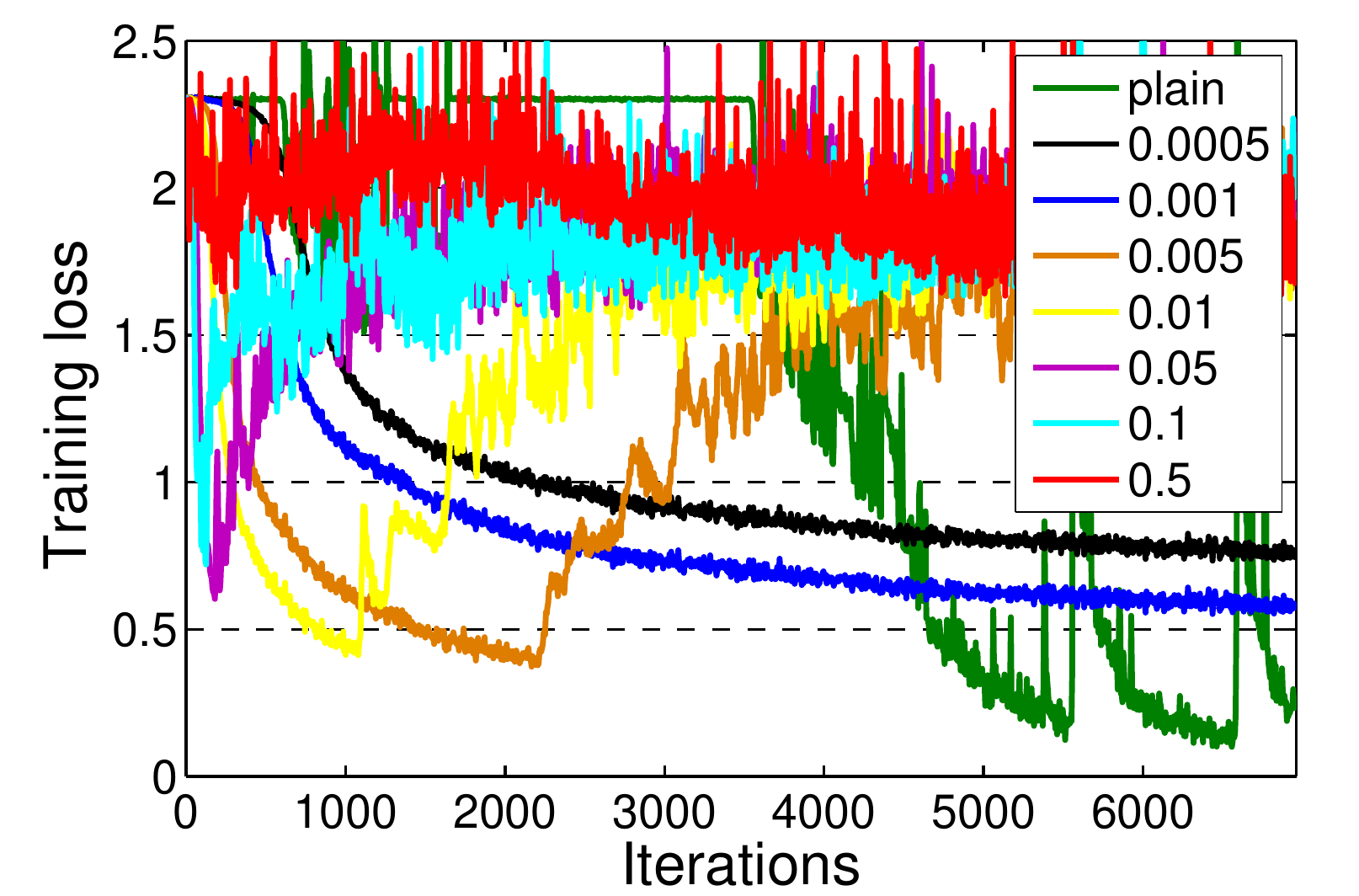}
  }
  \subfigure[CayT]{
  \includegraphics[width=0.23\linewidth]{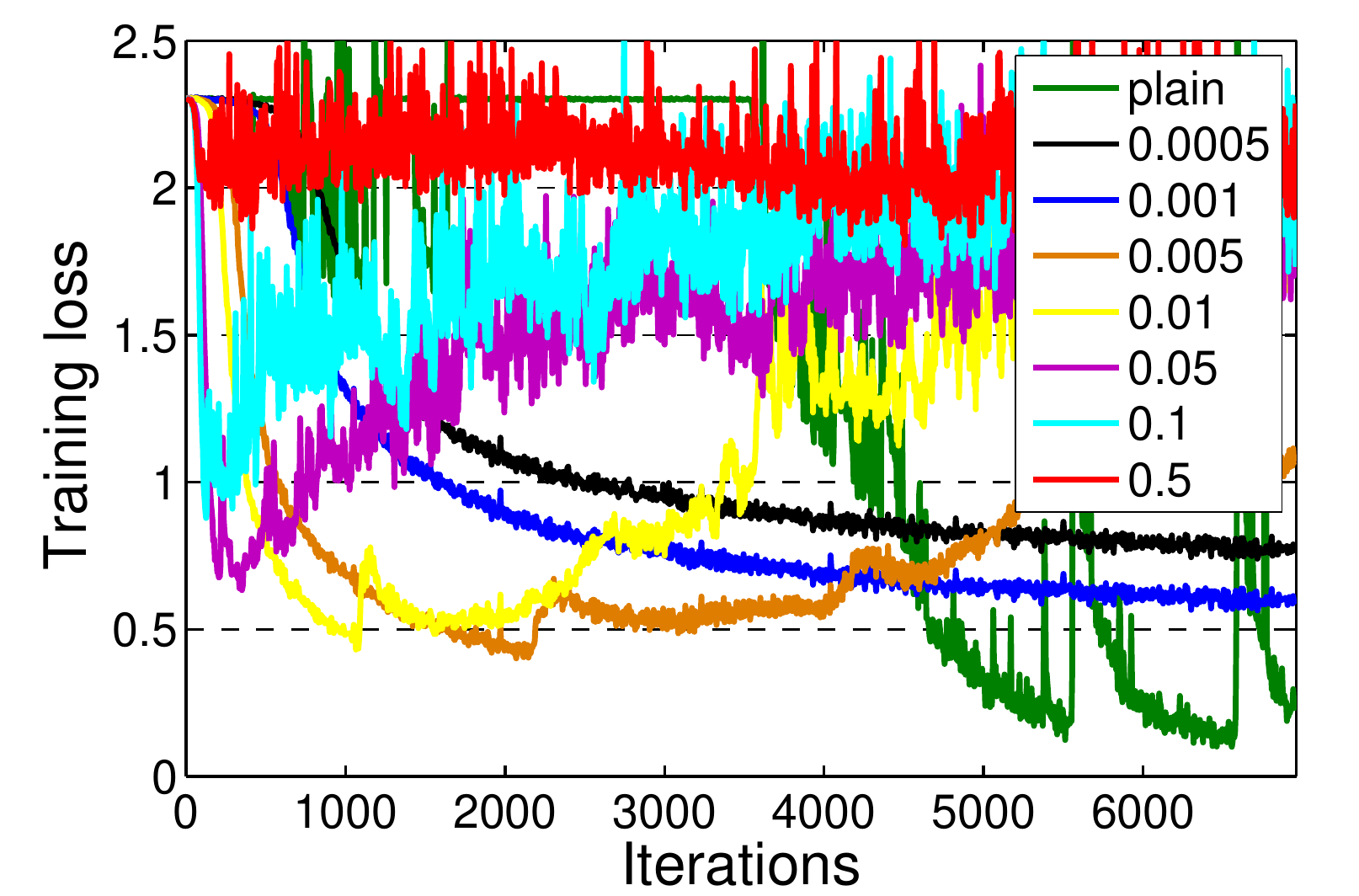}
  }
  \subfigure[Our OLM]{
  \includegraphics[width=0.23\linewidth]{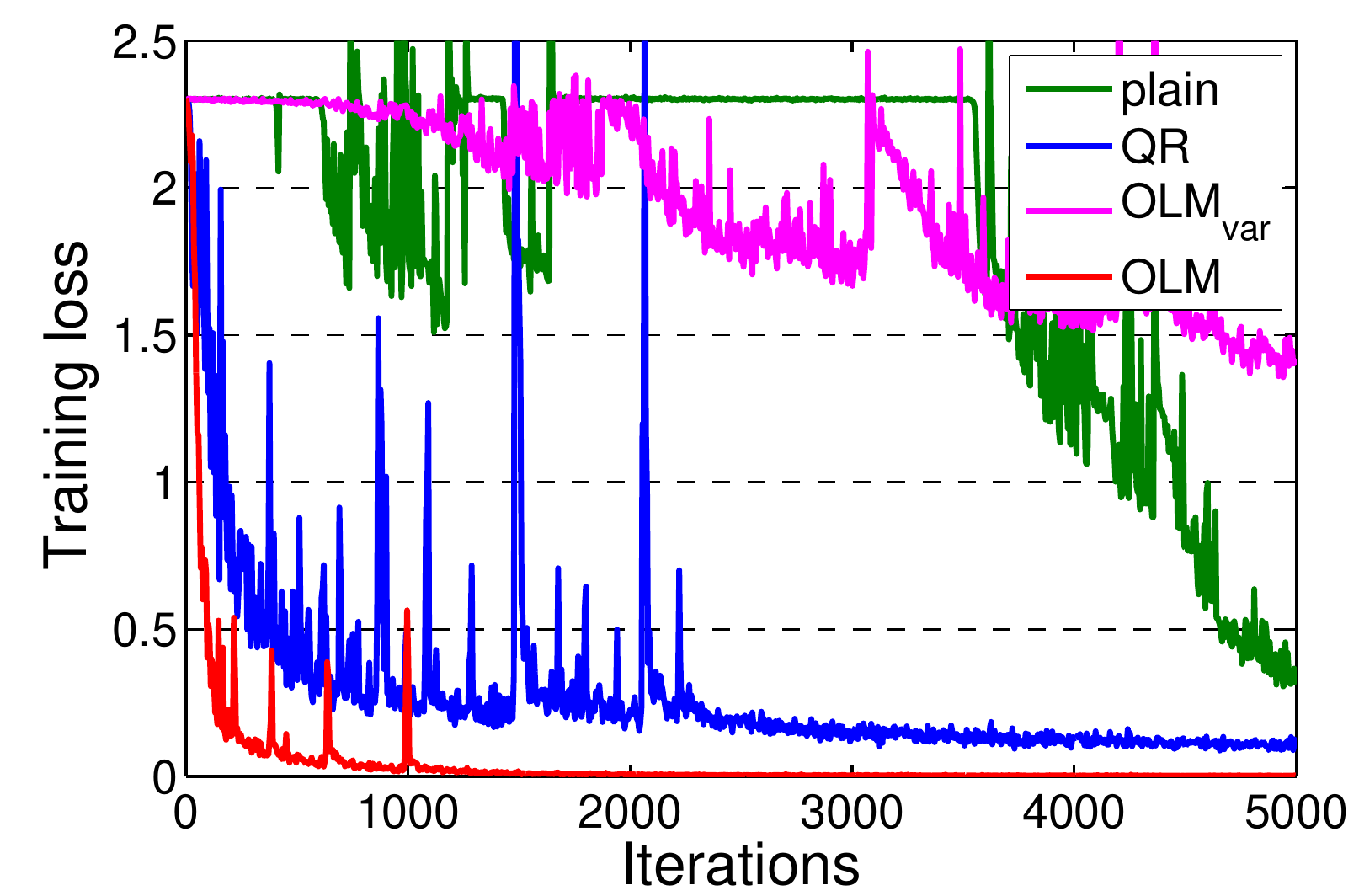}
  }
   \vspace{-0.15in}
  \caption{\small Results of Riemannian optimization  methods to solve OMDSM on MNIST dataset under the 8-layer MLP. We  train the model with batch size of 1024 and show the training loss curves for different learning rate of `EI+QR', `CI+QR' and  `CayT' compared to the baseline `plain'  in (a), (b) and (c) respectably. We compare our methods to baselines  and report the best performance among all learning rates based on the training loss for each method in (d).}
  \label{fig:exp_MLP_b1024_8layer}
  \vspace{-0.1in}
\end{figure*}

\begin{figure*}[]
\centering
  \vspace{-0in}
  \subfigure[EI+QR]{
  \includegraphics[width=0.23\linewidth]{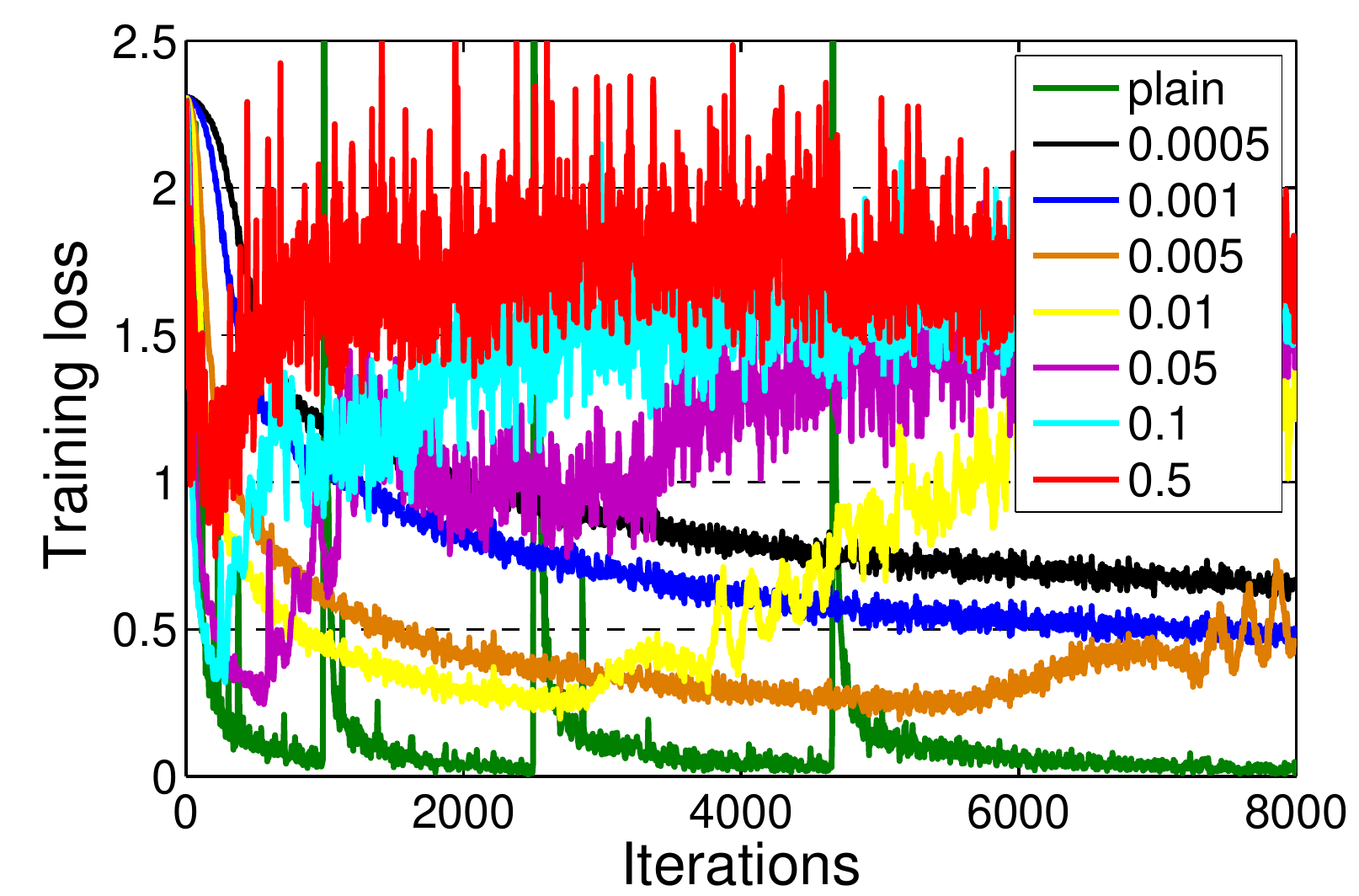}
  }
  \subfigure[CI+QR]{
  \includegraphics[width=0.23\linewidth]{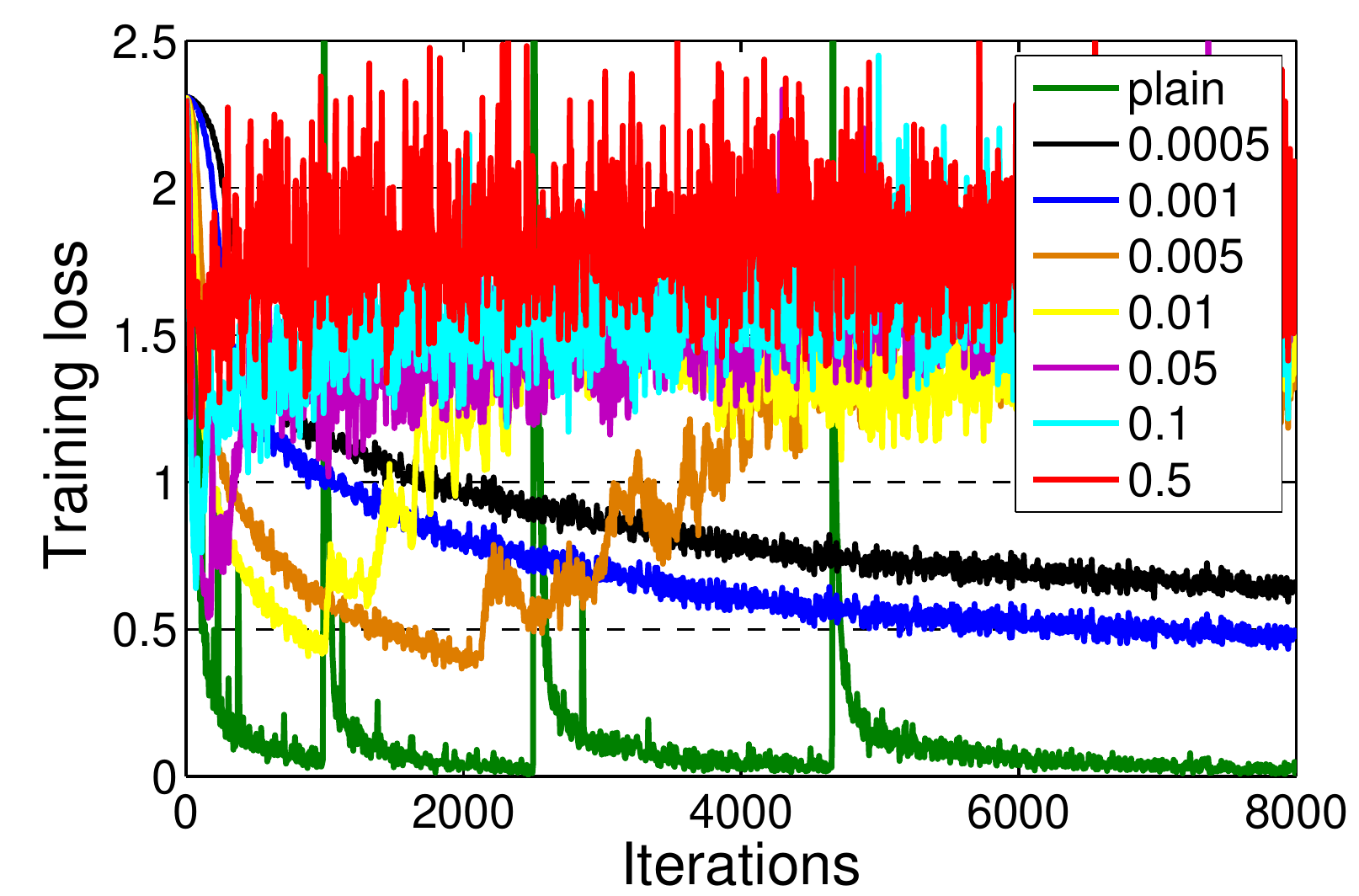}
  }
  \subfigure[CayT]{
  \includegraphics[width=0.23\linewidth]{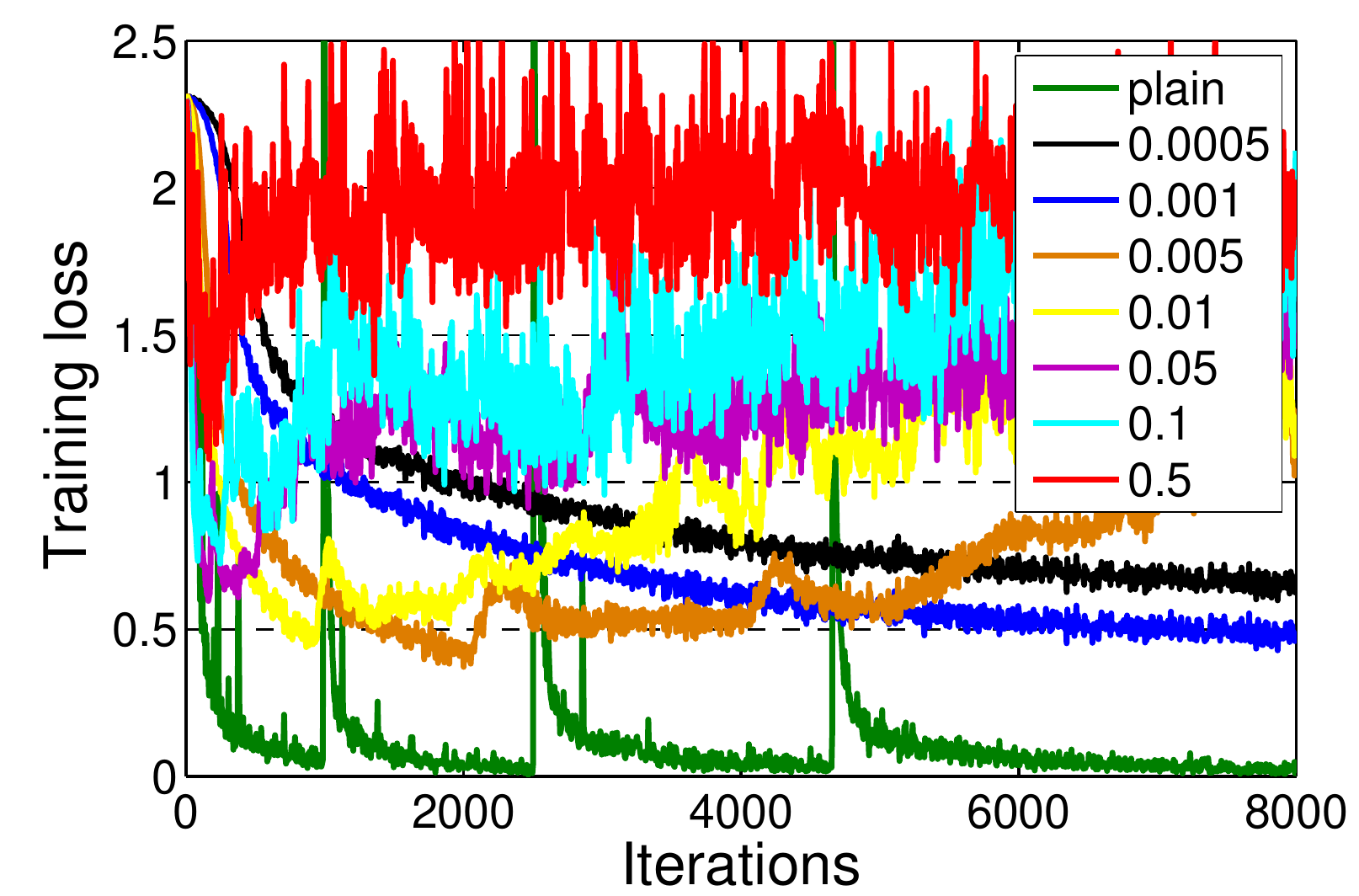}
  }
  \subfigure[Our OLM]{
  \includegraphics[width=0.23\linewidth]{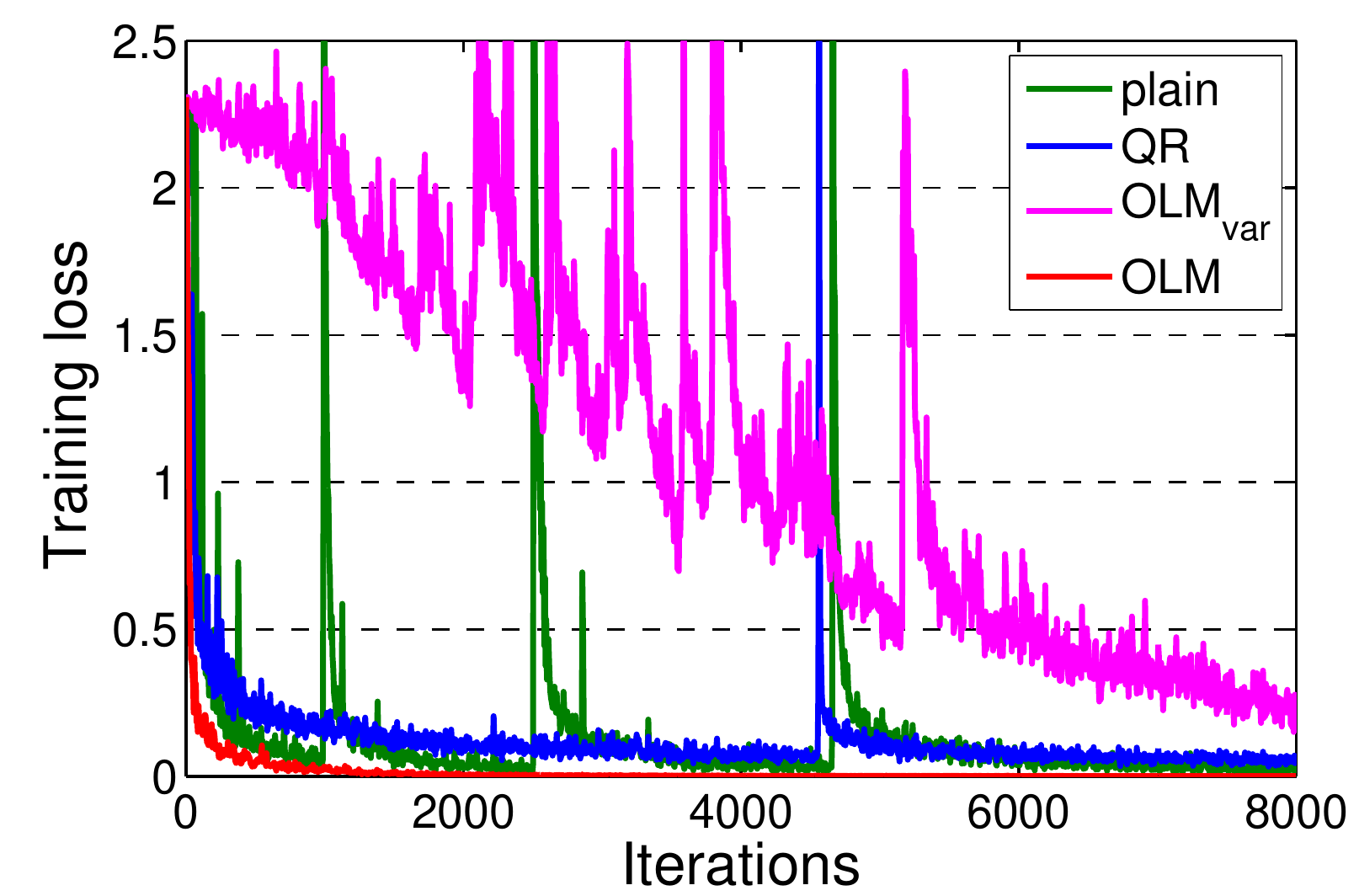}
  }
   \vspace{-0.15in}
  \caption{\small Results of Riemannian optimization  methods to solve OMDSM on MNIST dataset under the 4-layer MLP. We  train the model with batch size of 512 and show the training loss curves for different learning rate of `EI+QR', `CI+QR' and  `CayT' compared to the baseline `plain'  in (a), (b) and (c) respectably. We compare our methods to baselines  and report the best performance among all learning rates based on the training loss for each method in (d).}
  \label{fig:exp_MLP_b512_4layer}
  \vspace{-0.1in}
\end{figure*}

\begin{figure*}[]
\centering
  \vspace{-0in}
  \subfigure[EI+QR]{
  \includegraphics[width=0.23\linewidth]{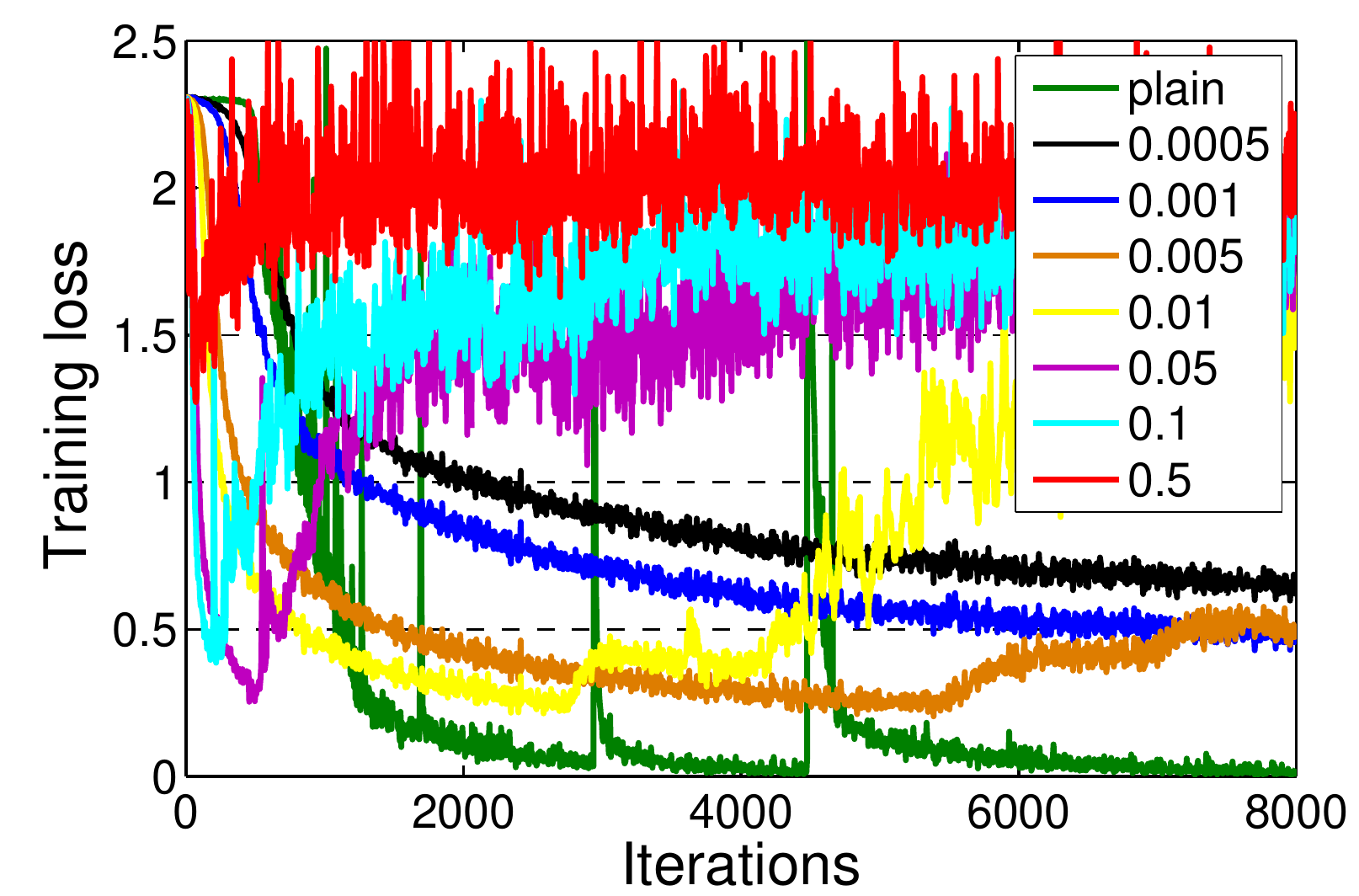}
  }
  \subfigure[CI+QR]{
  \includegraphics[width=0.23\linewidth]{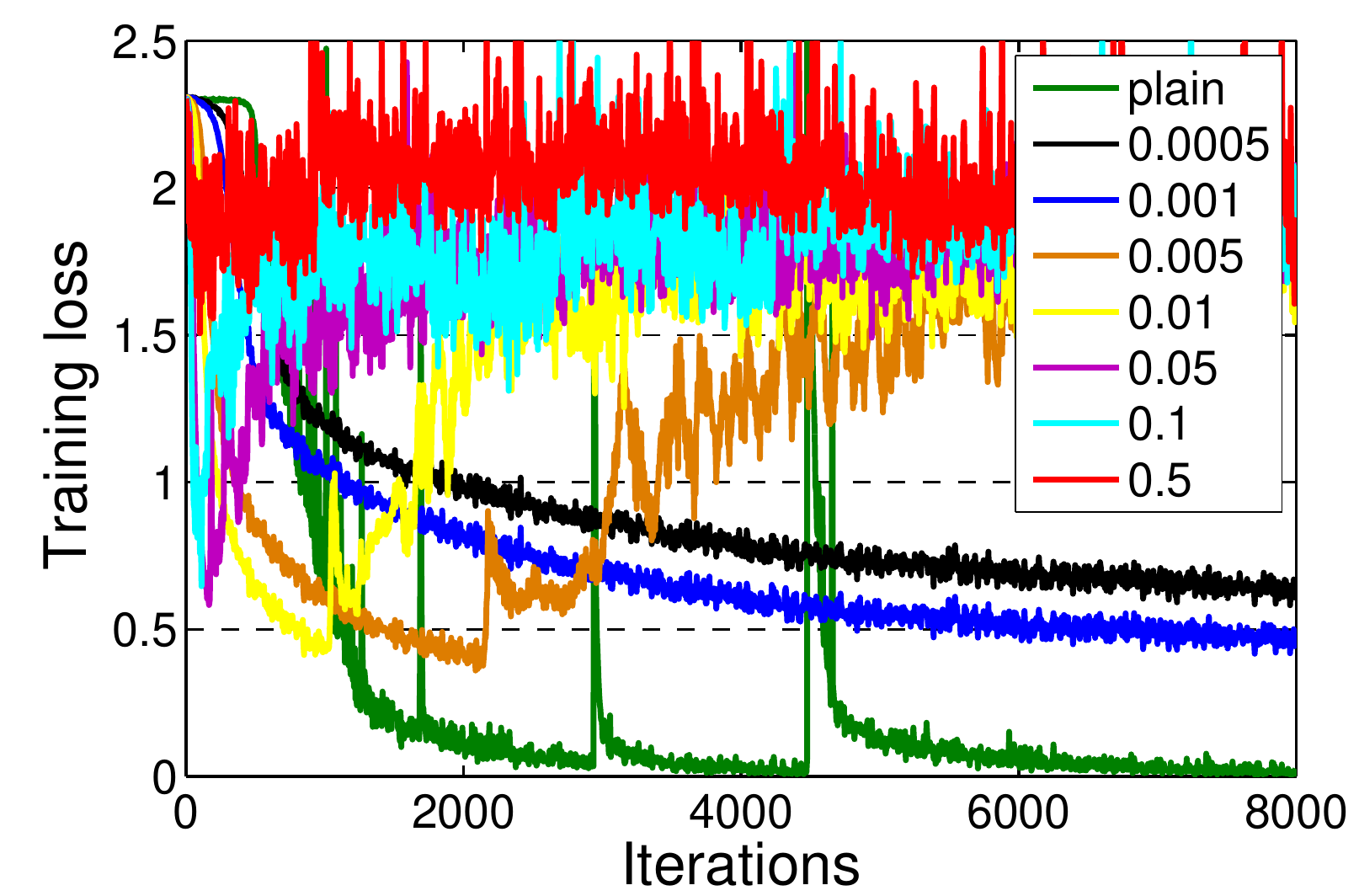}
  }
  \subfigure[CayT]{
  \includegraphics[width=0.23\linewidth]{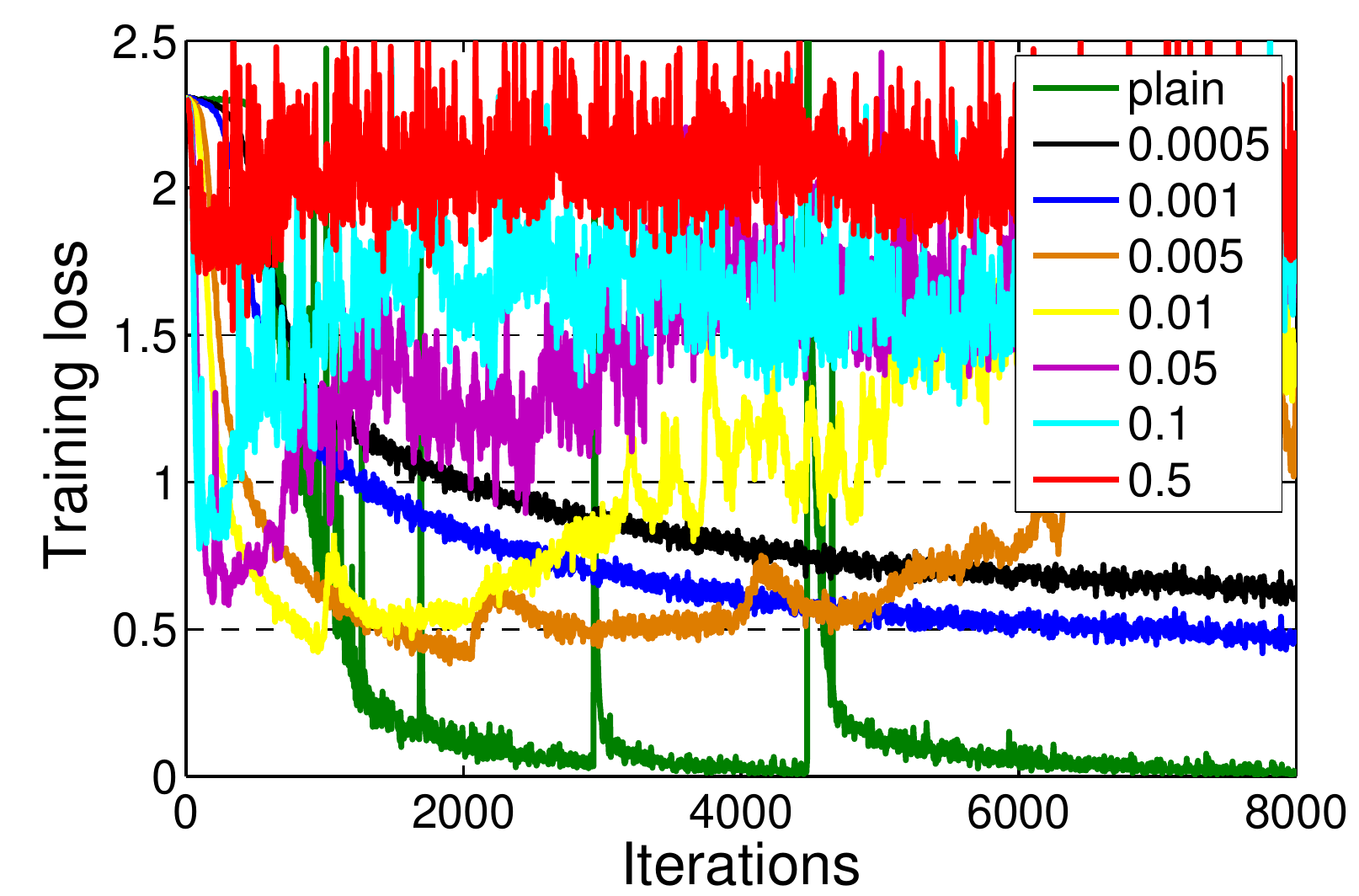}
  }
  \subfigure[Our OLM]{
  \includegraphics[width=0.23\linewidth]{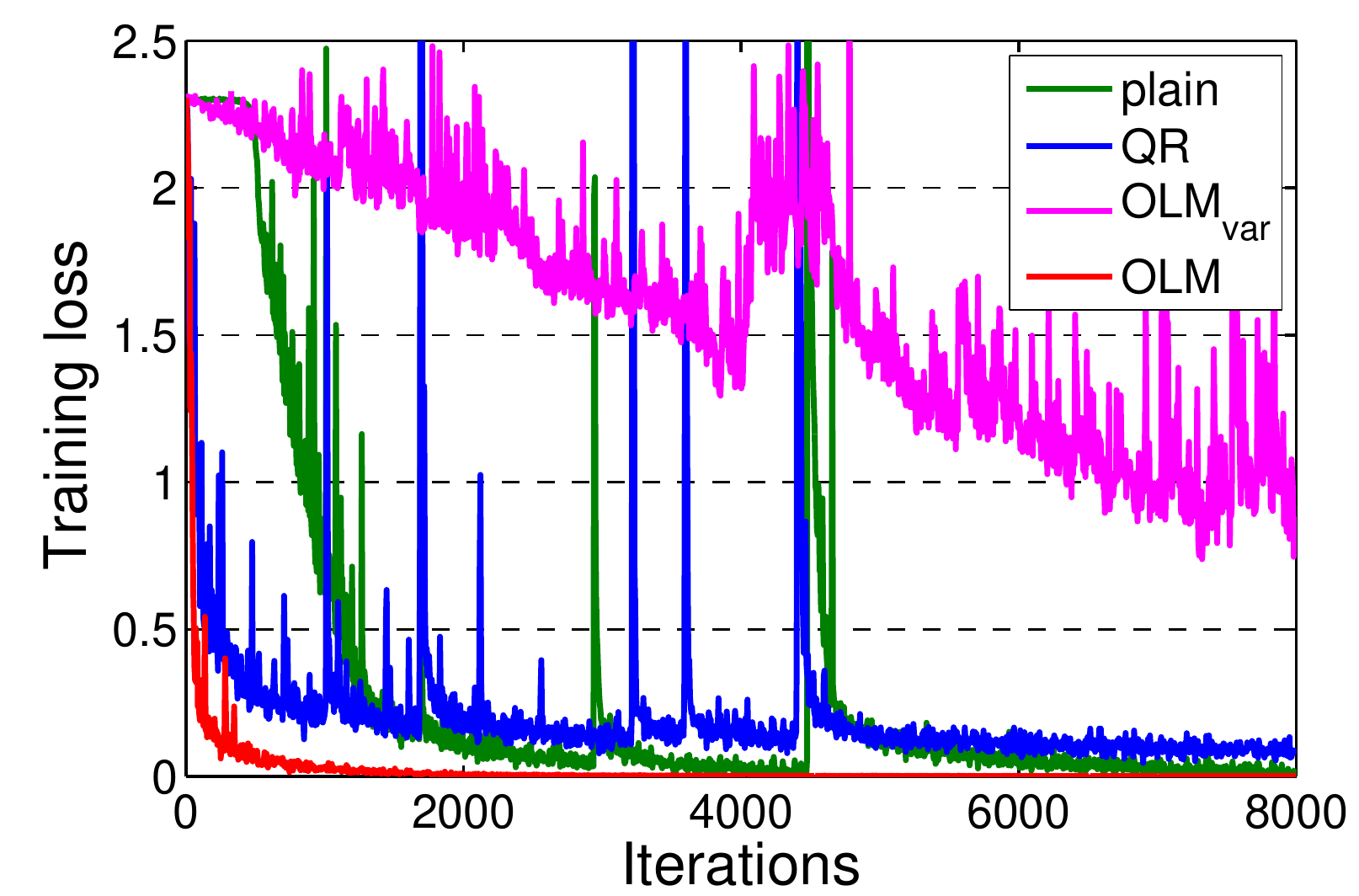}
  }
   \vspace{-0.15in}
  \caption{\small Results of Riemannian optimization  methods to solve OMDSM on MNIST dataset under the 6-layer MLP. We  train the model with batch size of 512 and show the training loss curves for different learning rate of `EI+QR', `CI+QR' and  `CayT' compared to the baseline `plain'  in (a), (b) and (c) respectably. We compare our methods to baselines  and report the best performance among all learning rates based on the training loss for each method in (d).}
  \label{fig:exp_MLP_b512_6layer}
  \vspace{-0.1in}
\end{figure*}

\begin{figure*}[]
\centering
  \vspace{-0in}
  \subfigure[EI+QR]{
  \includegraphics[width=0.23\linewidth]{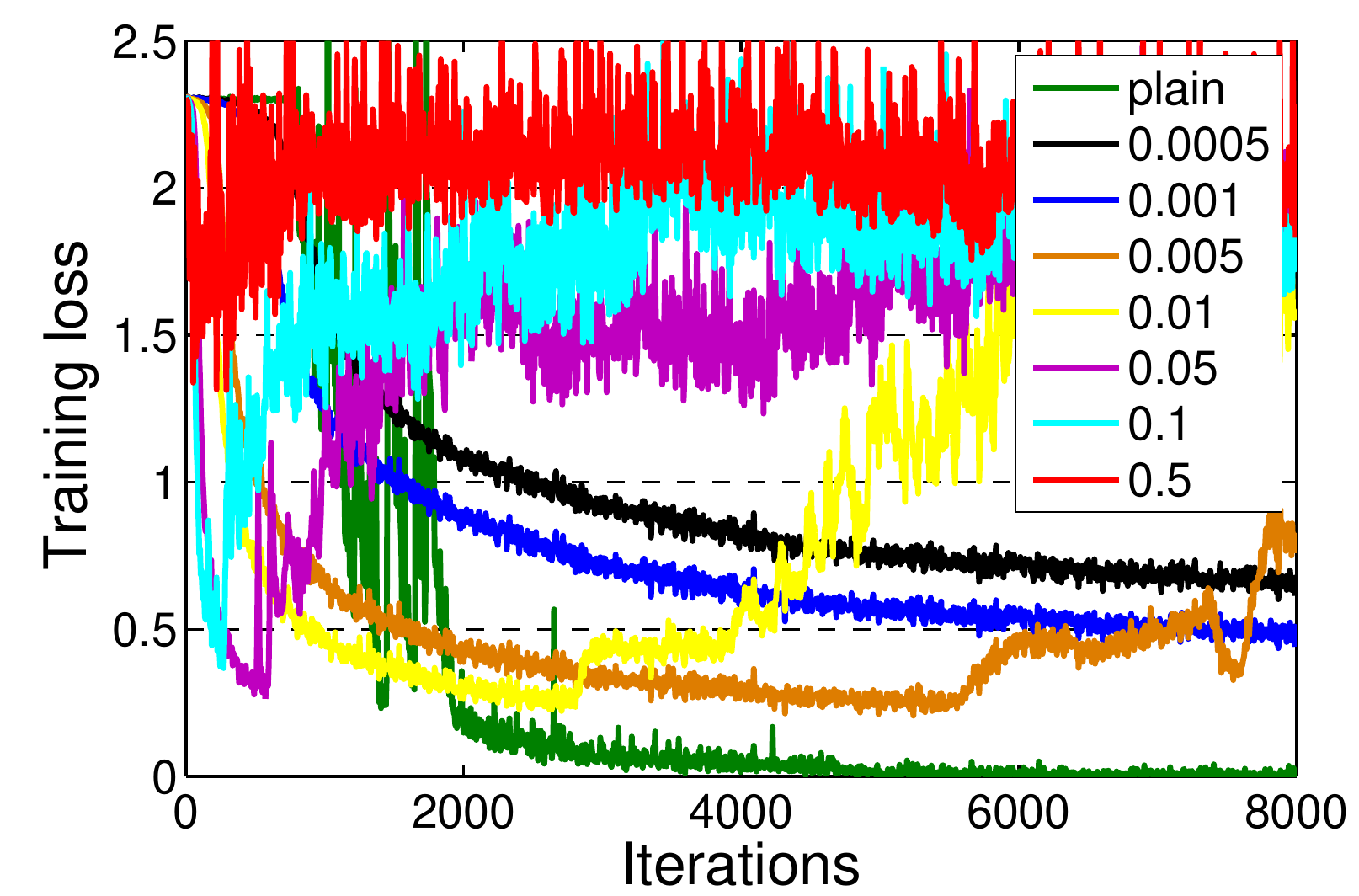}
  }
  \subfigure[CI+QR]{
  \includegraphics[width=0.23\linewidth]{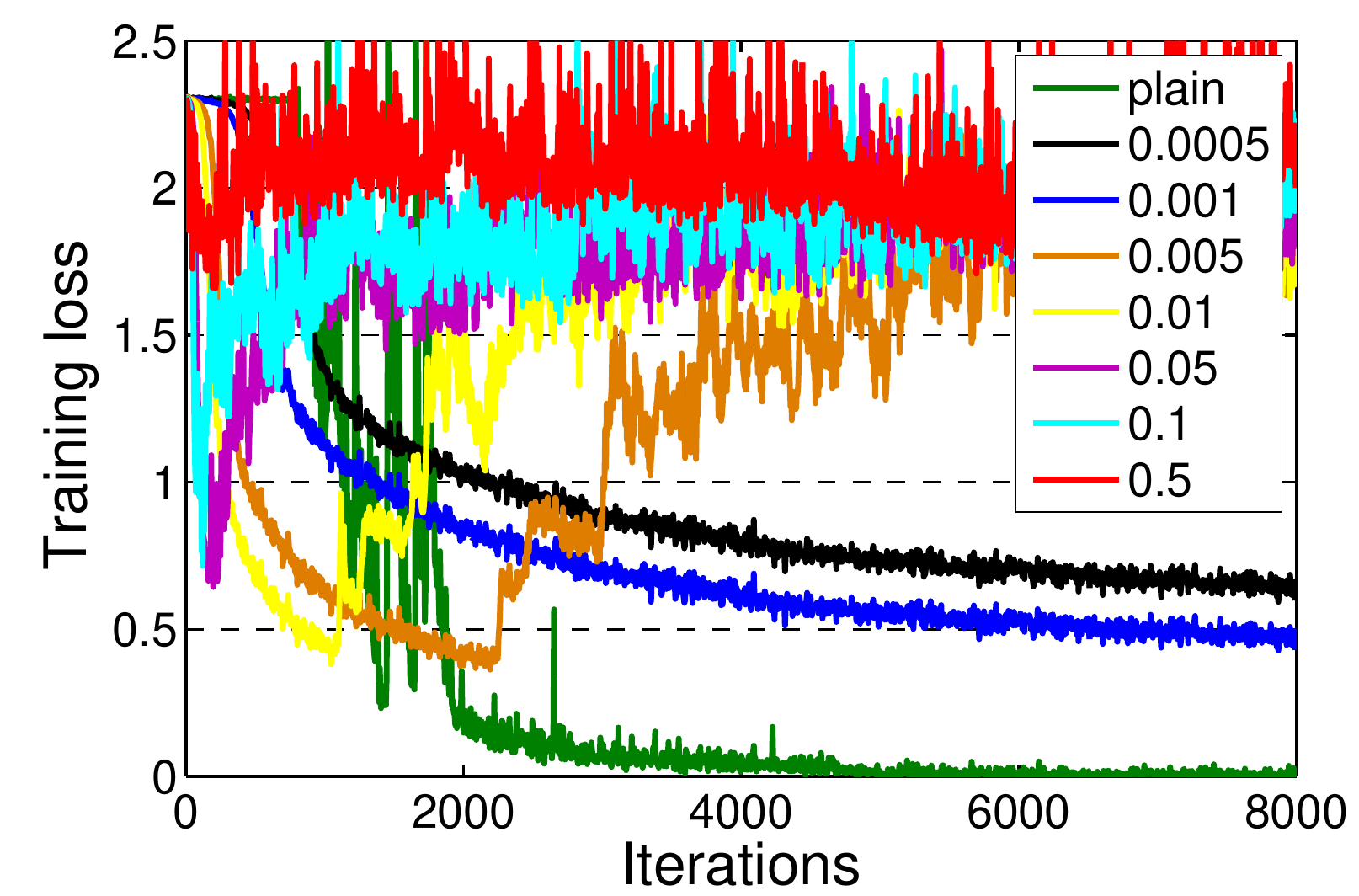}
  }
  \subfigure[CayT]{
  \includegraphics[width=0.23\linewidth]{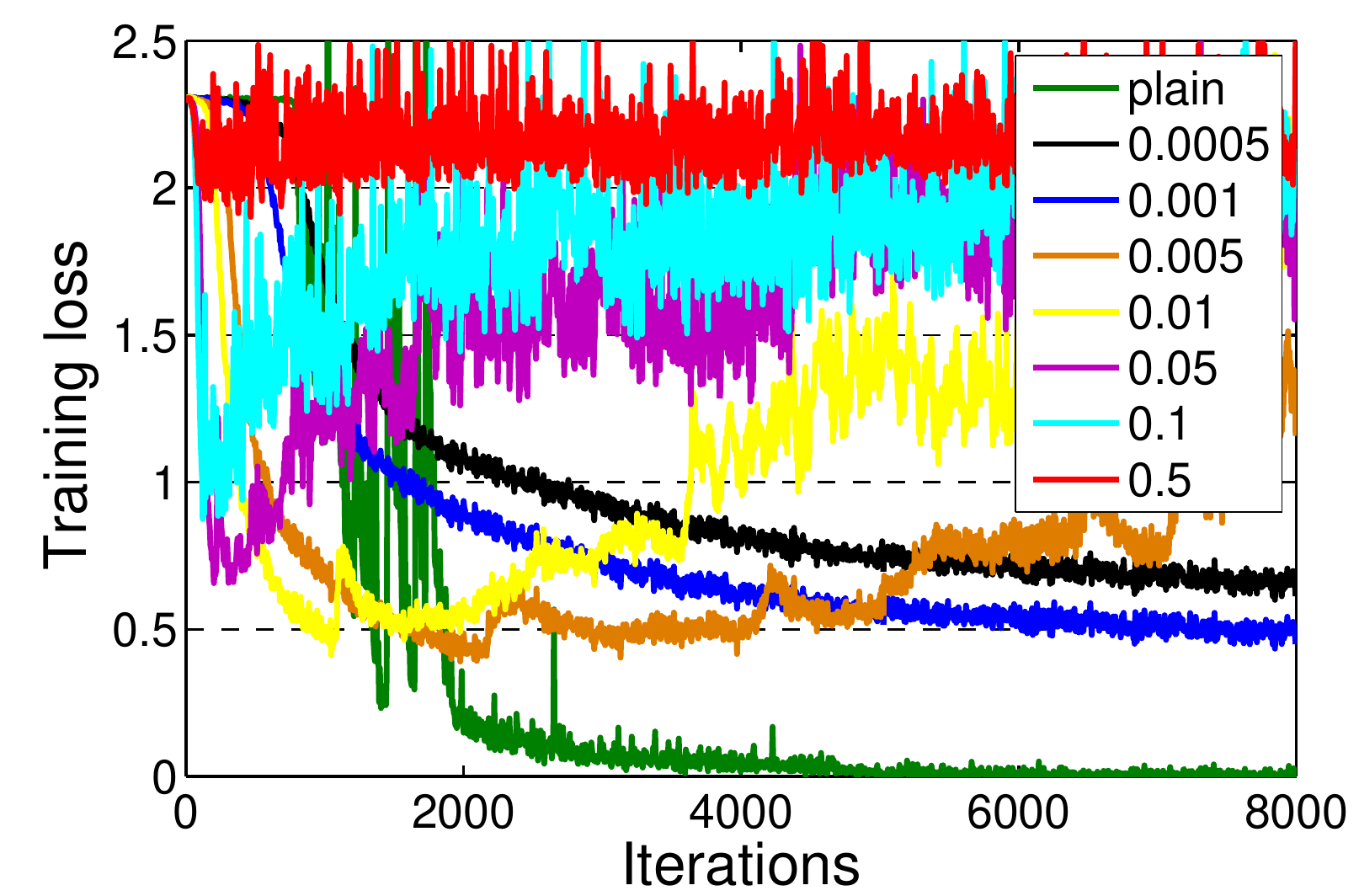}
  }
  \subfigure[Our OLM]{
  \includegraphics[width=0.23\linewidth]{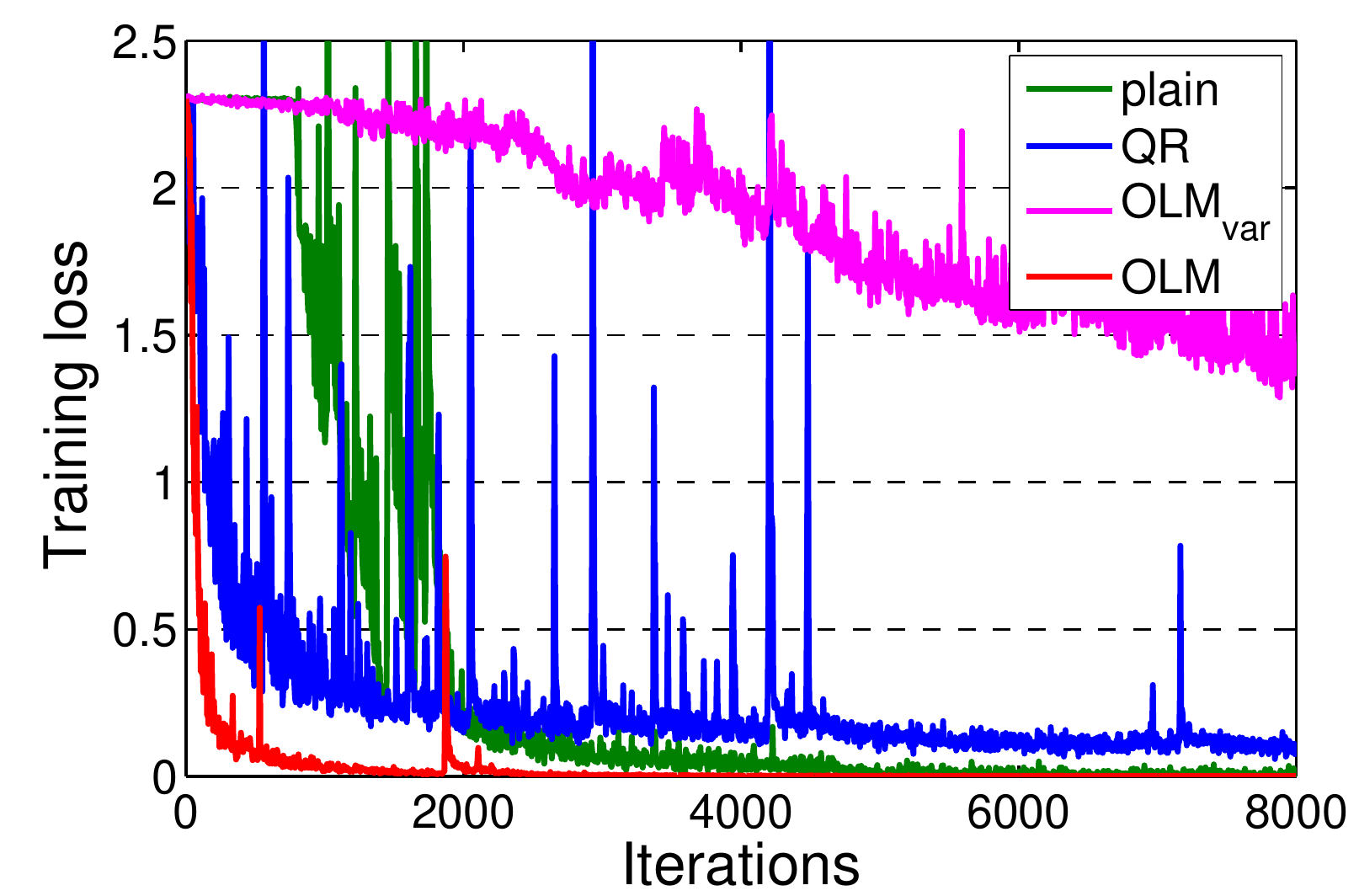}
  }
   \vspace{-0.15in}
  \caption{\small Results of Riemannian optimization  methods to solve OMDSM on MNIST dataset under the 8-layer MLP. We  train the model with batch size of 512 and show the training loss curves for different learning rate of `EI+QR', `CI+QR' and  `CayT' compared to the baseline `plain'  in (a), (b) and (c) respectably. We compare our methods to baselines  and report the best performance among all learning rates based on the training loss for each method in (d).}
  \label{fig:exp_MLP_b512_8layer}
  \vspace{-0.1in}
\end{figure*}

\begin{figure*}[]
\centering
  \vspace{-0in}
  \subfigure[EI+QR]{
  \includegraphics[width=0.23\linewidth]{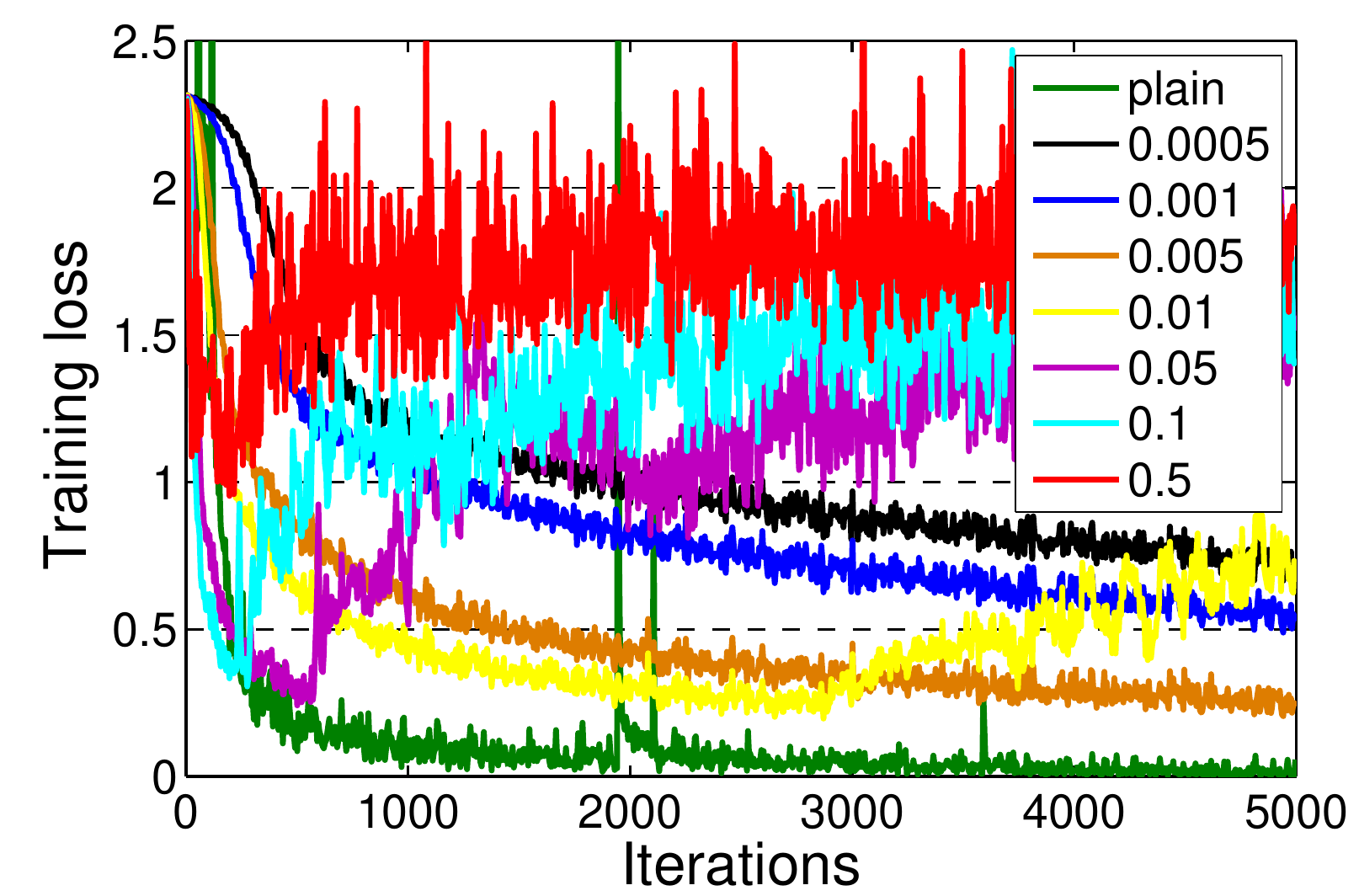}
  }
  \subfigure[CI+QR]{
  \includegraphics[width=0.23\linewidth]{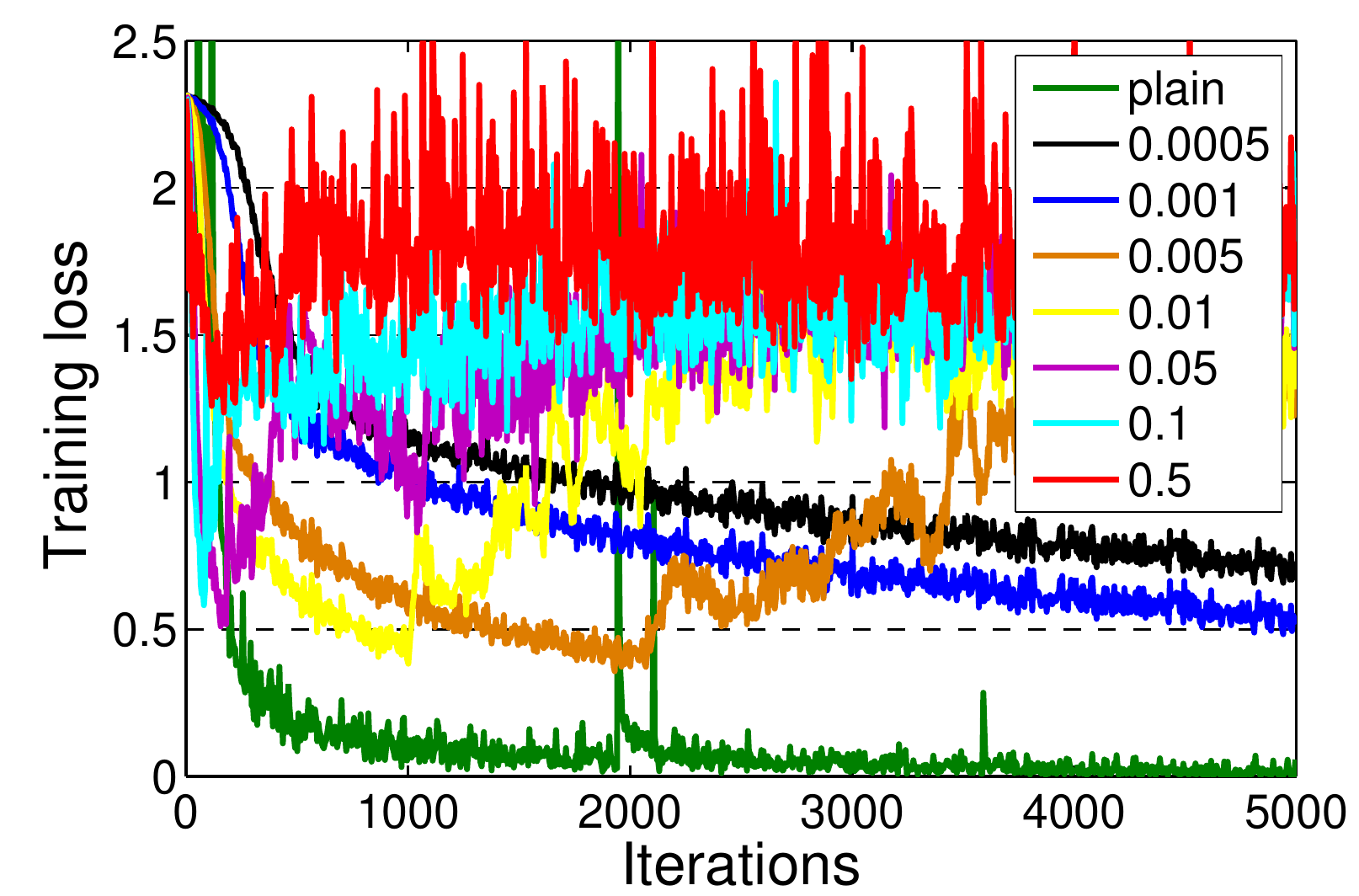}
  }
  \subfigure[CayT]{
  \includegraphics[width=0.23\linewidth]{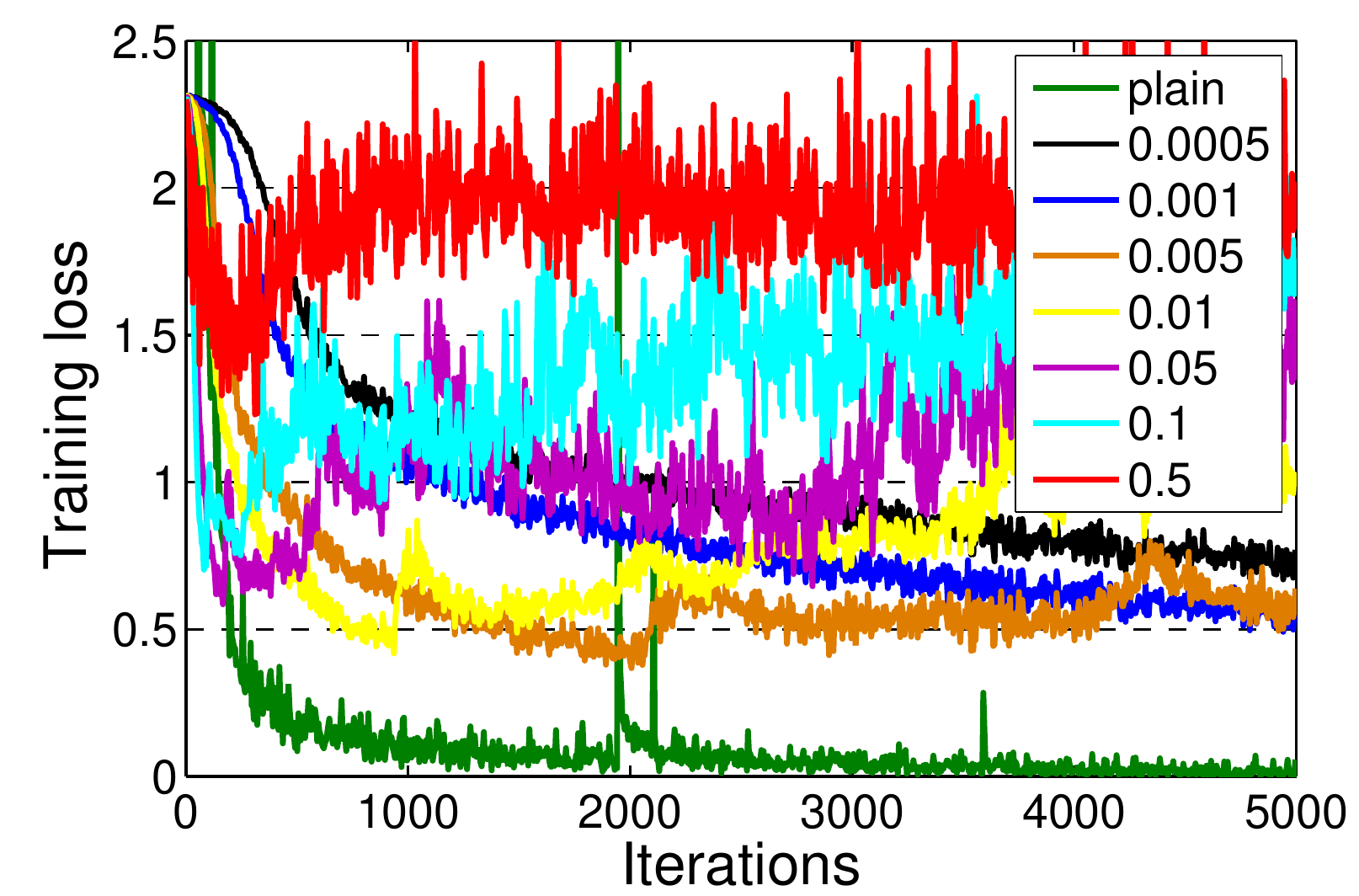}
  }
  \subfigure[Our OLM]{
  \includegraphics[width=0.23\linewidth]{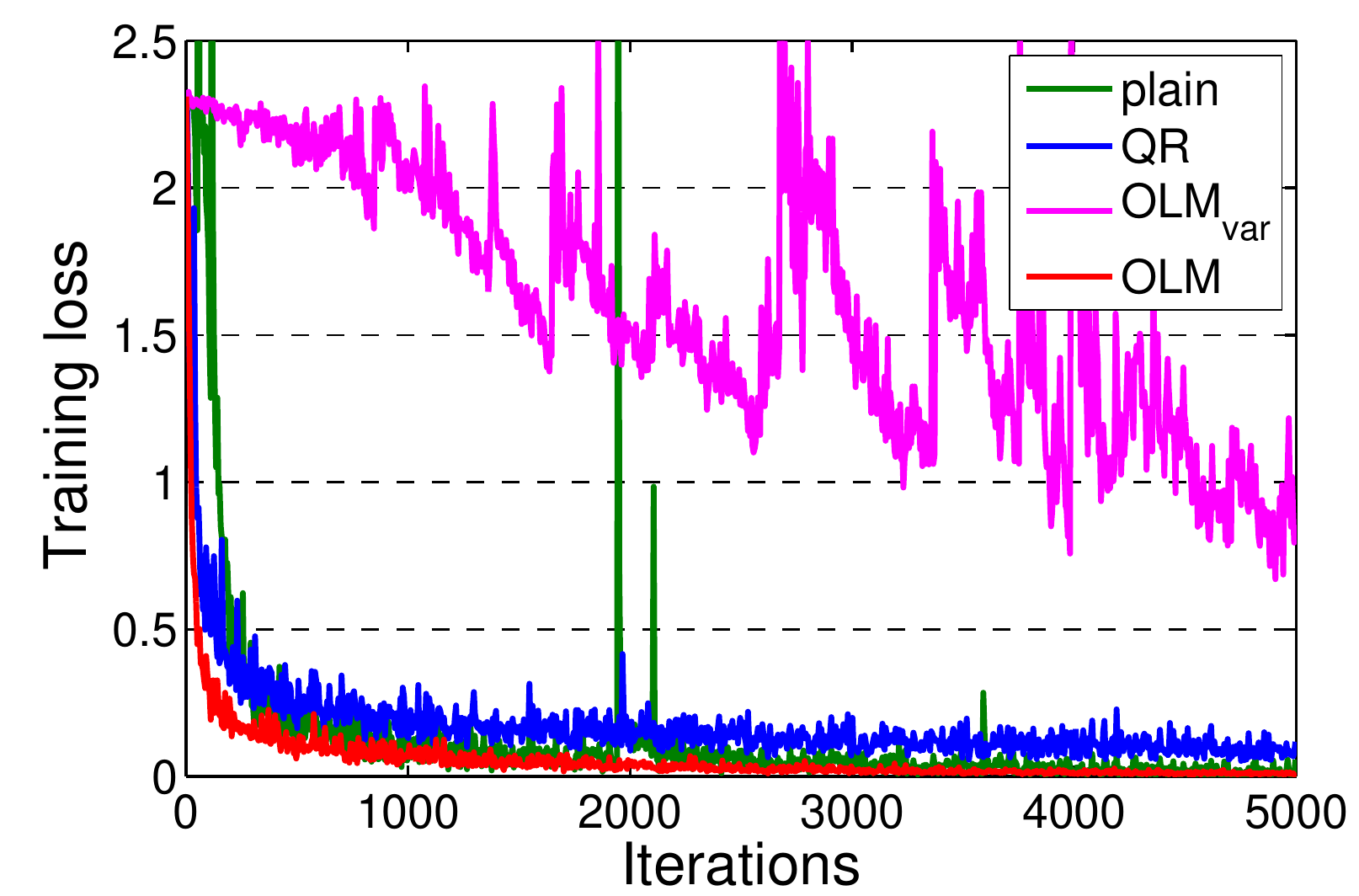}
  }
   \vspace{-0.15in}
  \caption{\small Results of Riemannian optimization  methods to solve OMDSM on MNIST dataset under the 4-layer MLP. We  train the model with batch size of 256 and show the training loss curves for different learning rate of `EI+QR', `CI+QR' and  `CayT' compared to the baseline `plain'  in (a), (b) and (c) respectably. We compare our methods to baselines  and report the best performance among all learning rates based on the training loss for each method in (d).}
  \label{fig:exp_MLP_4layer}
  \vspace{-0.1in}
\end{figure*}

\begin{figure*}[t]
\centering
  \vspace{-0in}
  \subfigure[EI+QR]{
  \includegraphics[width=0.23\linewidth]{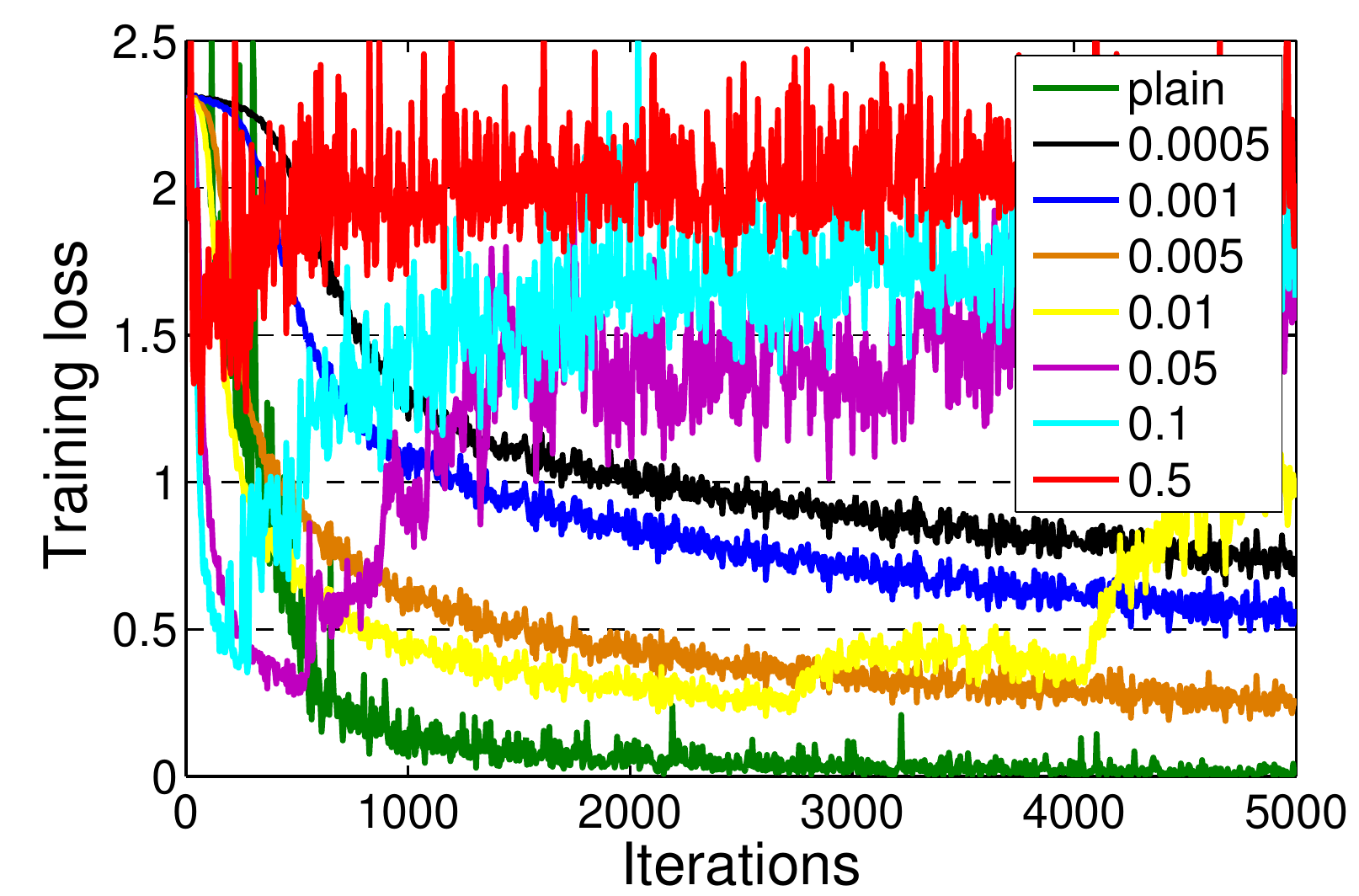}
  }
  \subfigure[CI+QR]{
  \includegraphics[width=0.23\linewidth]{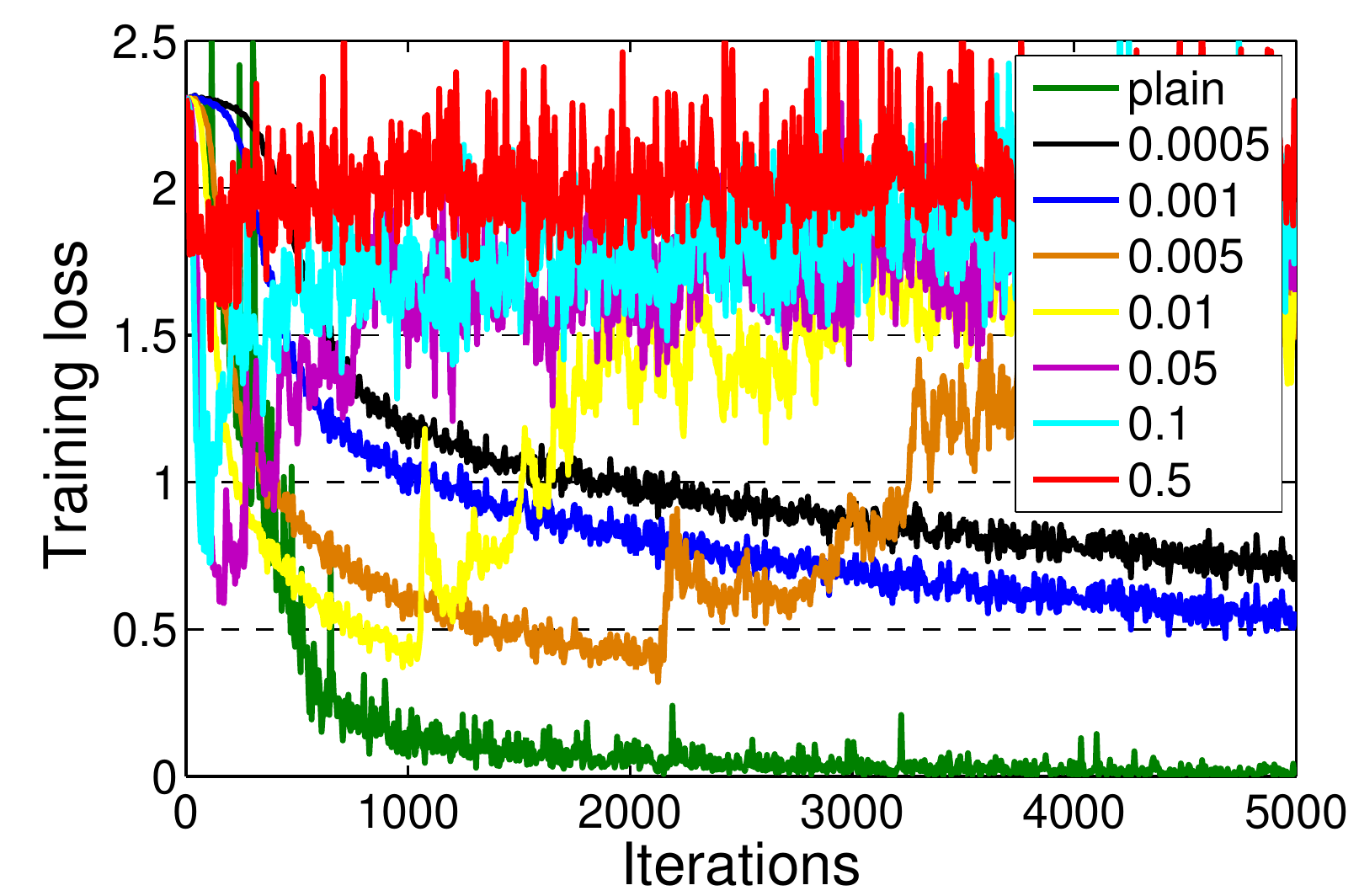}
  }
  \subfigure[CayT]{
  \includegraphics[width=0.23\linewidth]{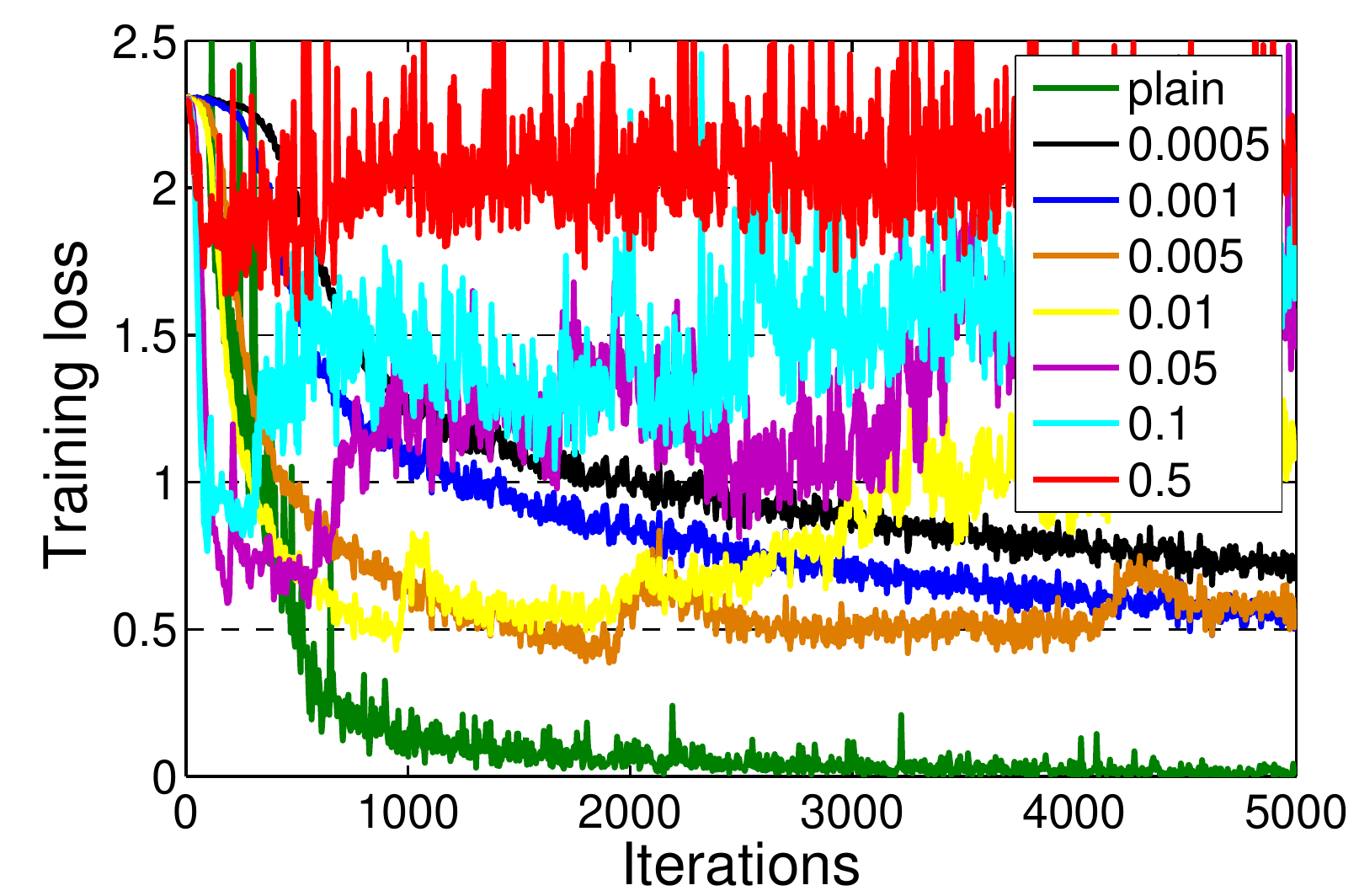}
  }
  \subfigure[Our OLM]{
  \includegraphics[width=0.23\linewidth]{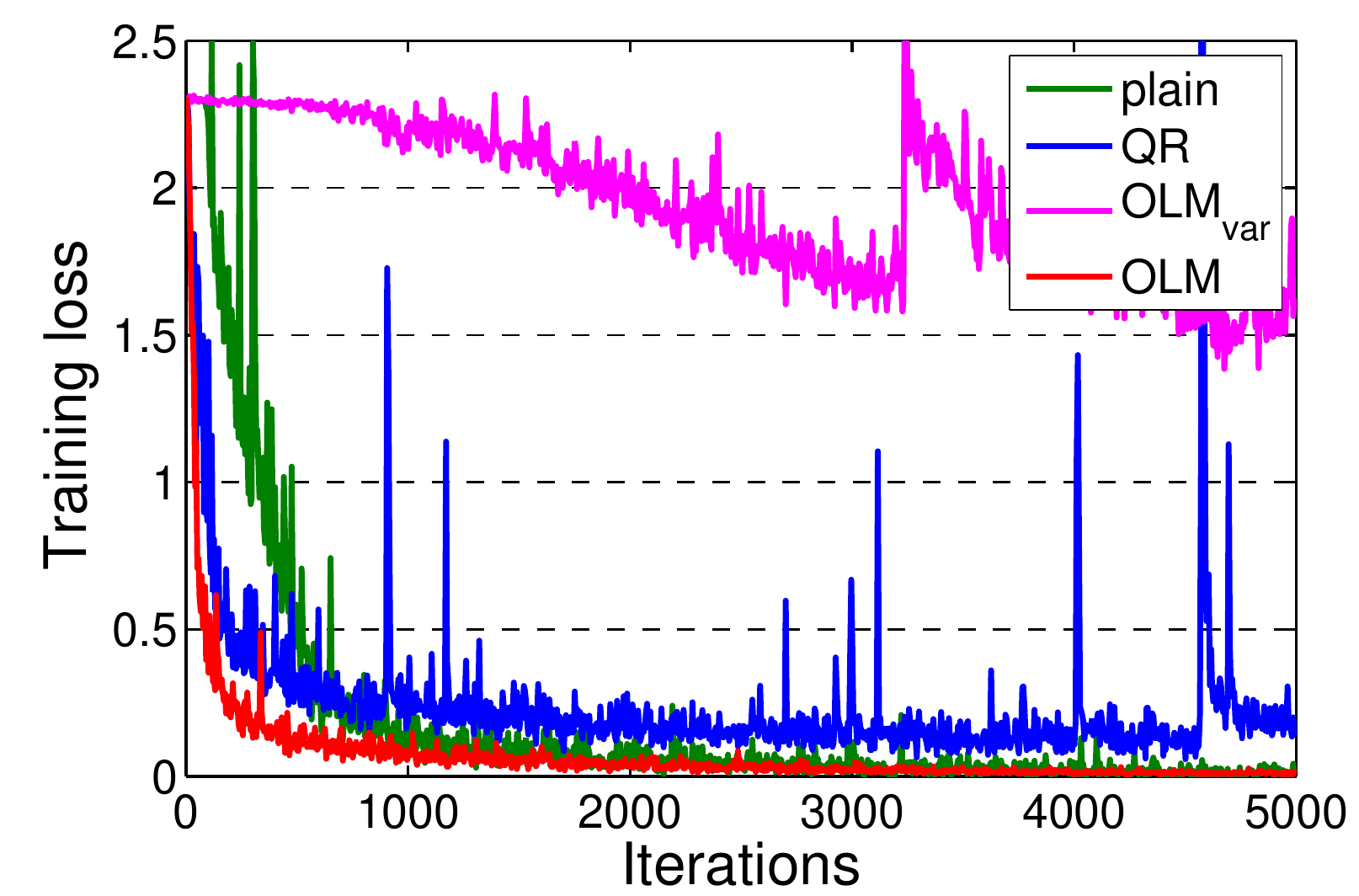}
  }
   \vspace{-0.15in}
  \caption{\small Results of Riemannian optimization  methods to solve OMDSM on MNIST dataset under the 6-layer MLP. We  train the model with batch size of 256 and show the training loss curves for different learning rate of `EI+QR', `CI+QR' and  `CayT' compared to the baseline `plain'  in (a), (b) and (c) respectably. We compare our methods to baselines  and report the best performance among all learning rates based on the training loss for each method in (d).}
  \label{fig:exp_MLP_6layer}
  \vspace{-0.1in}
\end{figure*}

\begin{figure*}[t]
\centering
  \vspace{-0in}
  \subfigure[EI+QR]{
  \includegraphics[width=0.23\linewidth]{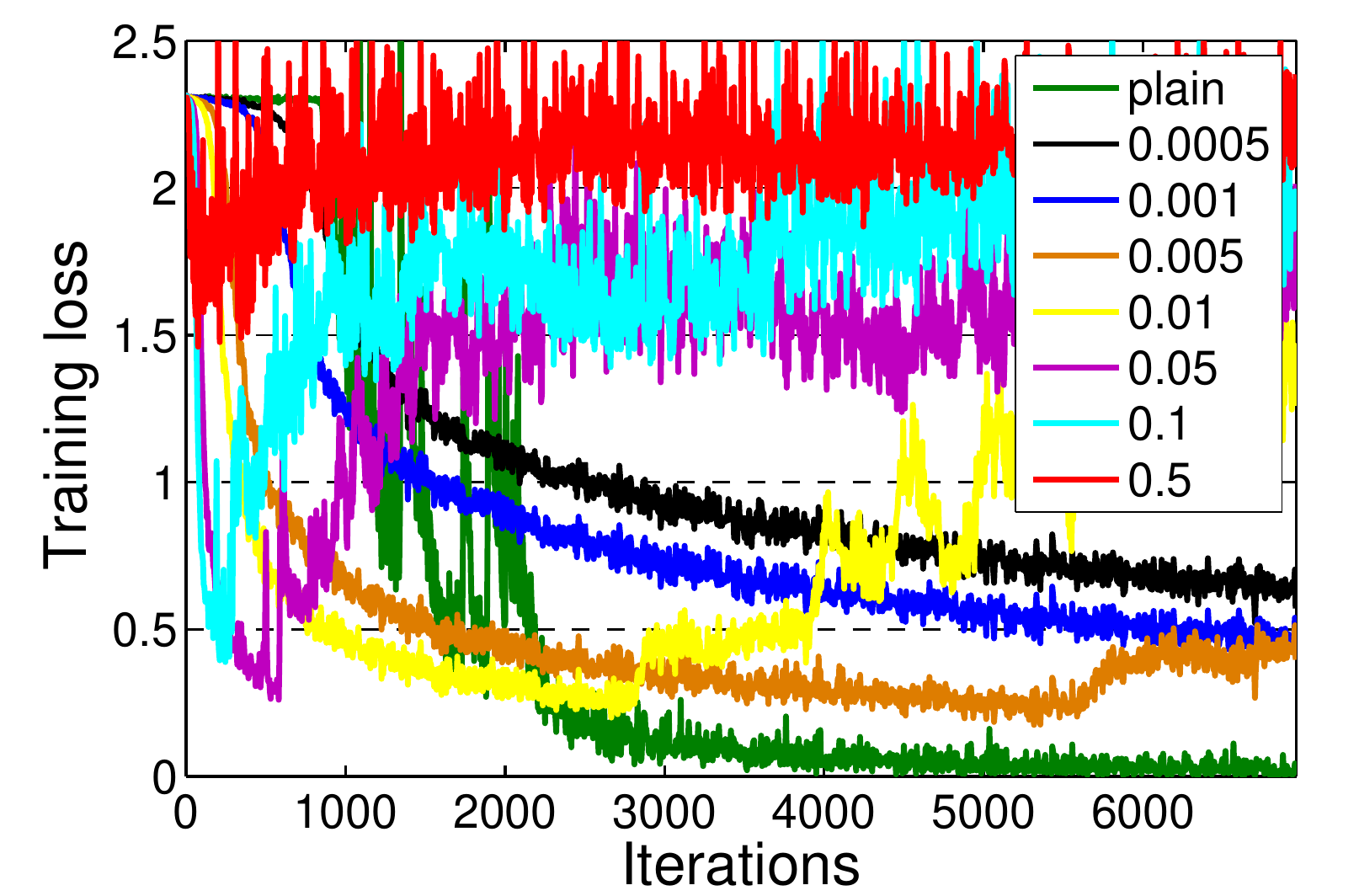}
  }
  \subfigure[CI+QR]{
  \includegraphics[width=0.23\linewidth]{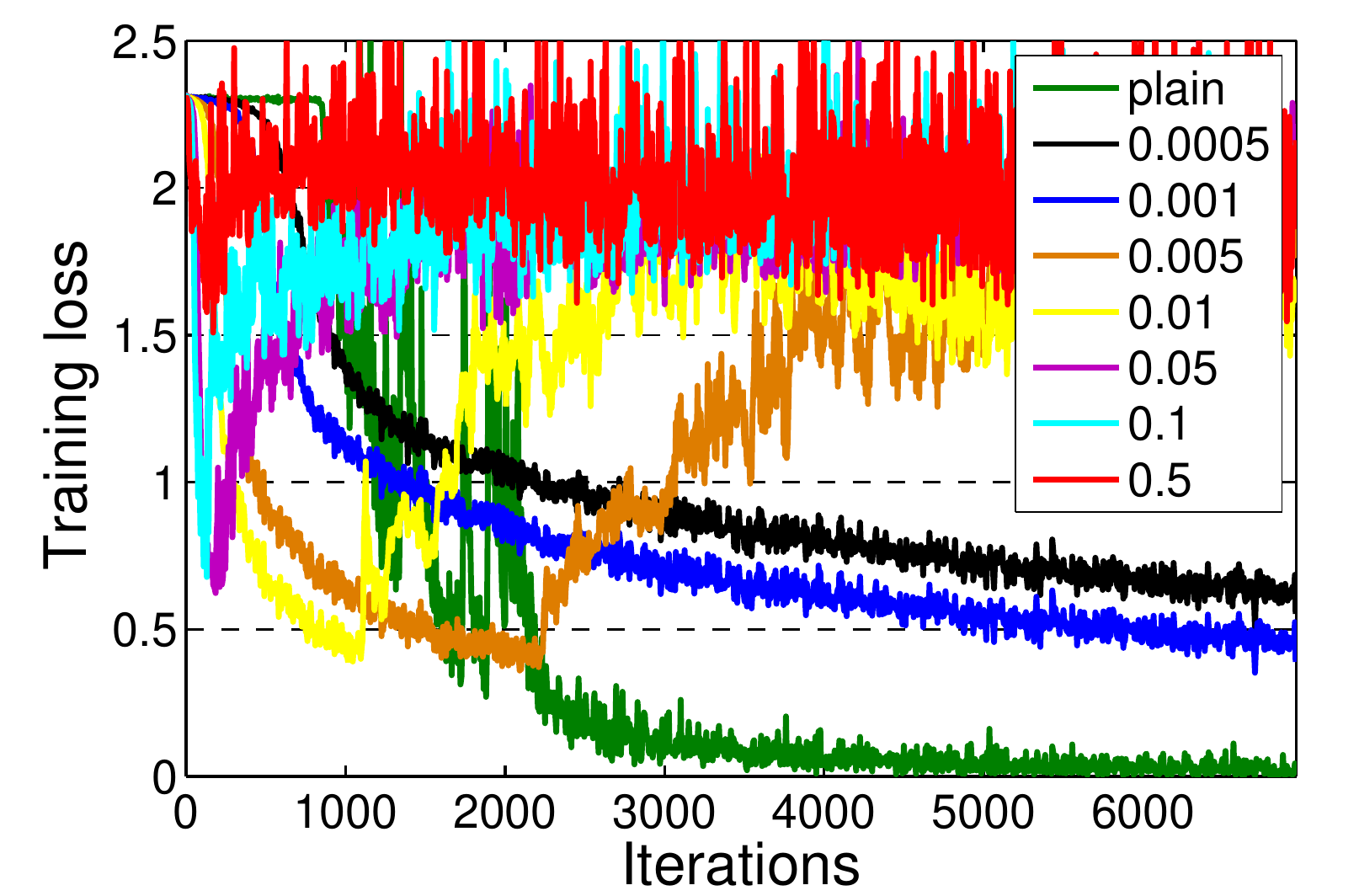}
  }
  \subfigure[CayT]{
  \includegraphics[width=0.23\linewidth]{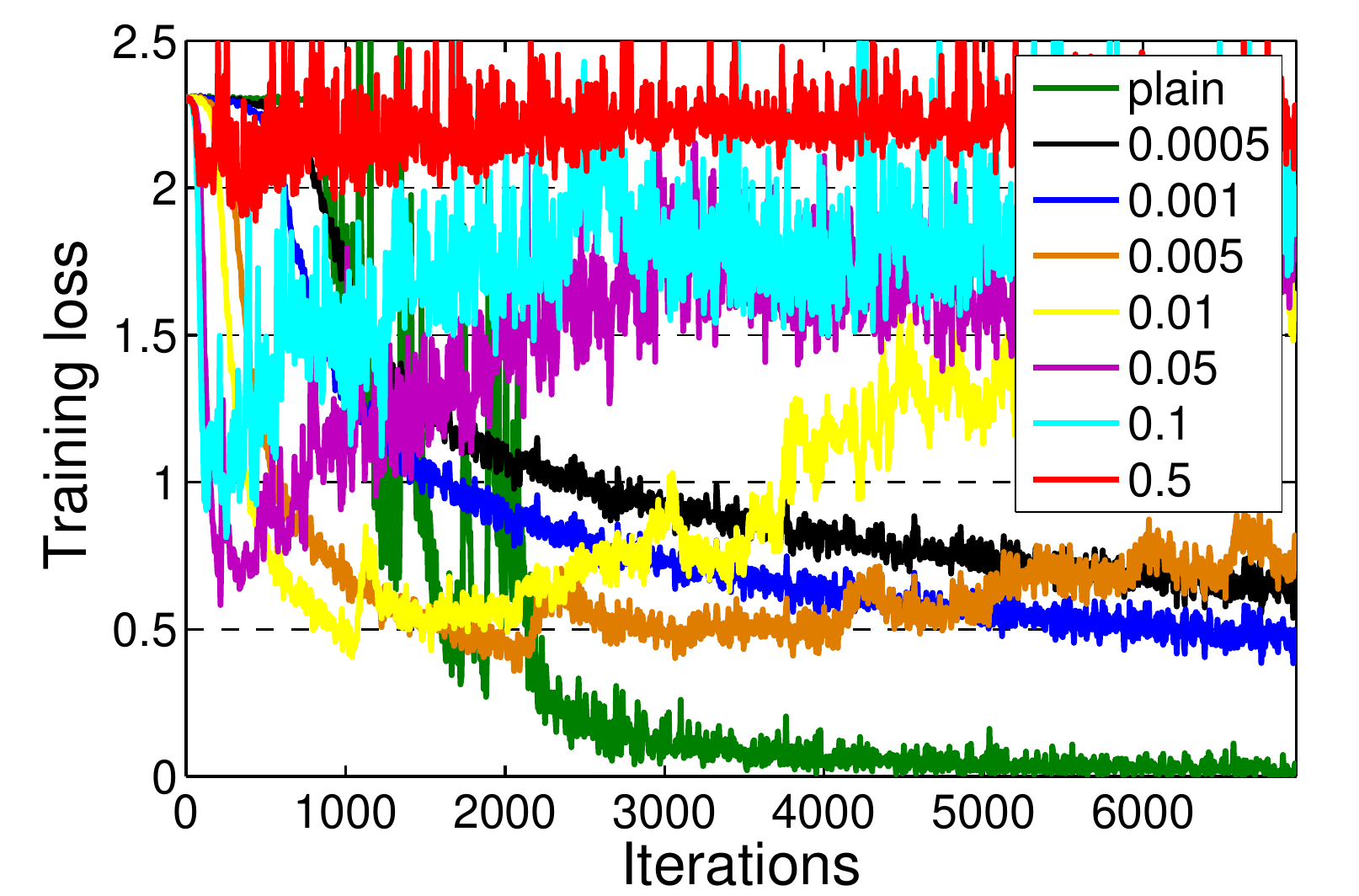}
  }
  \subfigure[Our OLM]{
  \includegraphics[width=0.23\linewidth]{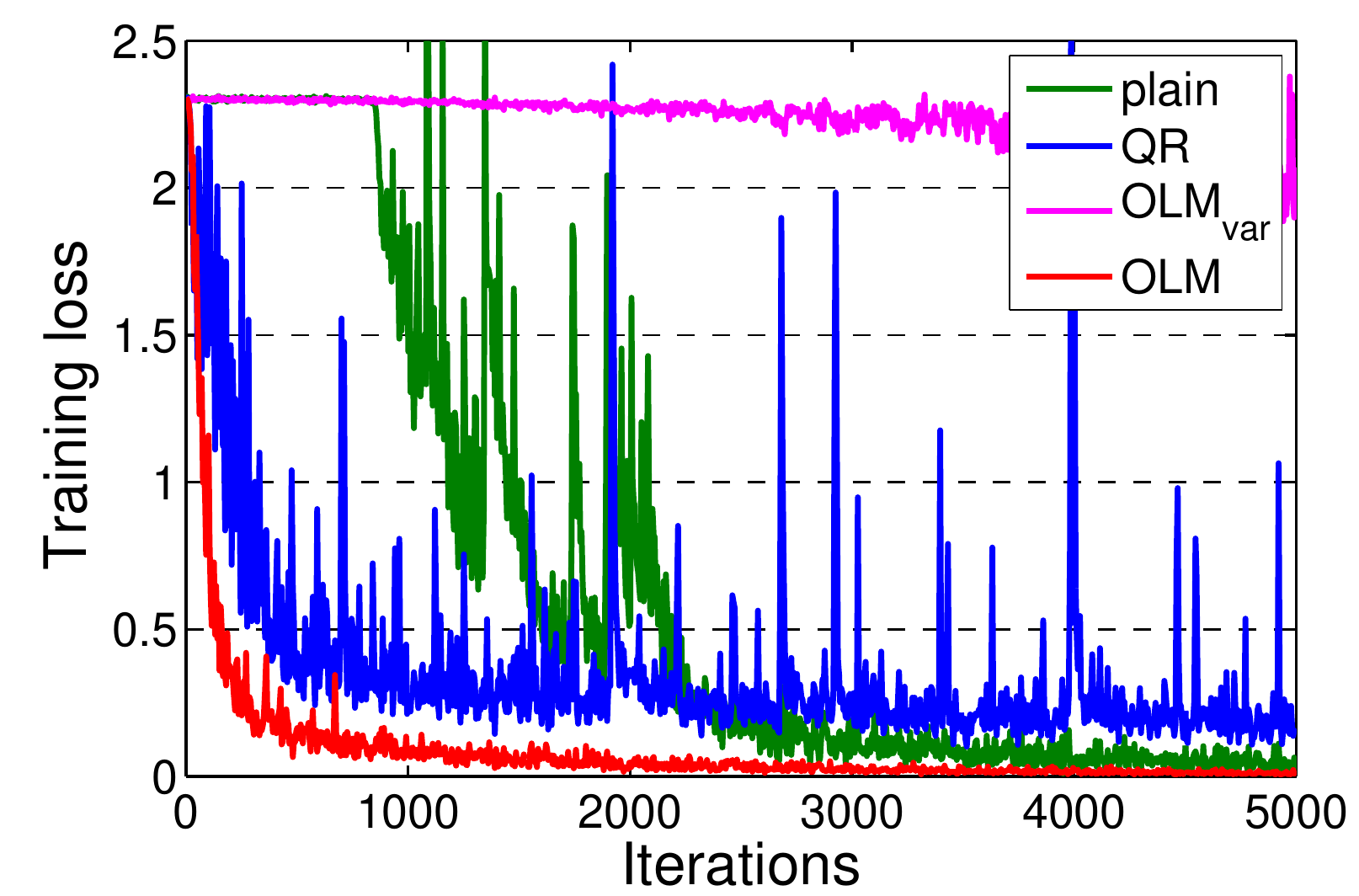}
  }
   \vspace{-0.15in}
  \caption{\small Results of Riemannian optimization  methods to solve OMDSM on MNIST dataset under the 8-layer MLP. We  train the model with batch size of 256 and show the training loss curves for different learning rate of `EI+QR', `CI+QR' and  `CayT' compared to the baseline `plain'  in (a), (b) and (c) respectably. We compare our methods to baselines  and report the best performance among all learning rates based on the training loss for each method in (d).}
  \label{fig:exp_MLP_8layer}
  \vspace{-0.1in}
\end{figure*}

Here, we show more experimental results for  Solving OMDSM on MNIST dataset under MLP architectures. The experimental setups are described in the paper. Figure \ref{fig:exp_MLP_b1024_6layer} and \ref{fig:exp_MLP_b1024_8layer} show the results under the respective 6-layer and 8-layer MLPs  with mini-batch size of 1024. We also train the model with mini-batch size of 512 and the results under the 4-layer, 6-layer and 8-layer MLPs are shown in Figure \ref{fig:exp_MLP_b512_4layer}, \ref{fig:exp_MLP_b512_6layer} and \ref{fig:exp_MLP_b512_8layer} respectively.  We further train the model with mini-batch size of 256 and the results under the 4-layer, 6-layer and 8-layer MLPs are shown in Figure \ref{fig:exp_MLP_4layer}, \ref{fig:exp_MLP_6layer} and \ref{fig:exp_MLP_8layer}.

These comprehensive experiments strongly support our empirical conclusions that: (1) Riemannian optimization methods probably do not work for the OMDSM problem, and if work, they must be under fine designed algorithms or tuned hyper-parameters; (2) deep feed-forward neural networks (e.g., MLP in this experiment) equipped with orthogonal weight matrix is easier for optimization by our `OLM' solution.

\end{document}